\ifdefined\pdfminorversion
\fi
\documentclass[11pt]{article}
\usepackage[letterpaper,textwidth=17cm,textheight=22cm,centering]{geometry}
\usepackage[authoryear,sort&compress,round]{natbib}
\usepackage{mathpazo}
\usepackage[scaled=0.96]{helvet}
\usepackage{fancyhdr}
\usepackage{mathtools}
\usepackage{multirow}
\usepackage{algorithm}
\usepackage{algorithmic}
\usepackage{caption}
\usepackage{subcaption}
\usepackage{adjustbox}
\usepackage{tabularx}
\usepackage{svg}
\usepackage{wrapfig}
\usepackage{titlesec}
\usepackage[most]{tcolorbox}

\usepackage{latexsym}
\usepackage[T1]{fontenc}
\usepackage[utf8]{inputenc}
\usepackage{microtype}
\usepackage{inconsolata}

\usepackage{graphicx}
\graphicspath{{./}}
\usepackage{tikz,fontawesome5}
\usepackage{xurl} \usepackage{array}
\usepackage{booktabs}
\usepackage{longtable}
\usepackage{colortbl}
\usepackage{placeins}
\usepackage{float}
\usepackage{enumitem}
\usepackage{amsmath}
\usepackage{amsfonts}
\usepackage{amssymb}
\usepackage{nicefrac}
\usepackage{fvextra}
\usepackage{xcolor}
\usepackage[normalem]{ulem}
\usepackage{hyperref}
\hypersetup{
  colorlinks=true,
  linkcolor=blue,
  citecolor=blue,
  urlcolor=blue
}

\newcommand{\OneRIUPerParticipant}{One PIU per participant}
\newcommand{\PairwiseRIUOverlap}{Pairwise PIU overlap}
\newcommand{\ExtraParticipantWithoutRIU}{Extra participant without a PIU}
\newcommand{\OneCommonRIU}{One common PIU}
\newcommand{\AllButOneRIUConcentration}{All-but-one PIU concentration}
\newcommand{\CoordinatorSelectedSpeaker}{Coordinator-selected speaker}
\newcommand{\BroadcastQuery}{Broadcast query}
\newcommand{\RoundRobin}{Round robin}
\newcommand{\CurrentSpeakerSelection}{Current-speaker selection}
\newcommand{\SimultaneousResponses}{Simultaneous responses}

\newcommand{\PairwiseRIUOverlapCell}[1][c]{\makecell[#1]{Pairwise PIU\\overlap}}
\newcommand{\ExtraParticipantWithoutRIUCell}[1][c]{\makecell[#1]{Extra participant without\\a PIU}}
\newcommand{\AllButOneRIUConcentrationCell}[1][c]{\makecell[#1]{All-but-one PIU\\concentration}}
\newcommand{\BroadcastQueryCell}{\makecell{Broadcast\\query}}
\newcommand{\RoundRobinCell}{\makecell{Round\\robin}}
\newcommand{\CurrentSpeakerSelectionCell}{\makecell{Current-speaker\\selection}}
\newcommand{\SimultaneousResponsesCell}{\makecell{Simultaneous\\responses}}

\usepackage{cpdef}

\titlelabel{\thetitle.\quad}
  \renewenvironment{abstract}
   {\begin{tcolorbox}[
      enhanced,
      colback=black!6,
      colframe=black!6,
      boxrule=0pt,
      arc=2pt,
      left=10pt,
      right=10pt,
      top=8pt,
      bottom=8pt,
      before skip=0.25em,
      after skip=0.25em]
    \bfseries}
   {\par\vspace{0.55em}\normalfont\footnotesize\sffamily
   
   \noindent\href{https://teamgym.vercel.app/}{\textcolor{black!85}{\faGlobe\enspace Project Page}}\qquad
   \href{https://github.com/chanwoo-park-official/TeamGym}{\textcolor{black!85}{\faGithub\enspace Code}}\qquad
   \href{https://huggingface.co/datasets/TeamGym/teamgym-datasets}{\raisebox{-.30ex}{\includegraphics[height=2.1ex]{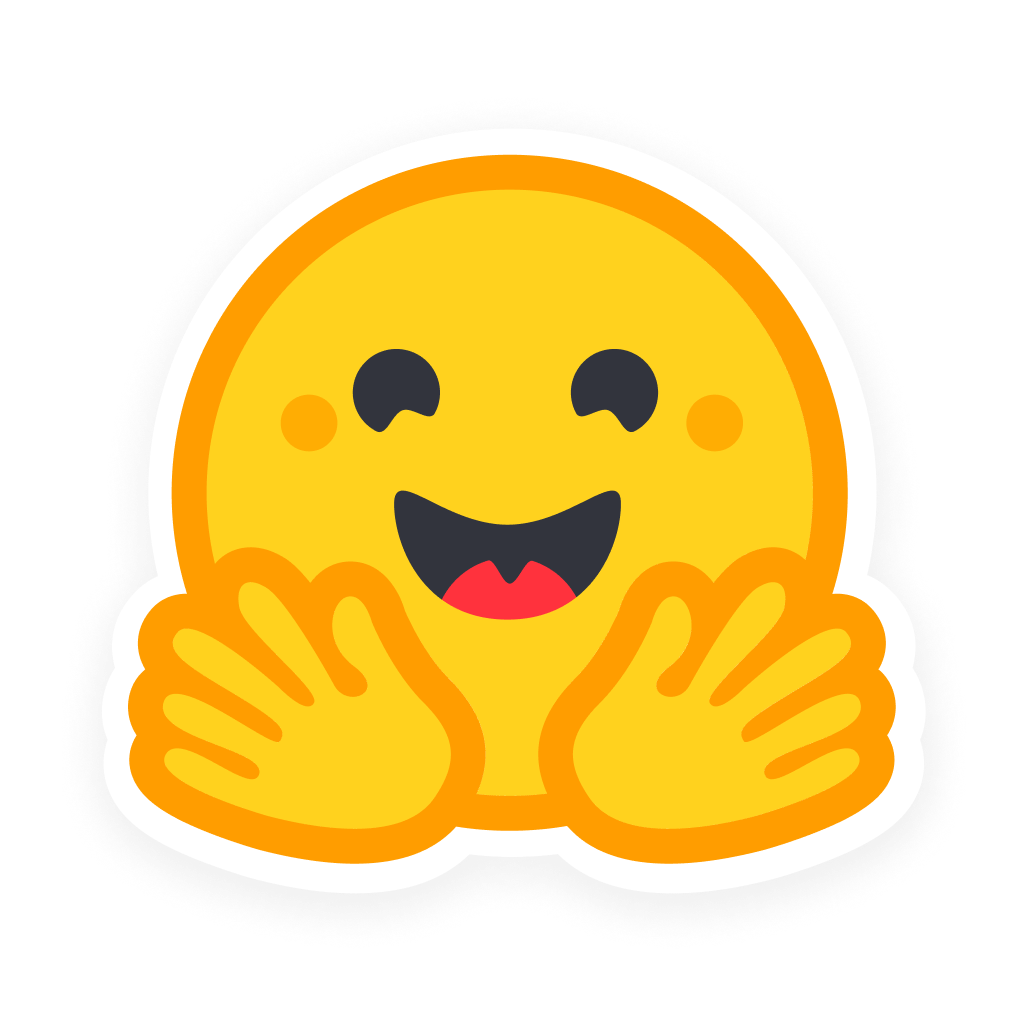}}\enspace\textcolor{black!85}{Hugging Face}}\par\end{tcolorbox}}

\makeatletter

  \renewcommand{\maketitle}{\bgroup\setlength{\parindent}{0pt}
  \vspace*{-2.5em}
  \begin{flushleft}
    \textbf{\LARGE \@title}\vspace*{1.25em}

    \textbf{\@author}
  \end{flushleft}\egroup}

\fancypagestyle{firstpagestyle}{\fancyhf{}

    \fancyhead[C]{\raisebox{22pt}{\makebox[\textwidth][l]{\raisebox{-0.5\height}{\includegraphics[height=30pt]{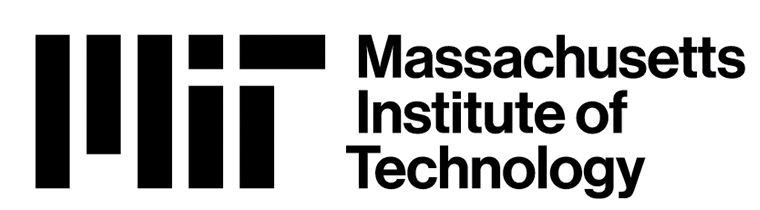}}}}}

}
\makeatother

\renewcommand{\tablename}{\textbf{Table}}
\renewcommand{\thetable}{\textbf{\arabic{table}}}

\makeatletter
\def\thickhline{\noalign{\ifnum0=`}\fi\hrule height 1.2pt
  \futurelet\@tempa\@xhline}
\providecommand*{\toclevel@subsubsubsection}{4}
\newcounter{subsubsubsection}[subsubsection]

\providecommand{\subsubsubsection}{\@startsection{subsubsubsection}{4}{\z@}{-1.5ex plus -0.5ex minus -.2ex}{0.5ex plus .2ex}{\normalfont\normalsize\bfseries\raggedright}}
\renewcommand{\paragraph}[1]{\par\vspace{1.5ex plus 0.5ex minus .2ex}\noindent{\normalsize\bfseries #1}\hspace{1em}\ignorespaces}
\makeatother

\newcommand{\safevspace}[1]{\vspace{0mm}}
\newcommand{\safesmallvspace}[1]{\vspace{0mm}}

\definecolor{armygreen}{rgb}{0.29, 0.33, 0.13}
\definecolor{cadmiumgreen}{rgb}{0.0, 0.42, 0.24}
\definecolor{darkgreen}{RGB}{0,100,0}
\DeclareRobustCommand{\revised}[1]{{#1}}
\newcommand{\ours}[1]{\texttt{TeamGym}}
\newcommand{\uncert}[1]{\,{\tiny$\pm$\,#1}}

\title{{\fontsize{24}{27}\selectfont\fontfamily{phv}\bfseries TeamGym:}\enspace
{\LARGE\fontfamily{phv}\bfseries A Gym for Team Decision-Making by AI}\\[0.18em]
{\LARGE\fontfamily{phv}\bfseries Agents with Distributed Information}}

\author{Chanwoo Park~$^{1,\star}$, Jaywoong Jeong~$^{2,\star}$,
Zachary Wojtowicz~$^{1}$, Yubin Kim~$^{1}$, Taekyun Lee~$^{3}$,\\
Kaiqing Zhang~$^{4}$, Asuman Ozdaglar~$^{1}$\\
\vspace{0.5em}
\normalfont{\small
$^{1}$MIT\quad
$^{2}$KAIST\quad
$^{3}$UT Austin\quad
$^{4}$University of Maryland}\\
\normalfont{\small $^{\star}$Equal contribution.}\vspace{0.25em}}

\newcommand{\inputdetailevaluationappendix}{\subsection{Trajectory Examples}
\label{app:trajectory-examples}

We present two representative episode trajectories that illustrate how PIUs
move from participants' initial local observations into the communication
transcript and final decision\revised{, followed by a matched set of four Finance episodes}.

\subsubsection{Dependency Case and Episode Trajectory: CVS 2022}

\paragraph{Problem and dependency structure.}
This representative FinLongDocQA problem asks for the ratio
of CVS Health's unique digital customers to 2022 revenue in each of three
segments, expressed as millions of customers per \$1 billion of revenue, and
then asks which segment has the largest ratio. Under the \OneRIUPerParticipant{}
condition, Participant A holds the common customer count,
while Participants B--D each hold one segment denominator.

\begin{center}\small
\setlength{\tabcolsep}{5.8pt}
\begin{tabular*}{0.998\textwidth}{@{\extracolsep{\fill}}c
>{\raggedright\arraybackslash}p{0.32\textwidth}
>{\raggedright\arraybackslash}p{0.19\textwidth}
>{\raggedright\arraybackslash}p{0.22\textwidth}@{}}
\toprule
PIU ID & PIU & \makecell{Participant assigned\\the PIU} & \makecell{Answer components\\using the PIU} \\
\midrule
1 & 47M unique digital customers & Participant A & Q1, Q2, Q3, Q4 \\
2 & Health Care Benefits revenue: \$91.4B & Participant B & Q1, Q4 \\
3 & Pharmacy Services revenue: \$169.2B & Participant C & Q2, Q4 \\
4 & Retail revenue: \$106.6B & Participant D & Q3, Q4 \\
\bottomrule
\end{tabular*}
\end{center}

The PIUs used by the four answer components are therefore
\[
Q_1=\{1,2\},\qquad Q_2=\{1,3\},\qquad
Q_3=\{1,4\},\qquad Q_4=\{1,2,3,4\}.
\]
Counting references component by component gives ten PIU uses, but only four distinct
PIUs.  No participant can compute even one requested ratio alone, and the
final comparison requires all four PIUs.

\paragraph{Reference calculation.}
Using the same 47M company-wide customer count as the numerator for each
requested segment calculation gives
\[
\frac{47}{91.4}=0.51,\qquad
\frac{47}{169.2}=0.28,\qquad
\frac{47}{106.6}=0.44.
\]
Thus Health Care Benefits has the highest ratio, at 0.51M unique digital
customers per \$1B of revenue.

\paragraph{Observed success by evaluation condition.}
The table below reports the result for this problem instance under each model
and evaluation condition. A check means that all four answer components were satisfied; a cross
means that all four were missing in the final decision.

\begin{center}\small
\begin{tabular}{lccc}
\toprule
Model & Single-Agent All-PIU & Multi-Agent PIU-Only & Multi-Agent Full-Context \\
\midrule
Claude Sonnet 4.5 & $\checkmark$ & $\checkmark$ & $\checkmark$ \\
Claude Haiku 4.5 & $\checkmark$ & $\times$ & $\times$ \\
\bottomrule
\end{tabular}
\end{center}

\paragraph{Successful episode trajectory: Sonnet 4.5.}
In Single-Agent All-PIU, the model directly read the four facts, computed
0.51, 0.28, and 0.44, and selected Health Care Benefits. In both multi-agent
conditions, all four PIUs entered the communication transcript in the responses
to the first four coordinator queries:
A disclosed 47M; B disclosed \$91.4B; C disclosed \$169.2B; and D disclosed
\$106.6B.  The coordinator used the same company-wide customer count in all
three ratios, performed the divisions, and returned the same ranking.  The semantic
judge found all four answer components correct and grounded in all three conditions.

\paragraph{Failure episode trajectory 1: Haiku 4.5, Multi-Agent PIU-Only.}
The coordinator issued 15 queries to A--D, but every
participant response was empty. No PIU entered the communication transcript.
The final decision explicitly refused to calculate because it lacked the customer
count and segment revenues. The final decision consequently omitted all four
answer components. This is a communication failure rather than an arithmetic
failure: the same model solved the problem in Single-Agent All-PIU.

\paragraph{Failure episode trajectory 2: Haiku 4.5, Multi-Agent Full-Context.}
Participant A found the consolidated 47M customer count, but initially reported
that the needed segment table was unavailable.  Participant C then incorrectly
attached \$169.2B to Health Care Benefits and could not recover the other two
denominators; later participant messages converged on the claim that the
accessible excerpts were insufficient. After 15 coordinator queries, the
coordinator returned ``unable to complete'' instead of the three ratios.  The
final decision omitted all four answer components. Here the source material
created an evidence-retrieval and fact-mapping failure even though every
PIU remained present in the source material provided to its assigned participant.

\paragraph{Case-study implication.}
The evaluation holds the problem instance fixed while exposing two distinct
collaboration bottlenecks. Multi-Agent PIU-Only can fail because a PIU in a participant's initial local observation never
enters the communication transcript; Multi-Agent Full-Context can fail because relevant facts are
lost or misassigned within much longer source material. In Single-Agent All-PIU,
both models solve the problem instance when the PIUs are centralized.

\subsubsection{Composite-Problem Episode Trajectory: Kirkland's 2023}

\paragraph{Generation is not execution.}
This representative FinDVer problem contains six questions, but
Q1--Q5 are not generated during the episode and Q6 is not appended afterward.
All six questions, their links to PIUs, the reference decision, and
the PIU distribution are frozen together as one composite problem instance before
evaluation.  The episode tests whether a coordinator can recover and combine
that already-fixed evidence.

\begin{quote}
\textbf{Data generation:}\quad
10-K source $\rightarrow$ candidate units $\rightarrow$ six PIUs
$\rightarrow$ Q1--Q5 analyses and Q6 synthesis $\rightarrow$ verification of the links between answer components and PIUs
$\rightarrow$ PIU distribution $\rightarrow$ retained problem instance.

\smallskip
\textbf{Evaluation execution:}\quad
initial common observation $\rightarrow$ coordinator queries $\rightarrow$
evidence accumulates in the communication transcript $\rightarrow$ Q1--Q5 and Q6
final decision $\rightarrow$ answer-component correctness and grounding judgment.
\end{quote}

The construction question is therefore ``which facts must jointly determine
the reference decision?''; the evaluation question is ``can the model find those facts and
combine only what the filing supports?''

\paragraph{From filing disclosures to one analyst problem.}
The six-PIU problem joins four review axes that an equity or credit analyst would
normally consider together.

\begin{center}\small
\begin{tabular}{lp{0.45\linewidth}l}
\toprule
Review axis & Filing evidence & Composite item \\
\midrule
Supply chain & Sourcing diversification and country-of-origin mix & Q1 \\
Profitability & 310-bp gross-margin improvement and drivers & Q2 \\
Execution risk & Capital constraints and tightly managed cash flow & Q3 \\
Financing & Term-loan terms, balance-date status, later draws & Q4--Q5 \\
Integrated judgment & Operating improvement plus liquidity pressure & Q6 \\
\bottomrule
\end{tabular}
\end{center}

Q6 asks whether the complete narrative is fully supported: Kirkland's was
improving operations through sourcing diversification and gross-margin
improvement, while still facing liquidity pressure and using new financing as
it executed a financial turnaround strategy.  This is not a vote over Q1--Q5;
it combines evidence with different scopes and dates into one bounded filing
interpretation.

\paragraph{Distributed evidence discovery, queries 1--6.}
Every participant had the generic participant description ``financial filing context
reviewer,'' so the coordinator initially did not know the PIU-ownership mapping.

\begin{center}\small
\begin{tabular}{rllp{0.39\linewidth}}
\toprule
Query & Participant & Item & Evidence recovered \\
\midrule
1 & A & Q4 & No term-loan balance at Feb. 3; \$34M credit borrowing \\
2 & B & Q1 & Diversification and fiscal 2023 country mix \\
3 & C & Q3, Q5 & Tight cash management; later \$6M and \$5M draws \\
4 & D & Q3 & Capital shortage could limit strategic execution \\
5 & E & Q4 & \$12M term loan; at least \$5M initial draw required \\
6 & F & Q2 & 310-bp margin improvement, drivers, and offset \\
\bottomrule
\end{tabular}
\end{center}

Query 1 illustrates genuine discovery rather than replay of a reference-decision ordering.
The coordinator asked A for sourcing information, but A did not own that PIU.
A instead disclosed its financing evidence.  The coordinator used the updated
communication transcript to query another reviewer for the still-missing sourcing component,
and recovered all six PIUs by query 6.

\paragraph{Cross-source consistency checks, queries 7--15.}
Once the decision-relevant facts were present, the remaining questions tested
how strongly they could be connected.  F confirmed that its margin paragraph
did not itself call the improvement part of the financial turnaround.  E
confirmed that its term-loan-terms paragraph did not state a strategic use of
proceeds.  By contrast, C's paragraph jointly disclosed later borrowing,
tightly managed cash flow, and execution of the financial turnaround strategy.
The defensible synthesis is therefore that the company used additional
financing while executing the turnaround under cash pressure--not that a
particular loan directly caused or funded a particular operating improvement.

\paragraph{Successful three-stage episode trajectory: Sonnet 4.5, Multi-Agent PIU-Only.}
\textbf{1. Distributed evidence discovery:} queries 1--6 mapped the six anonymous
reviewers to sourcing, margin, capital-risk, term-loan, balance-date, and
post-year-end evidence. \textbf{2. Checking support for cross-passage conclusions:}
queries 7--15 distinguished direct filing statements from unsupported
cross-passage inferences and
preserved the February 3 balance-date versus subsequent-borrowing distinction.
\textbf{3. Integrated conclusion:} the final decision reported Q1--Q5 and
concluded that Q6 was fully supported. The episode used the complete
budget of 15 participant responses, and all six answer components were
correct and grounded.

\paragraph{Efficiency and interpretation.}
The episode trajectory is deliberately informative rather than optimal.  An initial
misdirected query was necessary because PIU ownership was hidden, whereas
several queries after query 6 repeated checks of whether the filing supported
the proposed cross-passage conclusions after the core facts were already
complete. For presentation, the 15 coordinator queries and responses are most
clearly compressed into the three stages above.  The example then shows more
than parallel subproblem solving: it shows distributed evidence recovery,
verification of permissible cross-source links, and construction of one
realistic operational-and-liquidity judgment.

}

\begin{document}

  \maketitle
  \thispagestyle{firstpagestyle}
  {\footnotesize\begin{abstract}
\safesmallvspace{}
Many real-world decisions require teams to integrate information distributed across participants. Yet AI agents are still primarily trained and benchmarked in isolation.
We introduce \ours{}, an environment for evaluating and training language model-based agents on team decision-making problems constructed from real-world legal, finance, medical, and software engineering corpora. \ours{} introduces a method to identify a minimal set of facts, each element of which is necessary and all of which are collectively sufficient for a correct final decision, which we call \emph{pivotal information}. This enables \ours{} to both control the distribution of pivotal information and track how well agents identify and integrate pivotal information when making team decisions. 
Examining 8 language models on 12 source corpora, we find that current AI agents often fail to effectively share information: Distributing pivotal information across participants lowers success rate by 22.9 percentage points on average relative to giving the same information to a single model (a \emph{communication gap}). We also document that participants do not effectively retrieve decision-relevant information on their own: A team's success rate is lowered by a further 19.9 points when participants must identify and extract pivotal information from source material rather than receiving it directly (a \emph{retrieval gap}).
To help remedy these deficits, we use \ours{} to train policies through reinforcement learning on team decision loss. Across two policy-training experiments, we find that training improves final-decision quality and induces emergent coordination in participation, speaker selection, and communication, even though the reward evaluates only the final decision.
\end{abstract}
}

\vspace{0.25em}
\noindent\begin{minipage}{\textwidth}
    \centering
    \makebox[\textwidth][c]{

        \includegraphics[width=\textwidth]{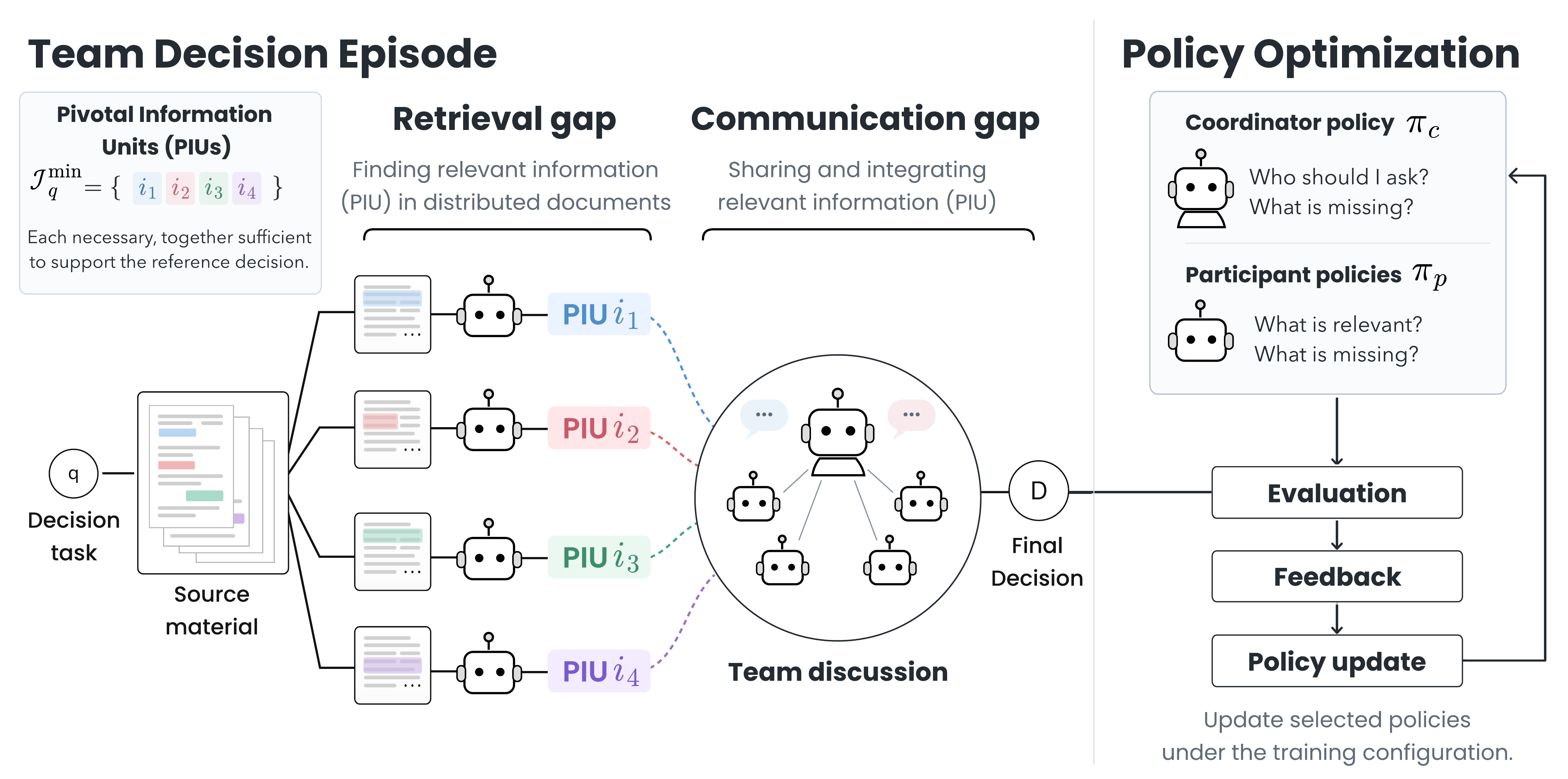}

    }
    \captionsetup{font=footnotesize,skip=2pt}
    \captionof{figure}{\revised{\textbf{\ours{} overview.}
    \ours{} constructs problem instances with pivotal information, evaluates
    retrieval and communication gaps, and trains participant and coordinator
    policies.}}
    \label{fig:teamgym-overview}
\end{minipage}

\safevspace{}
\safevspace{}
\safesmallvspace{}

\section{Introduction}
\label{sec:introduction}
\safevspace{}
\safesmallvspace{}

The frontier of AI is turning to multi-agent systems, in which AI agents are organized into teams that communicate to solve problems \citep{li2023camel,wu2023autogen,hong2024metagpt, kim2026capable}. 
Yet AI models are still trained and benchmarked largely as individual agents, which raises the question of whether they acquire the capacities and dispositions needed to be effective teammates. 
In human groups, collective intelligence, which is linked to social sensitivity and turn-taking, predicts team performance better than individual intelligence \citep{woolley2010collective}. Although AI agents differ from humans, this finding motivates evaluating them as teams rather than only as individuals.

\safesmallvspace{}
A central challenge in team decision-making is that decision-relevant information is typically distributed across members. Reaching a collective decision requires members to identify and retrieve relevant information, communicate it efficiently, and integrate it into a final decision. Team theory has long studied this challenge \citep{radner1962team,marschak1972economic}, while empirical research on human group decision-making finds that people often struggle to identify and integrate critical private information \citep{stasser1985pooling,lu2012twenty}.

\safesmallvspace{}
We introduce \ours{}, a gym for constructing team decision problem instances from source corpora, and for evaluating and training participant and coordinator policies under controlled information distributions and communication structures. 
\ours{} targets questions that existing benchmarks and multi-agent systems leave open. 
Single-agent benchmarks measure individual model capabilities \citep{wu2024smartplay,liu2024agentbench}, while research on multi-agent LLM systems focuses on designing harnesses that solve a given problem \citep{li2023camel,wu2023autogen,hong2024metagpt, kim2026teambench}. 
\revised{\ours{} examines how participants retrieve and communicate decision-relevant information, how team performance varies with information distribution and communication structure, and how training can improve participant and coordinator policies.}

\paragraph{Method.}
\ours{} constructs controlled team decision problem instances from real-world
source corpora. For each instance, it derives a decision objective from one or
more corpus questions and either retains a corpus-provided reference decision
or constructs one from corpus annotations and source materials. Source
annotations and reasoning associated with the reference decision guide the
selection of candidate evidence units, such as sentences, paragraphs, tables,
or code blocks. Source-specific grounding and model-based complete-set and
leave-one-out checks retain units that together support the reference decision
but whose individual removal leaves it unsupported or unresolved. \revised{These units are \textit{pivotal information units (PIUs)} and form
 the instance's \textit{pivotal information}.}

\safesmallvspace{}
Holding the problem instance and its pivotal information fixed, \ours{} can vary who
initially has each PIU (the \textit{PIU distribution}), whether participants receive
PIUs directly or source material containing them, the communication structure,
and token budgets. The team collectively receives the pivotal information, but
the evidence is distributed across participants to make communication necessary
for a fully supported final decision. The communication structure specifies whether a
coordinator controls communication, how speakers
are selected, and which decision maker makes the final decision. Token budgets
constrain what decision makers may observe and generate. To understand what
reduces team performance, we define two success-rate gaps. The
\textit{communication gap} is the drop when the same PIUs are distributed
across participants rather than given to one model. The \textit{retrieval gap}
is the further drop when the team must retrieve those PIUs from source material
rather than receive them directly.
Beyond evaluation, \ours{} supports training participant and coordinator
policies under controlled PIU distributions, communication structures, and
token budgets.

\safesmallvspace{}
We construct problem instances from source corpora in
Table~\ref{tab:introduction-source-corpora}
\citep{hendrycks2021cuad,wang2023maud,ecfrTitle12Part1,ecfrTitle12Part3,yu2027mslr,shen2022multilexsum,holzenberger2020sara,zhao2024findver,wang2026finlongdocqa,wisedocs2026mlcrbenchmark,wang2025evidencebench,raghavendra2026sweatlas,huang2026deepswe}.
Source materials include contracts, regulations, SEC filings,
biomedical papers, and source code, while the decision formats are
finite-choice, written, and numeric.

\safevspace{}
\paragraph{Findings.}
\revised{The resulting comparisons hold the problem instance and pivotal
information fixed. Each gap measures the combined effect of the changes between
its two evaluation conditions (Appendix~\ref{app:gap-scope}).}
\textbf{(1)} Across eight language models and 12 source corpora,
the decision success rate is 22.9 percentage points lower when the same
PIUs are distributed across a team than when one model receives them all, and
a further 19.9 points lower when the same team must retrieve its PIUs from
source material rather than receiving them directly. \textbf{(2)}
\revised{Across domains, relative communication-structure performance varies
across tested PIU distributions. For example, on finance problem instances, \RoundRobin{}'s
advantage over \CoordinatorSelectedSpeaker{} grows under concentrated PIUs.
Broadening or removing narrow coordinator queries reduces this gap and
improves evidence sharing with speaker selection fixed
(\S\,\ref{sec:cross-domain-riu-distribution-communication-structure-results}).
In SARA, withholding messages from participants given source material but
no assigned PIU improves final decisions.}
\textbf{(3)} Optimizing final-decision quality changes the
communication process itself. With coordinator control, joint two-policy
training produces a compact, exhaustive information-elicitation protocol;
without a coordinator, training the shared participant policy improves
end-to-end final-decision quality and completion and induces emergent
coordination in participation, speaker selection, and communication. Tracing
PIUs through these trajectories shows that increased participation makes more
pivotal information available on FinDVer validation, but not on held-out
FinLongDocQA.

\safesmallvspace{}
\ours{} therefore provides a controlled environment in which team
decision-making capabilities can be evaluated and trained under different
information distributions and communication structures.

\safesmallvspace{}
\safevspace{}
\section{Related Work}
\label{sec:related-work}
\safevspace{}
\safevspace{}

\paragraph{Team decision theory and organizational economics.}
Team decision theory studies cooperative decision makers with a team objective
and a decentralized information structure. A team problem couples a shared
objective with an information structure and one policy for each decision maker
\citep{radner1962team,marschak1972economic,yuksel2013networked}. Classical
analysis largely studies the properties of different computation structures and policies, not how they can be learned and implemented computationally.
Under \textbf{tractable joint-policy optimization}
\phantomsection\label{assump:strategic-computation}\textbf{(A1)}, computing this
collection is assumed feasible before execution, although the exact problem can
be intractable even in finite-horizon decentralized control
\citep{bernstein2002complexity}. \ours{} instead evaluates and trains
language-model policies, comparing policy collections before and after
training under the same conditions.

\safevspace{}
\paragraph{Mechanism design.}
Under the \textbf{direct-report mechanism assumption}
\phantomsection\label{assump:direct-type-report}\textbf{(A2)}, mechanism design
represents private information as a type that can be elicited directly under the applicable equilibrium assumptions, and that participants know their type and can communicate it if they so choose
\citep{maskin2002implementation,myerson1979incentive,myerson2008perspectives}.
In practice, however, attention and cognitive limitations constrain agent's ability to identify, extract, and communicate information contained in long source material, even if it is in principle available to them
\citep{mookherjee2006decentralization}. \ours{} assumes a shared team payoff
and studies bounded natural-language messages derived from source material.

\safevspace{}
\paragraph{Communication constraints.}
Research in organizational economics, planning, and reinforcement learning examines
communication costs and message limits
\citep{cremer2007language,wu2011online,kim2019learning,zhu2024surveycommunication}.
The idealization of \textbf{nonbinding communication resources}
\phantomsection\label{assump:unconstrained-communication}\textbf{(A3)} permits
decision makers to access and transmit relevant information without binding
observation or message constraints. \ours{} instead makes its resource budgets
explicit.

\safesmallvspace{}
We defer additional related work on multistage communication, human-AI and
group decision making, multi-agent LLM systems, benchmark contrasts, and the
source corpora to Section~\ref{app:related-work}.

\safevspace{}
\safesmallvspace{}

\section{\ours{} Environment}
\label{sec:interaction-interface}
\safevspace{}

\subsection{Decision Problem and Communication}
\label{sec:method}
\safevspace{}

\label{sec:mathematical-formulation}

We formalize \ours{} as a finite-horizon team decision problem
\citep{radner1962team,marschak1972economic,yuksel2013networked,nayyar2013common}.
A problem instance $q$ fixes underlying source materials, a decision objective,
a decision format, a set $\cD_q$ of possible final decisions, and a
problem-specific evaluator that assigns each $d\in\cD_q$ a team payoff
$U_q(d)\in[0,1]$.
A \textit{decision maker} is a role whose policy may generate a query, send a
message, or make the final decision. A \textit{participant} is a decision maker
that may send messages during communication.
A \textit{coordinator} is a decision maker that may control
communication by selecting or querying participants and may make the final
decision.
Initially, each participant
receives an initial local observation, while all decision
makers receive the same initial common observation containing the decision
objective and decision format. The policy $\rho_i$ maps the information
currently available to decision maker $i$ to a distribution over the queries,
messages, or final decisions permitted for its role; $\rho$ denotes the
collection of decision-maker policies.
The \textit{communication structure} $\cC$ specifies whether a coordinator
controls communication, which participants may send messages and when, whether
messages are sequential or simultaneous, and which decision maker makes the
final decision. The \textit{resource budgets} $B$, including token budgets,
constrain what decision makers may observe and generate and how often
participants may send messages. Queries and messages form the
\textit{communication transcript}; a decision maker's \textit{local transcript}
is the portion available to it under $\cC$ and $B$. The initial local and common
observations, together with the observation rules induced by $\cC$ and the
observation limits in $B$, define the \textit{decentralized information
structure}.
An \textit{episode} is one complete execution through the final-decision stage
for fixed $q$, initial observations, $\cC$, $B$, and $\rho$. For these fixed
inputs, the resulting interaction induces a distribution over final decisions,
and we evaluate $\rho$ by its expected team payoff,
$\mathbb{E}[U_q(d)]$. Holding $q$ and the initial observations fixed while
varying $\cC$, $B$, or $\rho$ changes the interaction conditions without
changing the problem instance.

\begin{figure}[htbp]
    \centering
    \includegraphics[width=\textwidth]{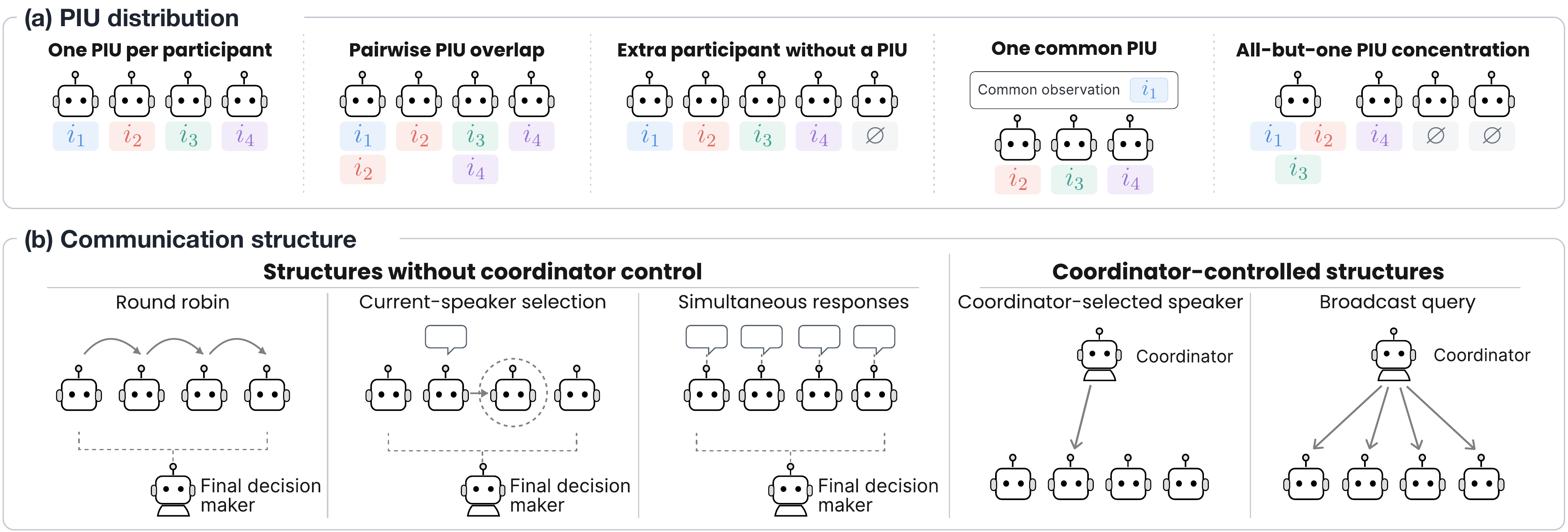}
    {\phantomsubcaption\label{fig:piu-distribution-conditions}}%
    {\phantomsubcaption\label{fig:communication-structure-taxonomy}}%
    \caption{\revised{\textbf{PIU distributions and communication structures.}
    \textbf{(a)} Colored boxes denote distinct PIUs, and \(\varnothing\) marks
    a participant assigned no PIU. PIU distributions preserve the problem
    instance and reference decision. \textbf{(b)} The first three protocols
    use a separate final decision maker. In the two coordinator-controlled
    protocols, the coordinator directs communication and makes the final decision.}}
    \label{fig:piu-communication-conditions}
\end{figure}

\label{sec:communication-structures}
\safesmallvspace{}
For evaluation, we consider five communication structures
(Figure~\ref{fig:communication-structure-taxonomy}), each proceeding in
scheduled steps called \textit{communication rounds}:
\textbf{(1) \RoundRobin{}}, under which one participant sends a message per
round in a fixed cyclic order; \textbf{(2)
\CurrentSpeakerSelection{}}, which starts with the first participant in that
order, after which each speaker selects the next participant; \textbf{(3)
\SimultaneousResponses{}}, under which all participants respond in one round;
\textbf{(4) \CoordinatorSelectedSpeaker{}}, under which the coordinator selects
and queries one participant per round; and \textbf{(5) \BroadcastQuery{}},
under which the coordinator sends the same query to all participants, whose
responses are generated in the same round. In \RoundRobin{},
\CurrentSpeakerSelection{}, and
\SimultaneousResponses{}, no coordinator controls communication, and a
separate final decision maker that does not participate in communication reads
the complete communication transcript and makes the final decision. In
\CoordinatorSelectedSpeaker{} and \BroadcastQuery{}, the coordinator controls
communication and makes the final decision.
Appendix~\ref{app:communication-structure-implementation} provides
message-generation and implementation details and connections to prior work.
\revised{Appendix~\ref{app:participant-resource-budget-details} reports the legal token-budget experiments.}

\FloatBarrier

\safevspace{}
\safevspace{}

\subsection{Problem-Instance Construction and Evaluation}
\label{sec:real-source-problems}
\safevspace{}

\ours{} constructs each problem instance $q$ from one source corpus using
materials such as contracts, SEC filings, biomedical papers, or source code.
Each $q$ is paired with a reference decision $d^{\mathrm{ref}}$ supported by
its source materials and used in construction and evaluation. For each $q$,
we identify an inclusion-minimal
set $\mathcal{I}^{\min}_q=\{i_1,\ldots,i_k\}$ called its \textit{pivotal information};
each $i_j$ is a \textit{pivotal information unit (PIU)}. Informally, this information
is sufficient to determine a final decision that achieves the highest possible
team payoff, whereas removing any one PIU leaves insufficient information to
guarantee such a decision.
Because DeepSWE admits multiple valid patches and costly repository execution,
it instead uses functional PIUs validated under a fixed, bounded protocol
(Appendix~\ref{app:deepswe-construction}).
Appendix~\ref{app:piu-formalization} formalizes this definition.
\safevspace{}

\paragraph{PIU construction.}
When a model's reasoning trace used to produce $d^{\mathrm{ref}}$ is available, we retain
it; otherwise, we generate reasoning conditioned on $d^{\mathrm{ref}}$,
following answer-conditioned rationale generation \citep{wang2023scott}.
To identify $\mathcal{I}^{\min}_q$, we use source annotations and the retained or generated
reasoning to locate candidate units in the source materials. Grounding and
model-based checks then test whether the candidates together support
$d^{\mathrm{ref}}$ and whether removing any one leaves the final decision
unsupported or unresolved. The candidates retained by these checks form
$\mathcal{I}^{\min}_q$.
We then verify that a model given all pivotal information can solve each
evaluated problem instance (Section~\ref{sec:problem-instances}).
Appendix~\ref{app:piu-construction-validation} reports an additional re-check
of PIU necessity on representative instances using a model not used for
construction. \revised{Appendix~\ref{app:piu-human-validation}
reports a separate human audit of PIU sufficiency and necessity.}

\begin{table}[htbp]
\centering
\scriptsize
\setlength{\tabcolsep}{2pt}
\renewcommand{\arraystretch}{0.94}
\caption{\textbf{\revised{Source corpora, decisions, and problem-instance counts.}}}
\label{tab:introduction-source-corpora}
\label{tab:problem-instance-counts}
\begin{tabular*}{\textwidth}{@{\extracolsep{\fill}}cclcc@{}}
\toprule
\textbf{Domain} & \revised{\textbf{Source corpus}} & \textbf{Decisions} &
\revised{\makecell{Constructed\\instances}} & \revised{\makecell{Evaluated\\instances}} \\
\midrule
\multirow{6}{*}{Legal}
& CUAD & \multirow{6}{*}{\makecell[l]{Contract profiles, merger reviews, compliance and enforcement,\\
legal and procedural conclusions, and tax calculations}} & 538 & 30 \\
& MAUD & & 557 & 30 \\
& eCFR Casebook & & 18 & 18 \\
& MSLR & & 11 & 11 \\
& Multi-LexSum & & 29 & 29 \\
& SARA & & 45 & 45 \\
\arrayrulecolor{black!18}\cmidrule(lr){1-5}\arrayrulecolor{black}
\multirow{2}{*}{Finance}
& FinDVer & \multirow{2}{*}{\makecell[l]{SEC-filing findings, calculations,\\comparisons, and classifications}} & 389 & 30 \\
& FinLongDocQA & & 249 & 30 \\
\arrayrulecolor{black!18}\cmidrule(lr){1-5}\arrayrulecolor{black}
\multirow{2}{*}{Medical}
& MLCR & \multirow{2}{*}{\makecell[l]{Evidence-grounded medical reviews\\and biomedical answers}} & 29 & 29 \\
& EvidenceBench & & 45 & 45 \\
\arrayrulecolor{black!18}\cmidrule(lr){1-5}\arrayrulecolor{black}
\multirow{2}{*}{\makecell[c]{Software\\engineering}}
& SWE Atlas--Codebase QnA & \multirow{2}{*}{\makecell[l]{Codebase QA and multi-turn, long-horizon repository-level\\
code repair with functionally evaluated patches}} & 32 & 32 \\
& DeepSWE & & 12 & 12 \\
\bottomrule
\end{tabular*}
\end{table}

\begin{figure*}[!h]
    \centering
    \includegraphics[width=\textwidth]{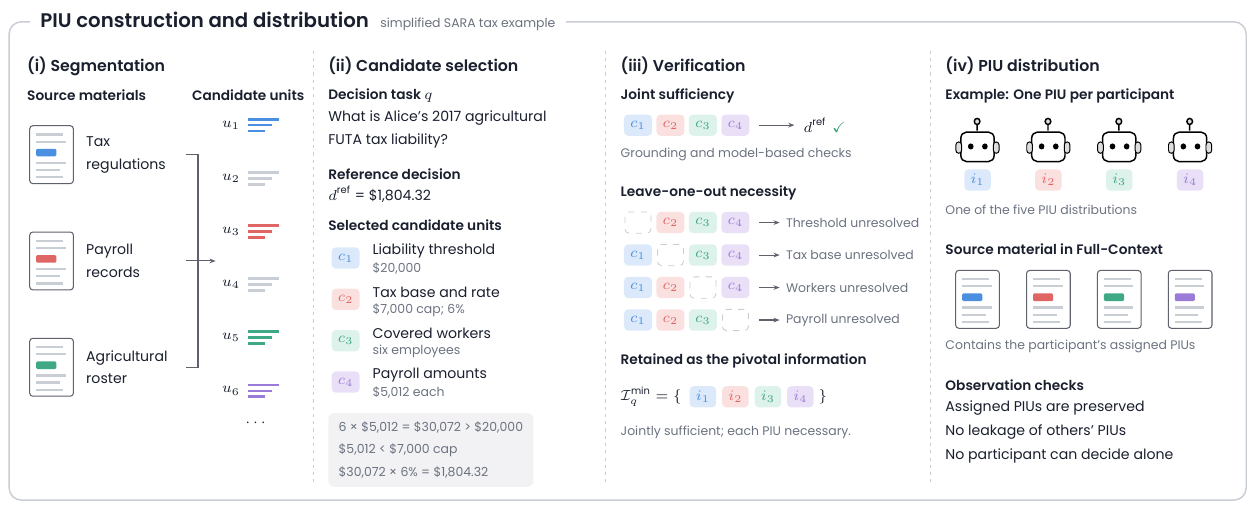}
    \caption{\revised{\textbf{PIU construction and distribution.}
    Candidate evidence units selected from source material are checked for
    joint sufficiency and leave-one-out necessity. The retained units form
    the pivotal information. The simplified SARA example illustrates
    \OneRIUPerParticipant{} (Figure~\ref{fig:piu-distribution-conditions}).
    The decision task and reference decision stay fixed across distributions.}}
    \label{fig:piu-construction-pipeline}
\end{figure*}
\FloatBarrier

\safevspace{}
\paragraph{PIU distribution.}
A PIU distribution specifies who initially has each PIU in
$\mathcal{I}^{\min}_q$: a PIU may be assigned to one or more participants
through their initial local observations or placed in the initial common
observation shared by the team. \revised{The five conditions are
\OneRIUPerParticipant{}, \PairwiseRIUOverlap{} (some PIUs are shared by two
participants), \ExtraParticipantWithoutRIU{}, \OneCommonRIU{} (placed in the
initial common observation), and \AllButOneRIUConcentration{} (one
participant receives all but one PIU). Not every problem instance admits
every condition (Figure~\ref{fig:piu-distribution-conditions}).} For a fixed $q$, the conditions vary
the PIU distribution while keeping the pivotal information and
$d^{\mathrm{ref}}$ unchanged. The initial common observation may also include
\textit{participant descriptions}, which summarize participants' subject
matter or function without revealing their assigned PIUs and can inform
speaker selection.
\par

\safevspace{}
\paragraph{Source-material distribution.}
Rather than receiving PIUs directly, participants' initial observations may
contain portions of the contracts, filings, papers, or source code used to
construct $q$. We divide this source material into units without dividing a
PIU, assign PIU-containing units according to the fixed PIU distribution, and
distribute the remaining units across participants. For every problem instance,
we then verify that no participant can produce a correct final decision from
its initial local observation and the initial common observation.
Appendix~\ref{app:piu-construction-validation} reports an additional re-check
of this requirement on representative instances using a model not used for
construction. Appendix~\ref{app:problem-instance-construction} provides the
formal source terminology and source-specific construction and distribution
procedures.
\par

\safevspace{}
\paragraph{\revised{Example.}}
\revised{Consider a SARA task that asks the team to compute payroll tax from worker
payments and the applicable tax rules. One PIU records George's payment of
\$5,012. In PIU-Only, the participant assigned that PIU receives the payment
fact directly. In Full-Context, the same participant receives a source
excerpt containing the payment and must identify it in the excerpt. Both
teams face the same question with the same distribution of pivotal facts.}

\paragraph{Evaluation.}
A \textit{problem family} groups problem instances constructed from the same
source corpus that require the same general kind of work and use the same
evaluation procedure. For a problem family with a binary success criterion,
\textit{episode success rate} is the proportion of episodes whose final
decisions succeed. Depending on the problem family, final decisions are
compared with reference decisions or assessed by an LLM judge using
problem-specific criteria. Software-engineering patches are checked by an
official repository verifier.
\par

\safevspace{}
\safevspace{}
\section{Evaluation with \ours{}}
\label{sec:evaluation}
\safevspace{}
\safesmallvspace{}
We evaluate LLMs on problem instances constructed from legal, finance, medical,
and software-engineering source corpora under three evaluation conditions:
(1) \textit{Single-Agent All-PIU}, in which one model receives all PIUs and makes
the final decision; (2) \textit{Multi-Agent PIU-Only}, in which the PIUs are
distributed directly across participants; and (3) \textit{Multi-Agent
Full-Context}, in which the same participants instead receive source material
containing their assigned PIUs. For each problem instance and LLM, the three
conditions share the same PIUs and evaluator; the two multi-agent conditions
also share the same PIU distribution, communication structure, and resource
budgets.

\begin{figure}[t]
    \centering
    \includegraphics[width=\textwidth]{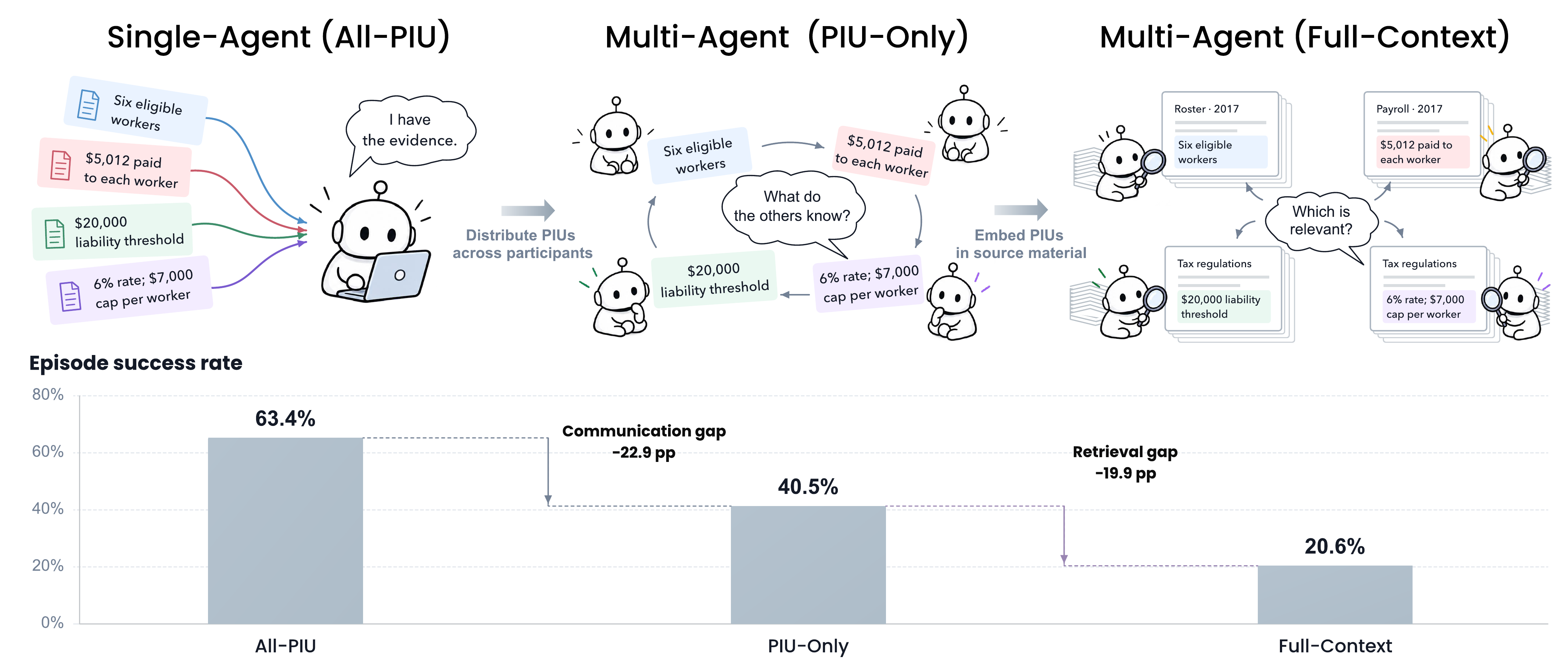}
    \caption{\textbf{Three evaluation conditions and aggregate performance gaps.}
    Single-Agent All-PIU provides all pivotal information to one model;
    Multi-Agent PIU-Only distributes those PIUs across participants; and
    Multi-Agent Full-Context embeds the same PIUs in participants' source
    materials. The communication gap compares the first two conditions, and
    the retrieval gap compares the two multi-agent conditions.}
    \label{fig:three-mode-comparison}
\end{figure}

Figure~\ref{fig:three-mode-comparison} summarizes how the information available
to the decision maker changes across these three conditions and shows the
corresponding aggregate communication and retrieval gaps.
\revised{Appendix~\ref{app:gap-scope} explains what each gap measures.
Figure~\ref{fig:experiment-roadmap} maps the studies, and
Appendix~\ref{app:experimental-settings-overview} summarizes the varied and
fixed settings.}

\safevspace{}
\subsection{Problem Instances}
\label{sec:problem-instances}
\safevspace{}

Table~\ref{tab:introduction-source-corpora} \revised{summarizes the source corpora, decisions, and constructed and
evaluated instance counts.} Before evaluation, we perform repeated
checks in which GPT-5.5 receives all PIUs under Single-Agent All-PIU. It
succeeds under the problem-specific evaluator at least once on every evaluated
problem instance, verifying that a model given all pivotal information can
solve the instance before the PIUs are distributed
(Section~\ref{sec:real-source-problems}).
Appendix~\ref{app:problem-instance-construction} reports additional
construction details and problem-instance statistics for each source corpus.

\safesmallvspace{}

\subsection{Cross-Domain Communication and Retrieval Gaps}
\label{sec:cross-domain-three-evaluation-conditions-results}
\safevspace{}

\begin{figure*}[!h]
    \centering
    \includegraphics[width=\textwidth]{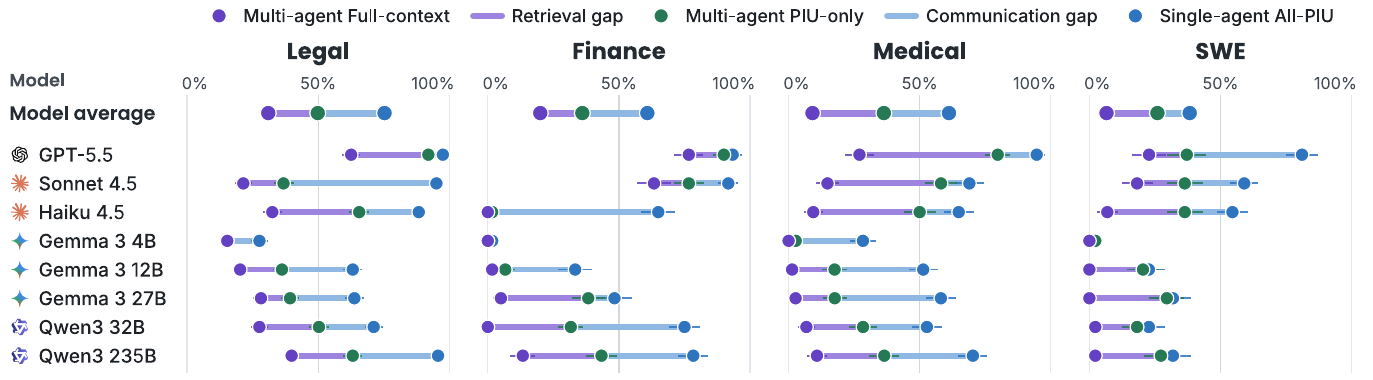}
    \safevspace{}
    \safevspace{}
    \safevspace{}
    \safevspace{}
    
    \caption{\revised{\textbf{Cross-domain episode success rates.}
    Across eight models and 12 corpora, success drops by 22.9 percentage points
    in the communication comparison and by a further 19.9 points in the
    retrieval comparison.} The model-average row weights models equally.
    Error bars indicate standard errors. Appendix~\ref{app:cross-domain-three-evaluation-conditions-results}
    reports source-corpus results, evaluation coverage, and failure handling.}
    \label{fig:cross-domain-three-evaluation-conditions-heatmap}
    \safevspace{}
    \safevspace{}
    \safevspace{}

\end{figure*}

For this cross-domain comparison, both multi-agent conditions use
\OneRIUPerParticipant{} (Figure~\ref{fig:piu-distribution-conditions}) and
\CoordinatorSelectedSpeaker{} \revised{(Figure~\ref{fig:communication-structure-taxonomy})}. The \textit{communication gap} is the decrease in
episode success rate from Single-Agent All-PIU to Multi-Agent PIU-Only, and the
\textit{retrieval gap} is the additional decrease from Multi-Agent PIU-Only to
Multi-Agent Full-Context. \revised{With the model and problem instance fixed, each gap measures the
aggregate change across its two evaluation conditions. Appendix~\ref{app:gap-scope}
describes the changes in team interaction, final decision-making, and source
presentation that enter these contrasts.} The retrieval contrast
corresponds to a departure
from \hyperref[assump:direct-type-report]{A2}, and the communication contrast
corresponds to a departure from
\hyperref[assump:unconstrained-communication]{A3}.
Figure~\ref{fig:cross-domain-three-evaluation-conditions-heatmap} reports the
episode success rates used to compute these gaps.
Appendix~\ref{app:cross-domain-three-evaluation-conditions-results}
additionally reports \textit{Single-Agent Full-Context} \revised{as a supplementary condition}, in which one model
receives all participants' source material combined in participant order,
removing team communication while retaining retrieval from source material.
\safevspace{}

\revised{\textbf{Strong individual problem-solving does not
guarantee effective teamwork:}} across all four domains, we observe
substantial communication and retrieval gaps, although their magnitudes vary
across domains and models.
Single-Agent All-PIU reflects individual reasoning and decision-making
capability when all pivotal information is directly available, but this
individual performance \textbf{does not by itself predict} the model's ability
to communicate and integrate PIUs distributed across participants or to retrieve them from source material for team
communication and integration. 

\begingroup
\safevspace{}
\safesmallvspace{}

\subsection{Mechanisms Linking Communication Structure to Episode Success}
\label{sec:selected-mechanism-results}
\safevspace{}

Communication structure may affect episode success rate by changing the
information supplied to the final decision maker. We examine two mechanisms
through which access to earlier participant messages can change this input:
\textit{correct synthesis} and \textit{error uptake}. Here, correct synthesis refers to an
auditor-labeled, reference-supported inference that combines information. The
intended criterion requires at least two distinct PIUs, but the current labels
also include some inferences combining one PIU with information from the
publicly visible decision options
(Appendix~\ref{app:selected-mechanism-synthesis}). Error uptake occurs
when a later participant uses an incorrect or unsupported claim from an earlier
visible message; rejecting or correcting that claim does not count.

\safesmallvspace{}
\revised{Section~\ref{sec:cross-domain-three-evaluation-conditions-results}
fixed the communication structure. Here we vary it under Multi-Agent PIU-Only
with \OneRIUPerParticipant{}.} In this setting,
\textbf{\RoundRobin{} and \CurrentSpeakerSelection{} achieve higher episode
success rates than \SimultaneousResponses{}} when the same Qwen3 model is used
for the participants and the final decision maker. When only the final decision
maker is replaced with GPT-5.5, however, \textbf{the ordering reverses:
\SimultaneousResponses{} achieves the highest rate}. We use frozen participant
trajectories to test how adding correct synthesis and removing error uptake
affect episode success. A separate GPT-5.5
auditor labels sentence-level components. Participants are not
rerun; we reconstruct the transcript input and regenerate only the final
decision. Appendix~\ref{app:selected-mechanism-design} reports the complete
design and sample sizes. To isolate these mechanisms, we add correct synthesis
while holding the PIUs supplied to the final decision maker fixed and
excluding sentences that directly state a final choice or answer; separately,
we remove only later sentences that use an earlier error while retaining the original error
(Table~\ref{tab:selected-mechanism-results}).

\noindent
\begin{minipage}[t]{0.46 \textwidth}
    \vspace{0pt}
\safevspace{}
\safesmallvspace{}

    \captionsetup{font=normalsize,justification=raggedright,
      singlelinecheck=false}
    \captionof{table}{\textbf{Transcript interventions.} Effects on episode
    success rate (pp). \revised{Asterisks mark effects whose paired 95\% bootstrap confidence
    intervals exclude zero.} ``Same'' uses the participant model as the final
    decision maker.}
\safevspace{}
\safevspace{}

    \label{tab:selected-mechanism-results}
    \vspace{0.2\baselineskip}
    \centering
    \fontsize{8}{9}\selectfont
    \setlength{\tabcolsep}{2.2pt}
    \renewcommand{\arraystretch}{1.12}
    \begin{tabular}{@{}cccc@{}}
    \toprule
    \multirow{2}{*}{\makecell[c]{Participant\\model}} &
    \multirow{2}{*}{\makecell[c]{Final decision\\maker}} &
    \multicolumn{2}{c}{Intervention effect (pp)} \\
    \cmidrule(lr){3-4}
    & & \makecell[c]{Add correct\\synthesis} &
    \makecell[c]{Remove error\\uptake} \\
    \midrule
    \multirow{2}{*}{Qwen3-32B} & Same &
    $+14.6^{*}$ &
    $+2.5$ \\
    & GPT-5.5 &
    $+5.7^{*}$ &
    $+1.3$ \\
    \midrule
    \multirow{2}{*}{Qwen3-235B} & Same &
    $+17.7^{*}$ &
    $+5.7^{*}$ \\
    & GPT-5.5 &
    $+8.2^{*}$ &
    $+7.6^{*}$ \\
    \bottomrule
    \end{tabular}
\end{minipage}\hfill
\begin{minipage}[t]{0.52\textwidth}
    \vspace{0pt}
    \safevspace{}
    \safesmallvspace{}

    \RaggedRight
    \hyphenpenalty=10000
    \exhyphenpenalty=10000
    \revised{\textbf{Access to earlier participant messages can improve decisions through
    correct synthesis or impair them through error uptake. The effects of both
    mechanisms depend in part on which model makes the final decision.}}
    Adding correct synthesis increases episode success rate in all four
    conditions, but replacing only the final decision maker with GPT-5.5
    reduces its effect. Removing error uptake has a statistically significant
    effect only for Qwen3-235B participants, and its effect is smaller than the
    correct-synthesis effect in every condition. Appendix~\ref{app:selected-mechanism-results} describes how correct synthesis
and error uptake are identified and how each intervention is constructed.
\end{minipage}

\safesmallvspace{}

\endgroup

\safevspace{}
\safesmallvspace{}
\subsection{PIU-Distribution and Communication-Structure Interaction}
\label{sec:cross-domain-riu-distribution-communication-structure-results}
\safevspace{}
\safesmallvspace{}

We evaluate this interaction across the 11 problem families for which we
observe multiple PIU-distribution conditions and
communication structures, testing whether the difference in episode success
rate between communication structures changes across PIU-distribution
conditions. DeepSWE is excluded because long-horizon repository execution and
functional verification make this comparison computationally costly.
Appendix~\ref{app:riu-distribution-communication-theory} relates this comparison
to team decision theory, communication complexity, and the hidden-profile
paradigm in group decision-making.

\safesmallvspace{}
{We estimate separate logistic models for Multi-Agent PIU-Only
and Multi-Agent Full-Context. For episode $e$, let $f_e$, $m_e$, $q_e$,
$\ell_e$, and $c_e$ denote the problem family, model, problem instance,
PIU-distribution condition, and communication structure. We specify
\safevspace{}

\safesmallvspace{}
\safevspace{}
\[
y_e \sim \operatorname{Bernoulli}(p_e), \qquad
\operatorname{logit}(p_e)
= \beta_0 + \gamma_{f_e} + \alpha_{m_e} + \pi_{\ell_e}
  + \lambda_{c_e} + \omega_{\ell_e,c_e},
\]

\safevspace{}
\safesmallvspace{}
where $y_e$ indicates episode success, $p_e$ is the probability of episode
success, $\gamma_{f_e}$ and $\alpha_{m_e}$ are problem-family and model fixed
effects, $\pi_{\ell_e}$ and $\lambda_{c_e}$ are PIU-distribution and
communication-structure terms, and $\omega_{\ell_e,c_e}$ is their interaction.
\OneRIUPerParticipant{} and \CoordinatorSelectedSpeaker{} are the reference
categories. We weight each episode by the inverse number of PIU-distribution
conditions available for its problem instance $q_e$, so problem instances with
more conditions do not receive greater aggregate weight, and cluster standard
errors by problem instance.}

\safesmallvspace{}
Across the 11 included problem families and eight models, we
reject the joint null that all PIU-distribution--communication-structure
interaction terms are zero in both Multi-Agent PIU-Only ($p<.001$) and
Multi-Agent Full-Context ($p<.001$).
\revised{\textbf{Thus, which communication structure performs
better depends on the PIU distribution.}} Appendix
Table~\ref{tab:cross-domain-riu-communication-interactions} defines and reports
the complete probability-scale interaction estimates, and
Appendix~\ref{app:problem-instance-riu-distribution-communication-structure-results}
reports the analysis with problem-instance fixed effects.

\revised{The tested PIU distributions can differ in team size by construction.
For example, across the six legal corpora in Multi-Agent PIU-Only,
\CurrentSpeakerSelection{} leads \SimultaneousResponses{} by 3.9 percentage
points under \OneCommonRIU{} and trails it by 9.8 points under
\ExtraParticipantWithoutRIU{}. The 13.8-point interaction compares
configurations that vary both information availability and team size.
The matched Finance and SARA interventions below test specific mechanisms
within their respective settings, leaving their contributions to the
cross-domain interaction unquantified.}
\par

\safevspace{}
\safesmallvspace{}
\subsubsection{Example 1: Narrow coordinator queries limit the PIUs communicated by one participant.}
\label{sec:piu-interaction-finance-example}
Across 59 finance problem instances, pooling Qwen3-32B and Qwen3-235B-A22B,
\textbf{the advantage of \RoundRobin{} over \CoordinatorSelectedSpeaker{}
is larger under \AllButOneRIUConcentration{}}: 31.4 percentage points,
compared with 17.8 under \OneRIUPerParticipant{}. The resulting $+13.6$-point
interaction is statistically significant. Yet 99.2\% of required participants
speak in the concentrated condition. Under \CoordinatorSelectedSpeaker{},
the selected participant receives a specific query. Its public description
summarizes its subject matter or function without listing its assigned PIU
categories, so the query may request only part of its evidence. Under
\RoundRobin{}, participants respond to the whole problem without a
coordinator query. We therefore test whether broadening or removing the
query reduces the communication-structure interaction when speaker selection
is fixed.

We fix speaker selection, order, response count, private source material,
and budgets. After selection, we append \textit{public PIU category labels}
that identify the topics of the selected participant's assigned PIUs, with
an instruction to report evidence for each category. The existing public
participant description remains unchanged; the labels disclose no private
facts, values, or answers. Messages and final decisions are regenerated.

We also replace the query with ``It is your turn to contribute,'' with and
without the additions. Holding the original \RoundRobin{} outcomes fixed,
we recalculate the communication-structure interaction for each coordinator
query condition. The original-query replay gives $+15.5$ points;
Table~\ref{tab:piu-interaction-finance-main} reports this $2\times2$ comparison.
\par\noindent
\begin{minipage}[t]{0.46\textwidth}
\vspace{0pt}
\captionsetup{font=normalsize,labelfont={color=black},textfont={color=black},
  justification=raggedright,singlelinecheck=false}
\captionof{table}{\textbf{Communication-structure interactions (pp).}
Each entry is the \RoundRobin{} advantage over \CoordinatorSelectedSpeaker{}
under \AllButOneRIUConcentration{} minus that under \OneRIUPerParticipant{}.
The Qwen3 \RoundRobin{} outcomes remain fixed. The right column appends
public PIU category labels and an instruction to report evidence for each
category; the left appends neither. \revised{Asterisks mark statistically significant interactions.}}
\label{tab:piu-interaction-finance-main}
\vspace{0.2\baselineskip}
\centering
\fontsize{8}{9}\selectfont
\setlength{\tabcolsep}{3pt}
\renewcommand{\arraystretch}{1.12}
\begin{tabular}{@{}lrr@{}}
\toprule
Coordinator query & \makecell{Without\\labels or\\instruction} & \makecell{With\\labels and\\instruction} \\
\midrule
Original coordinator query & $+15.5^{*}$ & $+5.9$ \\
\arrayrulecolor{black!18}\specialrule{0.3pt}{0.4pt}{0.4pt}\arrayrulecolor{black}
Coordinator query removed & $-5.4$ & $-9.3$ \\
\bottomrule
\end{tabular}
\end{minipage}\hfill
\begin{minipage}[t]{0.52\textwidth}
\vspace{0pt}
\RaggedRight
\revised{\textbf{Narrow coordinator queries contribute to the larger loss from PIU
concentration under \CoordinatorSelectedSpeaker{}.}}
Adding the labels and instruction reduces the interaction from $+15.5$ to
$+5.9$ points. Query removal changes it to $-5.4$ points without additions
and $-9.3$ with them. All three changes from the original-query replay are
statistically significant; none of the resulting interactions is significant.
Both interventions selectively increase substantive PIU communication.
These results support a role for query content, without establishing a zero
residual interaction or separate effects of the labels and instruction.
Appendix~\ref{app:piu-interaction-finance-design}--\ref{app:piu-interaction-finance-results}
reports the design, success rates, and inference.
\end{minipage}
\par
\safesmallvspace{}

\revised{Appendix~\ref{app:matched-episodes-ball} illustrates the narrow-query
mechanism in four matched episodes from one FinLongDocQA problem.}

\begingroup
\safevspace{}
\safesmallvspace{}
\subsubsection{Example 2: Messages from participants assigned no PIU impair the final decision.}
\label{sec:piu-interaction-sara-example}
In the cross-domain Multi-Agent Full-Context regression,
\AllButOneRIUConcentration{} shifts the comparison between
\SimultaneousResponses{} and \CoordinatorSelectedSpeaker{} by 4.0 percentage
points toward the latter relative to \OneRIUPerParticipant{}. This
interaction is statistically significant.
Source material could impair participants assigned
PIUs or cause participants assigned no PIU to produce harmful messages.
We examine these possibilities across 44 SARA problem instances, pooling
Qwen3-32B, Qwen3-235B-A22B, and GPT-5.5. We first compare Multi-Agent
PIU-Only with a condition that adds the original private source material.
In both conditions, participants receive their assigned PIUs explicitly as
facts relevant to the decision; in Multi-Agent Full-Context, they must
instead retrieve those facts from the source material. This comparison tests
whether adding source material changes the interaction even when the
relevant facts are already supplied directly. In Multi-Agent PIU-Only,
the advantage of \SimultaneousResponses{} over
\CoordinatorSelectedSpeaker{} is 10.6 percentage points larger under
\AllButOneRIUConcentration{} than under \OneRIUPerParticipant{}.
Adding source material changes this interaction to $-5.1$ points, a
statistically significant change of $-15.7$ points.

To locate the loss, we next independently vary source-material access for
participants assigned PIUs and those assigned no PIU, holding exact PIUs and
public descriptions fixed. Under \AllButOneRIUConcentration{} and
\SimultaneousResponses{}, source material only for participants assigned no
PIU reduces episode success from 69.7\% to 22.5\%; source material only for participants
assigned PIUs reduces it to 54.8\%. The larger loss thus occurs without
changing the latter group's inputs. However, source placement changes both
participant processing and messages, so it does not isolate their effect
on the final decision.

If messages from participants assigned no PIU cause the loss, removing them
should selectively improve success when their authors received source
material. We therefore freeze all participant messages and regenerate only
the final decision, comparing the complete transcript with one retaining only messages
from participants assigned PIUs. Both transcript conditions regenerate the final decision, so the
complete-transcript success rates differ slightly from those above.
Table~\ref{tab:piu-interaction-sara-main} reports this test.
\noindent
\begin{minipage}[t]{0.46\textwidth}
\vspace{0pt}
\captionsetup{font=normalsize,labelfont={color=black},textfont={color=black},
  justification=raggedright,singlelinecheck=false}
\captionof{table}{\textbf{Fixed-message intervention.} Episode success rates
under \AllButOneRIUConcentration{} and \SimultaneousResponses{}.
Rows vary private source material for participants assigned no PIU;
participants assigned PIUs receive no source material in either row.
Effects are the change from removing messages, in pp. The interaction is
the removal effect with source material minus that without it.
\revised{Asterisks mark statistically significant effects.}}
\label{tab:piu-interaction-sara-main}
\vspace{0.2\baselineskip}
\centering
\fontsize{8}{9}\selectfont
\setlength{\tabcolsep}{2.2pt}
\renewcommand{\arraystretch}{1.12}
\begin{tabular}{@{}lrrr@{}}
\toprule
\makecell[l]{Source\\material} & \makecell{Complete\\transcript} &
\makecell{Messages from\\participants\\assigned no PIU\\removed} & \makecell{Effect\\(pp)} \\
\midrule
Absent & 69.9\% & 68.9\% & $-1.0$ \\
Present & 21.2\% & 68.9\% & $+47.7^{*}$ \\
\midrule
\multicolumn{3}{l}{Interaction} & $+48.7^{*}$ \\
\bottomrule
\end{tabular}
\end{minipage}\hfill
\begin{minipage}[t]{0.52\textwidth}
\vspace{0pt}
\RaggedRight
\revised{\textbf{Messages from participants assigned no PIU can impair the final decision,
even when the evidence communicated by participants assigned PIUs is retained.}}
Removing those messages improves success when their authors receive source
material, with no statistically significant improvement when they do not.
The interaction between source-material access and message removal is
statistically significant. The result identifies a harmful contribution at
the final-decision stage;
it does not distinguish transcript length from distracting or contradictory
content. Appendix~\ref{app:piu-interaction-sara-design}--\ref{app:piu-interaction-sara-results}
reports the complete design, uncertainty estimates, and scope.
\end{minipage}
\par
\safesmallvspace{}

\par
\endgroup

\safevspace{}
\safesmallvspace{}
\subsection{Token-Budget Effects}
\label{sec:legal-participant-resource-budget-results}
\safevspace{}

Under Multi-Agent Full-Context with
\OneRIUPerParticipant{} and \CoordinatorSelectedSpeaker{}, we vary
private-note, message, and local-transcript token budgets, which respectively
limit unshared notes, messages, and newly available transcript text.
\revised{\textbf{Among them, the message budget has the strongest
relationship with episode success rate:}} for GPT-5.5 on 24 CUAD and 29
Multi-LexSum instances, increasing it from 32 to 128 tokens raises the rate
from 55.3\% to 78.6\%, averaged over private-note budgets, with no further gain
at 256 tokens; private-note differences are smaller, and larger
local-transcript budgets show no consistent improvement on Multi-LexSum.
Appendix~\ref{app:participant-resource-budget-details} reports complete results
and diagnostics.

\FloatBarrier

\safevspace{}
\safevspace{}

\section{Policy Training with \ours{}: Emergent Coordination}
\label{sec:policy-training}
\label{sec:policy-training-emergent-coordination}
\safevspace{}
\safevspace{}

We demonstrate policy training with \ours{} under two
communication structures: (1) \CoordinatorSelectedSpeaker{}, where we train
participant and coordinator policies and the coordinator selects and queries
participants and makes the final decision; and (2)
\CurrentSpeakerSelection{}, where we train a shared participant policy and each
current speaker either selects the next participant or ends communication. For
this demonstration, the participant that ends communication makes the final
decision, so no separate final decision maker is used.
\revised{The primary policies} are initialized from Qwen2.5-7B-Instruct~\citep{yang2024qwen2} and
trained using Group Relative Policy Optimization
(GRPO)~\citep{shao2024deepseekmath}. In both demonstrations, the LLM-judged
score assigned by GPT-5.5 serves as training reward. We
call systematic changes in participation, speaker selection, and communication
\emph{emergent coordination} when these behaviors are not specified as separate
rewards or training targets.

\safevspace{}
\safesmallvspace{}
\subsection{\CoordinatorSelectedSpeaker{}:
Training Participant and Coordinator}
\label{sec:coordinator-selected-policy-training}

\safevspace{}

Under \CoordinatorSelectedSpeaker{}, a shared participant policy generates
messages from source material, while a
coordinator policy generates queries and the final decision. We use
Multi-Agent Full-Context on CUAD- and MAUD-derived problem
instances~\citep{hendrycks2021cuad,wang2023maud}, converting each decision task
to a free-form written decision format while preserving its PIU distribution
and source material
(Appendix~\ref{app:policy-training-free-form-construction}). Training and
checkpoint selection use only CUAD-derived problem instances; the 26
MAUD-derived problem instances evaluate transfer. We compare
(1) two-stage policy training, which updates the participant and coordinator
policies successively to examine which behaviors change when each policy is
updated separately, and (2) joint two-policy training, which allows them to
co-adapt and serves as the primary end-to-end experiment.
Appendices~\ref{app:policy-training-free-form-evaluation} and
\ref{app:coordinator-controlled-policy-training} provide evaluation,
optimization, and checkpoint-selection details.

\safevspace{}
\safesmallvspace{}
\subsubsection{Two-Stage Policy Training}
\label{sec:emergent-coordination-sequential-training}
\safevspace{}

On the 36 CUAD-derived validation instances, the LLM-judged score increases
from 12.14\% to 16.84\% after the participant-only stage and to 23.33\% after
the coordinator-only continuation. Illustrative examples show that the
participant-only stage changes which PIUs appear in participant messages and,
through the resulting transcript, can change queries generated by the fixed
coordinator policy; the coordinator-only continuation can change queries and
final decisions even when participant messages remain substantively
unchanged. \revised{\textbf{Two-stage policy training can
therefore change participant messages, coordinator queries, and final
decisions, but its CUAD-derived validation gains do not transfer to the 26
held-out MAUD-derived problem instances.}}
Appendix~\ref{app:two-stage-policy-training} provides training details and
illustrative examples.

\safevspace{}
\subsubsection{Joint Two-Policy Training}
\label{sec:emergent-coordination-joint}
\safevspace{}
\safesmallvspace{}
\revised{\textbf{Joint two-policy training produces short
coordinator queries and queries every participant, despite not directly rewarding
either behavior.}} On the 36 CUAD-derived validation problem instances, joint
two-policy training increases the LLM-judged score from 11.97\% to 22.38\% and
the share of problem instances in which the coordinator queries every
participant from 80.6\% to 100\%. Because querying every participant does not
enter the training reward, this increase is an instance of emergent
coordination. Table~\ref{tab:emergent-coordination-query-protocols}
shows distinct coordinator-query behavior: joint two-policy training uses
shorter queries, queries every participant, and never copies the problem
question in the first query, whereas two-stage policy training uses longer
queries and more later queries that reuse terms from preceding participant
replies.
\revised{Figure~\ref{fig:policy-training-validation-cuad} compares the scheduled
CUAD validation scores from three joint-training runs and the single two-stage
run. The query diagnostics refer to the representative joint-training run.}

\begin{table}[h]
\noindent
\safevspace{}
\safevspace{}
\safevspace{}

\begin{minipage}[t]{0.55\textwidth}
    \vspace{0pt}
    \captionsetup{font=normalsize,justification=raggedright,singlelinecheck=false}
    \caption{\textbf{Coordinator-query behavior.}
    CUAD-derived validation (\(n=36\)); percentages except mean words per
    query. Behavior rates use eligible queries; reuse of preceding-reply terms
    is measured lexically.}
\safevspace{}
\safevspace{}

    \label{tab:emergent-coordination-query-protocols}
    \centering
    \fontsize{7.5}{8.5}\selectfont
    \setlength{\tabcolsep}{2pt}
    \renewcommand{\arraystretch}{1.08}
    \begin{tabular}{@{}cccc@{}}
    \toprule
    \makecell[c]{Measurement} & \makecell{Base\\policies} & \makecell{Two-stage\\training} & \makecell{Joint two-policy\\training} \\
    \midrule
    Every participant queried & 80.6 & 66.7 & \textbf{100.0} \\
    Mean words per query & 14.0 & 48.0 & \textbf{12.5} \\
    Question copied in first query & 0.0 & 88.6 & 0.0 \\
    \makecell{Later queries reusing\\preceding-reply terms} & 3.2 & 42.9 & 13.5 \\
    \bottomrule
    \end{tabular}
\end{minipage}\hfill
\begin{minipage}[t]{0.42\textwidth}
    \vspace{0pt}
    \RaggedRight
    \hyphenpenalty=10000
    \exhyphenpenalty=10000
    Joint two-policy training produces short queries, does
    not copy the problem question in the first query, and queries every
    participant on all 36 problem instances. Later queries that reuse terms
    from preceding participant replies remain infrequent, so the learned
    coordinator-query behavior is characterized primarily by shorter queries
    and querying every participant.
\end{minipage}
\safevspace{}
\safesmallvspace{}
\end{table}

On 26 MAUD-derived instances, under joint two-policy training the coordinator
queries every participant and every participant sends a public message in all
26, versus 20 under the base policies, while the aggregate LLM-judged score
changes only from 5.12\% to 5.58\%. Thus, coordination behavior transfers even
when final-decision quality changes little.
Appendix~\ref{app:joint-two-policy-training} provides training details and an
example. \revised{The paired analysis in Appendix~\ref{app:cuad-maud-transfer-analysis}
finds that participation transfers more clearly than evidence transmission
or decision quality.}

\safevspace{}
\safesmallvspace{}

\subsection{\CurrentSpeakerSelection{}:
Training a Shared Participant Policy}
\label{sec:current-speaker-participant-policy-training}
\safevspace{}
\safesmallvspace{}

\begin{figure}[htbp]
    \centering
    \includegraphics[width=\textwidth]{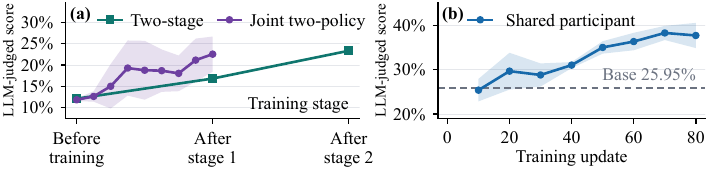}
    {\phantomsubcaption\label{fig:policy-training-validation-cuad}}%
    {\phantomsubcaption\label{fig:finance-current-speaker-validation-trajectory}%
    \label{fig:policy-training-validation-findver}}%
    \caption{\revised{\textbf{Policy-training validation scores.}
    \textbf{(a)} CUAD scores for the single two-stage run and mean $\pm$ SD
    across three joint-training runs. \textbf{(b)} FinDVer scores, mean $\pm$
    SD across three shared-policy runs, with the pretraining baseline.}}
    \label{fig:policy-training-validation-results}
\end{figure}

We train a shared participant policy under \CurrentSpeakerSelection{} in
Multi-Agent PIU-Only with
\OneRIUPerParticipant{}. The shared policy controls participant messages,
next-participant selection, decisions to end communication, and the final
decision. Public participant descriptions are visible to support participant
selection. Training uses 350 FinDVer problem instances and 188 two- or
three-participant FinLongDocQA problem instances. We select the checkpoint
using the 39-instance FinDVer validation
subset and report separate evaluations on the 21-instance FinLongDocQA
validation subset and 40 held-out four-participant FinLongDocQA problem
instances. The held-out test comparison therefore asks whether a policy trained
on two- and three-participant FinLongDocQA problem instances also improves
performance on four-participant instances.
\revised{\textbf{The trained shared participant policy improves
LLM-judged scores on both validation subsets and on held-out four-participant
FinLongDocQA problem instances.}}
\revised{Figure~\ref{fig:finance-current-speaker-validation-trajectory} shows
validation curves for three runs. Table~\ref{tab:finance-current-speaker-policy-training}
reports selected-checkpoint scores. The replay and communication diagnostics
below use the representative run and its original baseline.}
Appendix~\ref{app:coordinator-free-policy-training} provides the data-split, optimization, checkpoint-selection, and diagnostic details.

\begin{table}[htbp]
    \centering
    \caption{\revised{\textbf{Finance validation and test scores at selected checkpoints.}
    LLM-judged scores (\%). The trained column reports mean $\pm$ sample SD
    across three Qwen2.5-7B-Instruct runs. The test uses held-out
    four-participant FinLongDocQA teams.}}
    \label{tab:finance-current-speaker-policy-training}
    \revised{%
    \begin{tabular}{@{}llrrrr@{}}
    \toprule
    Corpus & Split & $n$ & Team size & Base & Trained \\
    \midrule
    FinDVer & Validation & 39 & 2--8 & 25.95 & $38.79\pm3.70$ \\
    FinLongDocQA & Validation & 21 & 2--3 & 54.05 & $67.70\pm3.16$ \\
    FinLongDocQA & Test & 40 & 4 & 25.53 & $40.83\pm8.82$ \\
    \bottomrule
    \end{tabular}}
\end{table}

\begin{table}[H]
\safevspace{}
\safevspace{}
\safevspace{}
\safesmallvspace{}

\noindent
\begin{minipage}[t]{0.38\textwidth}
    \vspace{0pt}
    \captionsetup{font=normalsize,justification=raggedright,singlelinecheck=false}
    \caption{\textbf{Cross-replay.} Mean LLM-judged scores (\%) completed by both trajectories.}
    \label{tab:finance-current-speaker-cross-replay}
    \par\vspace{0.2\baselineskip}
    \centering
    \fontsize{7.5}{8.5}\selectfont
    \setlength{\tabcolsep}{2pt}
    \renewcommand{\arraystretch}{1.08}
    \begin{tabular}{@{}ccc@{}}
    \toprule
    \multirow{2}{*}{Communication trajectory} &
    \multicolumn{2}{c}{Final-decision policy} \\
    \cmidrule(l){2-3}
    & Base & Trained \\
    \midrule
    Base & 50.03 & 49.40 \\
    Trained & 56.55 & 56.78 \\
    \bottomrule
    \end{tabular}
\end{minipage}\hfill
\begin{minipage}[t]{0.60\textwidth}
    \vspace{0pt}
    \captionsetup{font=normalsize,justification=raggedright,singlelinecheck=false}
    \caption{\textbf{Communication diagnostics.} Base and trained participant
    policies; percentages except unique speakers. 
    }
    \label{tab:finance-current-speaker-communication-diagnostics}
    \par\safevspace{}\vspace{0.52pt}\centering
    \fontsize{7.5}{8.5}\selectfont
    \setlength{\tabcolsep}{3pt}
    \renewcommand{\arraystretch}{1.08}
    \begin{tabular}{@{}>{\centering\arraybackslash}m{0.43\linewidth}cccc@{}}
    \toprule
    \multirow{2}{*}{Communication diagnostic} &
    \multicolumn{2}{c}{Validation} &
    \multicolumn{2}{c}{FinLongDocQA test} \\
    \cmidrule(lr){2-3}\cmidrule(l){4-5}
    & Base & Trained & Base & Trained \\
    \midrule
    Every participant speaks & 61.7 & 86.7 & 62.5 & 82.5 \\
    Mean participant coverage & 65.2 & 89.4 & 64.4 & 90.0 \\
    Mean unique speakers & 2.28 & 3.25 & 2.58 & 3.60 \\
    Self-handoff rate & 14.2 & 10.6 & 17.6 & 16.2 \\
    \bottomrule
    \end{tabular}
\end{minipage}
\safevspace{}
\safevspace{}
\safevspace{}
\end{table}

Here, a communication trajectory comprises the communication transcript,
next-participant selections, the decision to end communication,
and the identity and assigned PIU of the participant making the final decision.
The final-decision policy is the base or trained shared participant policy used
only to regenerate the final decision.
Cross-replay holds this trajectory and the final-decision prompt fixed while
varying the final-decision policy. \revised{Across 40 validation instances
completed by both the base and trained policies,}
replacing the base communication trajectory
with the trained trajectory, averaged across the two final-decision policies,
yields a 6.95-point gain. Replacing the base final-decision policy with the
trained policy does not produce a statistically significant improvement.
\revised{\textbf{Thus, within cross-replay, the validation gain
is concentrated in the communication trajectory rather than the final-decision
policy.}}
Training also
increases the proportion of participants who send public messages and the
number of distinct speakers while reducing the rate at which a speaker selects
itself as the next participant on validation and test. Because these
diagnostics do not enter the reward, the changes provide evidence of emergent
coordination in participation and next-participant selection.

\begin{figure}[htbp]
    \centering
    \includegraphics[width=\textwidth]{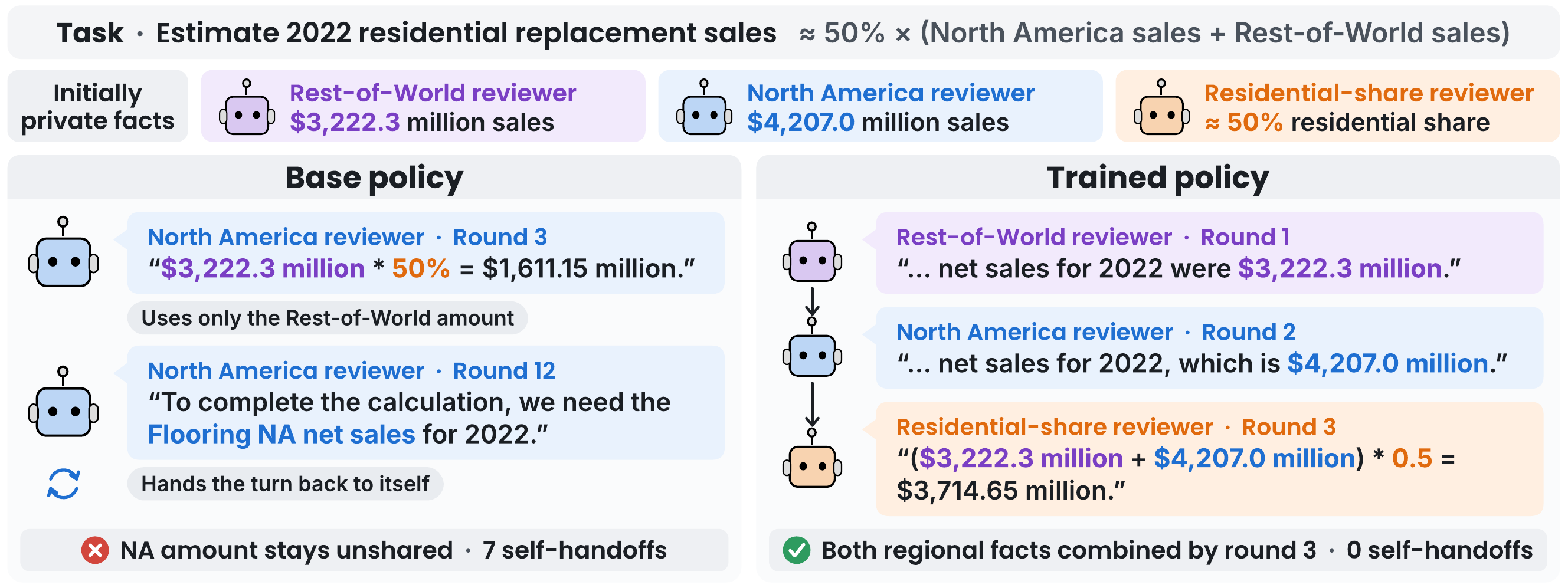}
    \caption{\revised{\textbf{Evidence sharing and coordination before and after training.}
    One FinLongDocQA validation instance with the base and trained
    Qwen2.5-7B-Instruct policies. Both teams receive the same distributed facts
    and use all 15 communication rounds. The trained team combines the
    required quantities in round~3 and requests finalization in round~15.}}
    \label{fig:finance-current-speaker-chat-example}
\end{figure}

\revised{Figure~\ref{fig:finance-current-speaker-chat-example} illustrates
evidence sharing and speaker selection in one FinLongDocQA validation episode.}

\safevspace{}
To test whether additional public messages make more PIUs available for the
final decision, we apply the same GPT-5.5 final-decision policy to every base
and trained communication transcript. Unlike cross-replay, this analysis retains
every trajectory, including valid public messages produced before an episode
that does not reach a final decision. On FinDVer, trained transcripts improve
performance under the same final-decision policy by 12.6 points and increase
the proportion of PIUs appearing in the transcript by 15.9 points. On held-out
FinLongDocQA, however, end-to-end score and final-decision completion improve,
but the changes in performance under the same final-decision policy and in the
proportion of PIUs appearing in the transcript are not statistically
distinguishable from zero.
\revised{\textbf{Together, these results show that broader
participation and higher end-to-end performance do not necessarily coincide
with a larger proportion of PIUs appearing in the communication transcript.}}
Table~\ref{tab:finance-current-speaker-trajectory-analysis} reports these
comparisons, and Appendix~\ref{app:finance-current-speaker-trajectory-analysis}
reports additional diagnostics.

\safesmallvspace{}
Across both fixed communication structures, training changes
unrewarded coordination behaviors as well as final-decision quality. Under
\CoordinatorSelectedSpeaker{}, the participant and coordinator policies change
participant messages, coordinator queries, and final decisions; under
\CurrentSpeakerSelection{}, the shared participant policy changes
participation, next-participant selection, and the communication trajectory.
\revised{Figure~\ref{fig:policy-training-validation-results} summarizes the
repeated Qwen training runs. Appendix~\ref{app:proposal-finance-training-seeds}
reports the per-run finance results. An independently trained
Llama-3.1-8B-Instruct policy improves held-out FinLongDocQA scores by 18.83
percentage points relative to its own baseline
(Appendix~\ref{app:proposal-llama-finance-pilot}).}

\section{Additional Related Work}
\label{app:related-work}

\paragraph{Mechanism design and communication.}
Multistage communication games, communication equilibria, and communication
requirements study communication before a decision
\citep{myerson1986multistage,forges1986approach,segal2007communication}.
Cheap-talk models instead study communication with preference misalignment
\citep{crawford1982strategic}.

\paragraph{Human-AI and group decision making.}
Human-AI decision making motivates evaluating AI systems as team
participants and examining the interaction process rather than only the final
decision \citep{lai2021towards,herath2025design,seeber2020machines}. Hidden-profile
studies show that human teams can fail to reveal uniquely held information, and
collective-intelligence research relates interaction structure to performance
\citep{stasser1985pooling,woolley2010collective}. Work on limited attention,
working memory, and information overload motivates finite resource constraints
\citep{kahneman1973attention,cowan2001magical,eppler2004concept}. These findings
motivate our controlled comparisons, but \ours{} does not evaluate human-AI
teams or treat its token budgets as models of human cognition. Work using LLMs
to simulate social settings has a different objective: plausible social
behavior is the object of study, whereas \ours{} treats the participant model
as a controlled component
\citep{park2022social,park2023generative,hansen2026simulating}.

\paragraph{Multi-agent LLM systems and benchmark contrasts.}
CAMEL, AutoGen, MetaGPT, and ChatDev study collaboration through
prompted agent interaction
\citep{li2023camel,wu2023autogen,hong2024metagpt,qian2024chatdev}. PEER learns
collaborative writing, MDAgents studies adaptive medical collaboration, and
other work studies debate and collaborative post-training
\citep{schick2023peer,kim2024adaptive,du2024multiagentdebate,park2025maporl}.
CooperBench evaluates cooperative coding over repositories
\citep{khatua2026cooperbench}. These systems study agent coordination, but
\ours{} fixes source-grounded PIUs, controls the PIU distribution and whether
participants receive PIUs directly or source material containing them, and
measures whether bounded communication makes the PIUs available for a final
decision. Multi-agent debate commonly gives agents the same
problem; \ours{} gives participants different initial local observations.

MultiAgentBench evaluates coordination under several communication
topologies \citep{zhu2025multiagentbench}. HiddenBench evaluates whether teams
surface and combine directly presented hidden facts
\citep{li2025hiddenprofile,li2026hiddenbench}, while SILO-BENCH studies
role-free collaboration on synthetic distributed-computation problems
\citep{zhang2026silobench}. Appendix~\ref{app:benchmark-comparisons} gives
compact comparisons with both benchmarks.
Collaborative Gym studies a user or user simulator collaborating with one LLM
agent, while our unit of analysis is a team with decision-relevant evidence
divided among participants \citep{shao2026collaborativegym}. Single-agent suites
such as SmartPlay and AgentBench do not require recovery of evidence held by
teammates \citep{wu2024smartplay,liu2024agentbench}.

The original evaluations for the finance, medical, and
software-engineering source corpora in
Table~\ref{tab:introduction-source-corpora} are single-agent or centralized. FinDVer and FinLongDocQA study financial
verification and long-document financial question answering
\citep{zhao2024findver,wang2026finlongdocqa}; MLCR and EvidenceBench provide
long medical-record cases and hypothesis-linked biomedical evidence
\citep{wisedocs2026mlcrbenchmark,wang2025evidencebench}; and SWE Atlas Codebase QnA
studies repository comprehension \citep{raghavendra2026sweatlas}. \ours{} uses the evaluated source corpora
to construct team decision problem instances with controlled PIU
distributions.

\section{Conclusion}
\label{app:deferred-conclusion}
\safevspace{}

We introduce \ours{}, a gym for constructing team decision problem instances
and for evaluating and training participant and coordinator policies under
controlled PIU distributions, communication structures, and token budgets.
Across 8 language models and 12 source corpora, distributing the same PIUs
across participants reduces episode success rate by 22.9 percentage points
relative to providing them to a single model, and requiring participants to
retrieve those PIUs from source material reduces it by a further 19.9 points.
The effectiveness of a communication structure also varies across
PIU-distribution conditions. Team decision-making therefore depends not only
on individual
model capabilities, but also on how participants retrieve, communicate, and
integrate distributed information.

Across two fixed communication structures, policy training improves
final-decision quality and induces emergent coordination in participation,
speaker selection, and communication. With coordinator control, joint
two-policy training produces a compact information-elicitation protocol;
without a coordinator, training improves end-to-end final-decision quality
and completion while increasing participation. Tracing PIUs through these
trajectories shows that increased participation makes more pivotal information
available on FinDVer validation, but not on held-out FinLongDocQA. This
distinction illustrates how \ours{} diagnoses how training changes team
decision making, rather than reporting only whether performance improves.

\revised{The joint two-policy and finance shared-policy training results
include three Qwen runs each, alongside a single Llama finance run.
The CUAD gains under \CoordinatorSelectedSpeaker{} are measured on the
validation instances used for checkpoint selection. On held-out MAUD,
participation transfers more clearly than evidence transmission or final-decision
quality. All of these policies were evaluated under their training communication
structures.} Extensions of \ours{} could relax the shared-payoff
assumption to study strategic information transmission, malicious or unreliable
participants, and teams with partially aligned objectives. They could also
introduce explicit retrieval costs and allow participants to choose which
source material to retrieve. By separating retrieval, communication, and final
decisions while holding the underlying problem fixed, \ours{} provides a
controlled foundation for diagnosing and training team decision-making
capabilities.

\subsection*{Reproducibility Statement}
The supplementary material documents the problem-instance construction
procedures, evaluation settings, prompts and evaluator protocols, statistical
analyses, and policy-training configurations and diagnostic analyses
(Appendices~\ref{app:problem-instance-construction},
\ref{app:communication-structure-implementation},
\ref{app:cross-domain-three-evaluation-conditions-results},
\ref{app:participant-resource-budget-details},
\ref{app:policy-training-free-form-evaluation},
\ref{app:coordinator-controlled-policy-training}, and
\ref{app:coordinator-free-policy-training}). The TeamGym code is available
in the \href{https://github.com/chanwoo-park-official/TeamGym}{GitHub repository}.
The forthcoming supplementary
archive will provide the code, configurations, dataset-access and preparation
instructions, prompts, and generated artifacts needed to reproduce all reported
problem-instance construction, evaluation, analysis, and policy-training results.

\subsection*{\revised{Ethics Statement}}
\revised{%
\ours{} uses publicly available source corpora cited in this paper. Human
annotators evaluated the sufficiency and necessity of source-derived information
units and were offered compensation for their time. No demographic or sensitive
personal information was requested from annotators; results are reported without
identifying them. Appendix~\ref{app:piu-human-validation} describes the protocol and results.
}

\subsection*{Use of AI Assistants}

The authors used OpenAI Codex for code development and manuscript editing.
For the legal, finance, medical, and software-engineering environment, language
models were also used within the research methodology for problem-instance
construction and semantic evaluation, as described in the corresponding
construction and evaluation sections. These uses include generating reasoning
during problem-instance construction when it was not provided, proposing and
checking problem annotations, and judging candidate final decisions against
reference decisions. The authors designed these procedures, reviewed their
outputs, and developed and approved all scientific claims in the final paper.

\section*{Acknowledgements}
CP gratefully acknowledges support from the Amazon AI Fellowship (2025--2027), including GPU resources for small-scale experiments and APIs, and scholarship funding from the Korea Foundation for Advanced Studies. CP also gratefully acknowledges computing resources provided by MIT’s Office of Research Computing and Data.

\clearpage
\appendix
\section{\revised{Experiment Roadmap, Settings, and Communication-Trajectory Results}}
\label{app:deferred-communication-trajectory-table}

\begin{table}[H]
\centering
\captionsetup{font=normalsize}
\caption{\textbf{Communication-trajectory analysis.} Arrows compare the base
and trained policies \revised{in the representative finance run}. All entries are percentages. ``Same final-decision
policy'' is the LLM-judged score obtained by applying one common GPT-5.5
final-decision policy to every communication transcript.}
\label{tab:finance-current-speaker-trajectory-analysis}
\fontsize{7.5}{8.5}\selectfont
\setlength{\tabcolsep}{2.8pt}
\renewcommand{\arraystretch}{1.08}
\begin{tabular}{@{}lrrrrr@{}}
\toprule
& \makecell{LLM-judged\\score}
& \makecell{Final-decision\\completion}
& \makecell{Same final-\\decision policy}
& \makecell{PIUs appearing in the\\communication transcript}
& \makecell{Exactly repeated\\public messages} \\
\midrule
FinDVer validation
& 21.85 $\rightarrow$ 41.08
& 51.3 $\rightarrow$ 84.6
& 34.10 $\rightarrow$ 46.72
& 38.0 $\rightarrow$ 53.9
& 18.9 $\rightarrow$ 50.9 \\
FinLongDocQA test
& 33.50 $\rightarrow$ 41.13
& 65.0 $\rightarrow$ 92.5
& 49.60 $\rightarrow$ 44.82
& 63.1 $\rightarrow$ 63.1
& 23.1 $\rightarrow$ 48.6 \\
\bottomrule
\end{tabular}
\end{table}

\clearpage
\begin{figure}[p]
    \centering
    \includegraphics[width=\textwidth,height=0.80\textheight,keepaspectratio]{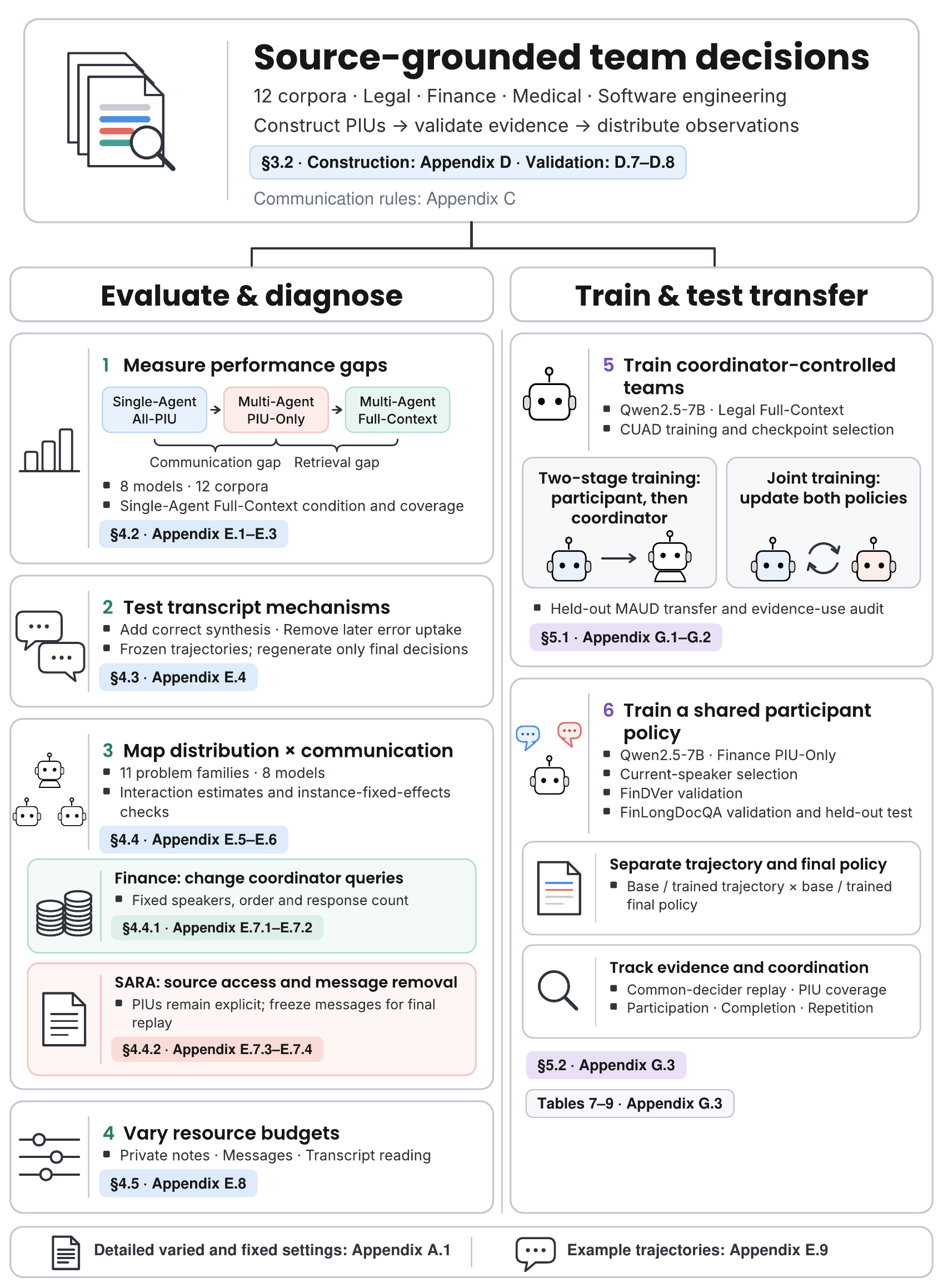}
    \caption{\revised{\textbf{Experiment roadmap.}
    The branches group the evaluation and policy-training studies.
    Appendix~\ref{app:experimental-settings-overview} describes what varies
    and what stays fixed in each comparison.}}
    \label{fig:experiment-roadmap}
\end{figure}
\clearpage

\subsection{\revised{Experimental Settings Overview}}
\label{app:experimental-settings-overview}

\revised{Table~\ref{tab:experimental-settings-overview} maps the varied and fixed
settings in each comparison. Fixed settings apply within a row; rows may use
different settings. Cited sections provide the details.}

\begingroup
\fontsize{9.5}{10.5}\selectfont
\setlength{\tabcolsep}{4pt}
\renewcommand{\arraystretch}{1.10}
\setlength{\LTpre}{3pt}
\setlength{\LTpost}{3pt}
\setlength{\aboverulesep}{1.1pt}
\setlength{\belowrulesep}{1.1pt}
\setlength{\lightrulewidth}{0.3pt}
\hyphenpenalty=10000
\exhyphenpenalty=10000
\arrayrulecolor{black!20}
\revised{
\begin{longtable}{@{}
  >{\centering\arraybackslash}m{0.24\textwidth}
  >{\RaggedRight\arraybackslash}m{0.32\textwidth}
  >{\RaggedRight\arraybackslash}m{\dimexpr0.44\textwidth-4\tabcolsep\relax}
  @{}}
\caption{\revised{\textbf{Experimental settings.} Factors varied and settings
held fixed within each comparison.}}\label{tab:experimental-settings-overview}\\
\toprule
\textbf{Comparison} &
\multicolumn{1}{c}{\textbf{What varies}} &
\multicolumn{1}{c}{\textbf{What stays fixed}}\\
\midrule
\endfirsthead
\multicolumn{3}{c}{\textbf{Table~\thetable\ continued}}\\
\toprule
\textbf{Comparison} &
\multicolumn{1}{c}{\textbf{What varies}} &
\multicolumn{1}{c}{\textbf{What stays fixed}}\\
\midrule
\endhead
\midrule
\multicolumn{3}{c}{\textit{Continued on the next page}}\\
\endfoot
\bottomrule
\endlastfoot

\rowcolor{black!5}
\multicolumn{3}{c}{\textbf{Evaluation comparisons}}\\*
\addlinespace[4pt]

\textbf{Communication gap}\par
\S\,\ref{sec:cross-domain-three-evaluation-conditions-results}\par
8 models, 12 corpora
&
Single-Agent All-PIU versus\newline
Multi-Agent PIU-Only\newline
Changed interaction and final-decision role
&
Instance, PIUs, model, evaluator\newline
Team setting uses\newline
\OneRIUPerParticipant{}\newline
\CoordinatorSelectedSpeaker{}
\\ \midrule

\textbf{Retrieval gap}\par
\S\,\ref{sec:cross-domain-three-evaluation-conditions-results}
&
Multi-Agent PIU-Only versus\newline
Multi-Agent Full-Context\newline
Source presentation\newline
Context load
&
Instance, PIUs, model, evaluator\newline
Team, nominal budgets\newline
\OneRIUPerParticipant{}\newline
\CoordinatorSelectedSpeaker{}
\\ \midrule

\textbf{Single-Agent Full-Context}\par
Supplementary condition\par
Appendix~\ref{app:cross-domain-single-full-context}
&
One model receives all participants'\newline
source material\newline
No team communication
&
Underlying problem, source material,\newline
PIUs, model, evaluator\newline
Evaluated inputs only\newline
Coverage and exclusions in the cited appendix
\\ \midrule

\textbf{Transcript interventions}\par
\S\,\ref{sec:selected-mechanism-results}\par
79 instances
&
Correct synthesis added\newline
Later error uptake removed\newline
Final decision by the participant's Qwen3 model or GPT-5.5
&
Multi-Agent PIU-Only\newline
\OneRIUPerParticipant{}\newline
Frozen participant trajectories\newline
\RoundRobin{} and \CurrentSpeakerSelection{}\newline
Regenerate only the final decision\textsuperscript{a}
\\ \midrule

\textbf{PIU distribution $\times$ structure}\par
\S\,\ref{sec:cross-domain-riu-distribution-communication-structure-results}\par
11 families, 8 models
&
Available PIU distributions\newline
$\times$ communication structures
&
Underlying problem, PIUs, evaluator\newline
Separate PIU-Only and Full-Context regressions\newline
Family and model effects included\newline
DeepSWE excluded
\\ \midrule

\textbf{Finance query interventions}\par
\S\,\ref{sec:piu-interaction-finance-example}\par
59 finance instances
&
Query retained / removed\newline
$\times$ category labels and instruction
present / absent\textsuperscript{b}
&
Multi-Agent Full-Context\newline
\CoordinatorSelectedSpeaker{}\newline
Model, source material, budgets\newline
Recorded speakers, order and count\newline
Original \RoundRobin{} outcomes
\\ \midrule

\textbf{SARA source access}\par
\S\,\ref{sec:piu-interaction-sara-example}\par
44 SARA instances
&
PIUs alone versus\newline
PIUs plus source material\newline
Targeted source-access controls\newline
Participants with / without PIUs\textsuperscript{c}
&
Within each distribution and structure\newline
Assigned PIUs, descriptions,\newline
model, evaluator, nominal caps\newline
PIUs are always supplied explicitly
\\ \midrule

\textbf{SARA message removal}\par
\S\,\ref{sec:piu-interaction-sara-example}
&
Complete transcript versus\newline
Messages from PIU holders only\newline
Crossed with nonholder source access
&
\AllButOneRIUConcentration{}\newline
\SimultaneousResponses{}\newline
Within each source-access condition\newline
Frozen messages and assignments\newline
Regenerate only the final decision\textsuperscript{c}
\\ \midrule

\textbf{Private-note and message budgets}\par
Appendix~\ref{app:legal-private-note-message-budget-details}
&
Private notes 64 / 128 / 256\newline
$\times$ messages 32 / 64 / 128 / 256 tokens
&
Multi-Agent Full-Context\newline
\OneRIUPerParticipant{}\newline
\CoordinatorSelectedSpeaker{}\newline
Model, $N$ messages, coordinator policy,\newline
final cap, 40,000-token private context\newline
Unrestricted transcript reading\textsuperscript{d}
\\ \midrule

\textbf{Local-transcript budgets}\par
Appendix~\ref{app:legal-local-transcript-budget-details}\par
29 Multi-LexSum instances
&
Participant reading cap\newline
FIFO versus random window\newline
Three random-window seeds
&
GPT-5.5, Multi-Agent Full-Context\newline
\OneRIUPerParticipant{}\newline
\CoordinatorSelectedSpeaker{}\newline
Private notes 256, messages 128 tokens\newline
Final cap, complete coordinator transcript
\\ \midrule

\rowcolor{black!5}
\multicolumn{3}{c}{\textbf{Policy training and supporting analyses}}\\*
\addlinespace[4pt]

\textbf{Legal two-stage training}\par
\S\,\ref{sec:emergent-coordination-sequential-training}
&
Participant weights first\newline
(base coordinator fixed)\newline
Then coordinator weights\newline
(selected participant fixed)
&
Multi-Agent Full-Context\newline
\CoordinatorSelectedSpeaker{}\newline
Instance-specific PIU distribution
and source material\newline
Free-form decisions, evaluator,\newline
legal budgets, CUAD selection
\\ \midrule

\textbf{Legal joint training}\par
\S\,\ref{sec:emergent-coordination-joint}
&
Participant and coordinator weights
updated together\newline
Base pair versus selected trained pair
&
Same legal interface and budgets\newline
\CoordinatorSelectedSpeaker{}\newline
Instance-specific PIU distribution
and source material, evaluator\newline
Separate policy parameter sets
\\ \midrule

\textbf{Legal transfer}\par
\S\,\ref{sec:coordinator-selected-policy-training}\par
26 MAUD instances
&
CUAD versus held-out MAUD cohort\newline
Base versus trained policies
on each MAUD instance
&
Trained checkpoints fixed during transfer\newline
Within each MAUD instance\newline
Full-Context/coordinator interface,\newline
PIU distribution, source material,\newline
evaluator, legal budgets\textsuperscript{e}
\\ \midrule

\textbf{Finance shared-policy training}\par
\S\,\ref{sec:current-speaker-participant-policy-training}
&
Base / trained shared policy\newline
FinDVer validation\newline
FinLongDocQA validation /\newline
four-participant test
&
Multi-Agent PIU-Only\newline
\OneRIUPerParticipant{}\newline
\CurrentSpeakerSelection{}\newline
Paired task, assigned PIUs,\newline
evaluator, finance budgets
\\ \midrule

\textbf{Cross-replay}\par
Appendix~\ref{app:deferred-communication-trajectory-table},
Table~\ref{tab:finance-current-speaker-cross-replay}
&
Base / trained trajectory\newline
$\times$ base / trained final-decision policy
&
40 jointly completed validation instances\newline
Within each trajectory\newline
Transcript, final decider's identity
and assigned PIU, final prompt
\\ \midrule

\textbf{Common final-decision policy}\par
Appendix~\ref{app:finance-current-speaker-trajectory-analysis}
&
Base versus trained transcripts\newline
Communication diagnostics\newline
Includes incomplete episodes
&
Paired task and evaluator\newline
One GPT-5.5 final-decision policy\newline
39 FinDVer validation and\newline
40 FinLongDocQA test instances\newline
Separate GPT-5.4 PIU audit
\\
\end{longtable}
\arrayrulecolor{black}

\begingroup
\fontsize{9}{10.5}\selectfont
\setlength{\parskip}{4pt}
\textsuperscript{a,b}Transcript and finance query interventions use Qwen3-32B
and Qwen3-235B-A22B. Finance varies queries under \OneRIUPerParticipant{}
and \AllButOneRIUConcentration{}, regenerating messages and decisions
with the source episode's speaker sequence fixed.
\textsuperscript{c}SARA adds GPT-5.5. Source access spans both distributions
and five structures; targeted controls use \AllButOneRIUConcentration{}
with \CoordinatorSelectedSpeaker{}, \BroadcastQuery{}, and
\SimultaneousResponses{}.
\textsuperscript{d}The note/message grid uses 24 CUAD and 29 Multi-LexSum
instances, with GPT-5.5 main and Qwen3-32B corpus-level results.

\textsuperscript{e}Legal checkpoints use CUAD selection; MAUD is excluded
from training and selection. The primary policies start from
Qwen2.5-7B-Instruct, using GRPO and GPT-5.5 reward. The joint and shared-policy
learning curves summarize three recorded histories each; the two-stage and
trajectory-diagnostic results retain their representative-run scope.
An additional finance run starts from Llama-3.1-8B-Instruct's own weights
(Appendix~\ref{app:proposal-llama-finance-pilot}).
Legal caps are 48-token messages, 64-token queries, and 256-token decisions.
Finance uses a 15-round budget, 600-token messages, and a 1,200-token decision
field (Appendix~\ref{app:coordinator-free-policy-training}).
Four-participant FinLongDocQA is absent from training; four-participant
FinDVer occurs in training.
\par
\endgroup
}
\endgroup
\clearpage

\section{Closest Benchmark Comparisons}
\label{app:benchmark-comparisons}
\label{app:hiddenbench-comparison}
\label{app:silobench-comparison}

\paragraph{HiddenBench.}
HiddenBench is the closest benchmark to \ours{} because both study team
decisions under decentralized information structures. HiddenBench provides
broad coverage of Hidden Profile settings (Version~4 evaluates 65 problem
instances and 15 models) and examines whether directly presented
participant-specific facts enter public discussion and improve group voting
\citep{li2025hiddenprofile,li2026hiddenbench}. \ours{} complements this breadth
by treating the source-grounded information and interaction environment as
controlled experimental variables: it varies PIU distribution, whether
participants receive PIUs directly or source material containing them,
communication structure, and token budgets while holding the underlying
problem and evidence requirements fixed. Paired contrasts and
PIU-distribution--communication-structure regressions quantify how performance
changes across these conditions, while PIU-level process tracing localizes
plausible stages at which required evidence is lost. \ours{} additionally
supports a participant-only stage, a coordinator-only continuation, and joint
two-policy training under \CoordinatorSelectedSpeaker{}, together with training
of a shared participant policy under \CurrentSpeakerSelection{}, with the
participant that ends communication making the final decision.
Thus, HiddenBench provides a broader Hidden Profile evaluation, whereas
\ours{} provides a more controlled, source-grounded environment for studying
how PIU distribution and communication structure jointly relate to team
performance.

\paragraph{SILO-BENCH.}
SILO-BENCH studies role-free agents solving synthetic distributed-computation
problems under direct-message, broadcast, and shared-file protocols, with
programmatically known answers and weak-scaling evaluations from 2 to 100
agents \citep{zhang2026silobench}. This design is particularly useful for
examining how emergent organization, consensus, and communication cost change
as both team size and the global problem grow. \ours{} instead holds each
real-world decision problem and its evidence requirements fixed while varying
PIU distribution, whether participants receive PIUs directly or source
material containing them, communication structure, and token budgets. Its
communication gap compares Single-Agent All-PIU with Multi-Agent PIU-Only,
while its retrieval gap compares Multi-Agent PIU-Only with Multi-Agent
Full-Context. These outcome-based contrasts are distinct from SILO-BENCH's
log-based Communication-Reasoning Gap. \ours{} also supports a participant-only
stage, a coordinator-only continuation, and joint two-policy training under
\CoordinatorSelectedSpeaker{}, together with training of a shared participant
policy under \CurrentSpeakerSelection{}, with the participant that ends communication
making the final decision. Consequently, SILO-BENCH
characterizes how emergent collaboration behaves under weak scaling, whereas
\ours{} more directly examines how controlled PIU distributions and
communication structures relate to performance on fixed, source-grounded team
decisions.

\section{Communication-Structure Implementation and Relation to Prior Work}
\label{app:communication-structure-implementation}

\paragraph{Message generation and structure-specific implementation.}
At each communication round, the environment constructs the current
observation permitted by the communication structure and resource budgets.
The applicable participant or coordinator policy receives this observation
and generates an action permitted for its role, such as a participant message,
coordinator query and speaker selection, or final decision. The implementation
uses role-specific prompt templates and structured output schemas, parses the
returned action, and applies structure-specific fallback rules. Model and
decoding settings are reported with the corresponding evaluation or training
setup. Under \CurrentSpeakerSelection{}, an omitted or invalid
next-participant selection defaults to the next participant in the fixed
cyclic order.

\paragraph{Relation to prior work.}
The three structures without coordinator control are related to public
deliberation and multi-agent debate
\citep{gerardi2007deliberative,du2024multiagentdebate}, whereas the two
coordinator-controlled structures are related to mediated communication and
fusion-center models
\citep{myerson1986multistage,forges1986approach,varshney1997distributed}.
Their speaker-selection rules also draw on mechanisms studied in conversation
analysis and multi-agent systems
\citep{sacks1974simplest,wu2023autogen,kim2019learning}. The five structures
are not intended as an exhaustive taxonomy of communication structures.

\section{Problem-Instance Construction and Statistics}
\label{app:problem-instance-construction}

We report source-specific construction choices, checks, and problem-instance
statistics for the legal, finance, medical, and software-engineering source
corpora. Sections~\ref{sec:mathematical-formulation}--\ref{sec:real-source-problems}
define the terminology and general construction requirements; here we provide
implementation details and source-specific statistics.

\paragraph{Source terminology.}
\label{app:source-terminology}
\begin{itemize}
    \item A \textit{source corpus} is the collection of records, annotations,
    or other artifacts from which problem instances are constructed.
    \item The \textit{underlying source materials} for a problem instance $q$
    are the particular records or artifacts selected from its source corpus.
    \item \textit{Source material} denotes the portions of the underlying
    source materials placed in participants' initial observations.
\end{itemize}

\subsection{Decision Sufficiency and Pivotal Information}
\label{app:piu-formalization}

For a problem instance $q$ and an evidence subset $S$, let $v_q(S)$ denote the
corresponding \textit{information view}: the decision objective, decision
format, any other fixed public information, and the evidence in $S$. The view
excludes omitted source information, the reference decision, evaluator-side
annotations, and other hidden construction metadata.

Let $\mathcal{Q}_q$ be a comparison class containing $q$ and other admissible
problem instances that preserve $q$'s public decision specification and set
$\cD_q$ of possible final decisions. Admissibility is defined relative to the
problem family: an admissible instance must satisfy the same domain constraints
and evaluation rule while potentially differing in information not contained
in the given view. For an information view $v$, define
\[
\Omega_q(v)
=
\left\{
\tilde q\in\mathcal{Q}_q :
\tilde q \text{ is consistent with } v
\right\}.
\]
Thus, the information in $v$ does not distinguish $q$ from the other instances
in $\Omega_q(v)$.

For each $\tilde q\in\mathcal{Q}_q$, let
\[
\mathcal{D}_{\tilde q}^{\star}
=
\arg\max_{d\in\cD_q} U_{\tilde q}(d)
\]
be its set of optimal final decisions. We call $S$
\textit{decision-sufficient} for $q$ if
\[
\bigcap_{\tilde q\in\Omega_q(v_q(S))}
\mathcal{D}_{\tilde q}^{\star}
\neq \varnothing.
\]
Equivalently, the information view identifies at least one final decision that
is optimal for every admissible problem instance consistent with that view.

We call $\mathcal{I}^{\min}_q=\{i_1,\ldots,i_k\}$ the \textit{pivotal information} for $q$,
and each $i_j\in\mathcal{I}^{\min}_q$ a \textit{pivotal information unit (PIU)}, if
\[
\bigcap_{\tilde q\in\Omega_q(v_q(\mathcal{I}^{\min}_q))}
\mathcal{D}_{\tilde q}^{\star}
\neq \varnothing
\]
and, for every $i_j\in\mathcal{I}^{\min}_q$,
\[
\bigcap_{\tilde q\in
\Omega_q(v_q(\mathcal{I}^{\min}_q\setminus\{i_j\}))}
\mathcal{D}_{\tilde q}^{\star}
=
\varnothing.
\]
The pivotal information therefore identifies a common optimal decision, whereas
after removing any PIU, no final decision is guaranteed to be optimal across
all admissible instances consistent with the remaining information.

The construction procedure does not enumerate $\Omega_q(v_q(S))$. Instead, it
operationalizes the complete-set and leave-one-out conditions using source
annotations, reference reasoning, grounding checks, and model-based sufficiency
checks. The resulting $\mathcal{I}^{\min}_q$ need not be unique: another inclusion-minimal
evidence set may also identify the decision. Separately, the centralized checks
under Single-Agent All-PIU test whether the reference decision can be produced
from the selected pivotal information.

\subsection{General Construction Procedure}
\label{app:general-problem-construction}

\paragraph{Reference decisions and reasoning.}
Each problem instance is constructed from one or more source questions; the
decision objective may retain one question or combine related questions. When
a reference decision and the reasoning used to produce it accompany the source
questions, we retain them. Otherwise, we construct the reference decision from
the available source annotations and underlying source materials. When the
reasoning is unavailable, we generate reasoning conditioned on the reference
decision; answer-conditioned rationale generation provides one example of
this procedure \citep{wang2023scott}.

\paragraph{Candidate-unit and removal checks.}
The fixed decision objective, reference decision, and reasoning used to produce
it identify the information in the underlying source materials that supports
the required answer components or reasoning steps. Dividing this information
into candidate units is source-specific; a candidate unit may be a sentence,
paragraph, or code block. The construction procedure requires each retained
candidate unit to be necessary for at least one required answer component or
reasoning step. When the retained or generated reasoning does not establish
whether this requirement holds, a model-based check using a model such as
GPT-5.5 first requires the model to produce the reference decision. The
procedure then removes one candidate unit at a time and asks whether uncertainty
remains about the reference decision given the decision objective and remaining
units. The removed unit is retained when uncertainty remains and excluded
otherwise. This process continues until removing any retained unit leaves
uncertainty about the reference decision; the retained units form $\mathcal{I}^{\min}_q$. When
necessity is established from the retained or generated reasoning, the released
construction code also provides the model-based removal check as an additional
check.

\paragraph{PIU and source-material distribution.}
For \PairwiseRIUOverlap{}, we use a seeded random ordering to assign PIUs to
the participant positions defined by the condition.
Appendix~\ref{app:multilexsum-pairwise-overlap-ordering} reports an additional
Multi-LexSum analysis that orders PIUs using embedding-based similarity while
preserving the condition. After the PIU distribution is fixed, the underlying
source materials are divided into units without dividing a PIU. Units that do
not contain a PIU are distributed across participants using source-specific
procedures that may account for their distance from a participant's PIUs and
balance the lengths of participants' initial local observations. Afterward, we
verify each participant's initial local observation together with the initial
common observation and, when necessary and consistent with the source,
truncate the material. For example, legal source materials may be truncated,
whereas source code is not truncated when doing so would prevent the program
from functioning.

\paragraph{Participant identifiers and participant descriptions.}
A participant description is a prompt-visible phrase describing a
participant's subject matter or function without disclosing the specific
content, values, or conclusions of the PIUs assigned to that participant.
Participant identifiers and participant descriptions disclosed to every
decision maker form part of the initial common observation.
Because participant descriptions indicate which participant is likely to hold
relevant information, decision makers can use them to inform speaker selection.
Participant descriptions do not determine the structural roles of participant,
coordinator, or separate final decision maker defined in
Section~\ref{sec:mathematical-formulation}. The source-specific procedures below
describe their construction and leakage checks.

\subsection{Legal}
\label{app:legal-problem-construction}
\label{app:legal-generation-procedure}

Table~\ref{tab:legal-source-corpus-summary} summarizes the six legal problem
families and their source corpora.

\begin{table}[!h]
\centering
\small
\setlength{\tabcolsep}{4pt}
\renewcommand{\arraystretch}{1.03}
\caption{\textbf{Legal problem families and source corpora.}}
\label{tab:legal-source-corpus-summary}
\begin{tabular}{@{}
    >{\centering\arraybackslash}m{0.14\textwidth}
    >{\centering\arraybackslash}m{0.37\textwidth}
    >{\centering\arraybackslash}m{0.20\textwidth}
    >{\centering\arraybackslash}m{0.23\textwidth}
    @{}}
\toprule
Source corpus &
Underlying source materials &
Reference decision &
Problem family \\
\midrule
CUAD & Expert-annotated commercial contracts & Constructed alternative & Contract-profile decision \\
MAUD & Expert-curated merger-agreement provisions & Constructed alternative & Merger-review decision \\
eCFR Casebook & Regulations and controlled transaction facts & Written conclusion & Compliance analysis \\
MSLR & Anonymized enforcement case files & Final sanction record & Enforcement conclusion \\
Multi-LexSum & Documents from a litigation collection & Case-grounded decision & Procedural analysis \\
SARA & Tax facts and supplied statutory provisions & Numeric result & Tax calculation \\
\bottomrule
\end{tabular}
\end{table}

\subsubsection{Common Legal Construction Procedure}

\paragraph{Step 1. Select the underlying source materials.}
A problem instance draws
from exactly one contract, agreement, transaction specification, case file,
litigation collection, or tax case, according to its source corpus.

\paragraph{Step 2. Specify the decision objective and format.}
For CUAD and MAUD, a construction
model uses one or more questions and their grounded excerpts from the source corpus
to construct a new finite-choice legal-review question and reference decision. It
constructs a decision set \(\cD_q\) of mutually exclusive, legally plausible alternatives with
comparable specificity. The other alternatives have similar complexity to
the reference decision, and the alternatives are written as natural legal
profiles rather than explicit sums of answer components. For CUAD and MAUD, the
grounded excerpts used to support the reference decision are candidate units.
The other four source corpora use written or numeric final decisions.
Construction uses the reference decision and the reasoning used to produce it
to identify candidate units in the underlying source materials. The
source-specific procedures below determine which candidate units form \(\mathcal{I}^{\min}_q\).

\paragraph{Step 3. Identify the pivotal information.}
We apply the PIU construction requirements in
Section~\ref{sec:real-source-problems}. If the checks identify
multiple inclusion-minimal sets that satisfy these requirements, CUAD and MAUD
retain one as the pivotal information before constructing the PIU distribution.
eCFR Casebook, MSLR, and Multi-LexSum identify the pivotal information directly
from units needed for distinct reasoning steps.

\paragraph{Step 4. Define the PIU distribution.}
We apply the PIU-distribution conditions in
Figure~\ref{fig:piu-distribution-conditions} subject to the construction requirements
in Section~\ref{sec:real-source-problems}. Some conditions are ineligible for
problem instances with only two PIUs; we do not add artificial PIUs to force a
condition.

\paragraph{Step 5. Construct participant descriptions.}
Each participant receives a participant description based on the PIUs assigned to that
participant and not included in the initial common observation. For each PIU, a
construction model produces a short phrase describing its broad subject or
function. The phrase must not disclose the PIU's specific content,
values, conclusions, or effect on the final decision. The phrases corresponding
to the PIUs assigned to a participant are combined deterministically to form
the participant description for that participant. A participant
assigned no PIU outside the initial common observation receives the participant
description \emph{context reviewer}, followed by the participant number. The
participant descriptions form part of the initial common observation. For eCFR Casebook,
MSLR, Multi-LexSum, and SARA, a separate model-based check assesses this
requirement, and descriptions that fail the check are revised.

\paragraph{Step 6. Check initial observations and complete problem-instance construction.}
We apply the participant-level source-material check in
Section~\ref{sec:real-source-problems}. Removable non-PIU information that
permits one participant to produce the correct final decision alone is deleted
and the check is repeated; a problem instance is rejected if such information
cannot be removed. Finite-choice labels are permuted after construction to
prevent position regularities.

\subsubsection{Problem Instances and PIU Distributions in the Legal Evaluation}
\label{app:legal-model-comparison-problems}
The legal model comparison evaluates the same 163 retained problem instances
under the three evaluation conditions. Because CUAD and MAUD contain
substantially more retained problem instances than the other legal source
corpora, we use 30 problem instances from each, with at most one problem
instance per contract or agreement. For eCFR Casebook, MSLR, Multi-LexSum, and
SARA, we use every retained problem instance. Each problem instance uses
the \OneRIUPerParticipant{} condition in the two team conditions.

The PIU-distribution--communication-structure evaluation uses 514 PIU
distributions under the five PIU-distribution conditions in
Figure~\ref{fig:piu-distribution-conditions}. For each of the eight models, the
evaluation design crosses each PIU distribution with the five communication structures in
Figure~\ref{fig:communication-structure-taxonomy} under Multi-Agent PIU-Only and
Multi-Agent Full-Context.

\subsubsection{Legal Problem-Instance and PIU-Distribution Statistics}
\label{app:legal-problem-instance-statistics}

\begin{figure}[!h]
    \centering
    \includegraphics[width=\textwidth]{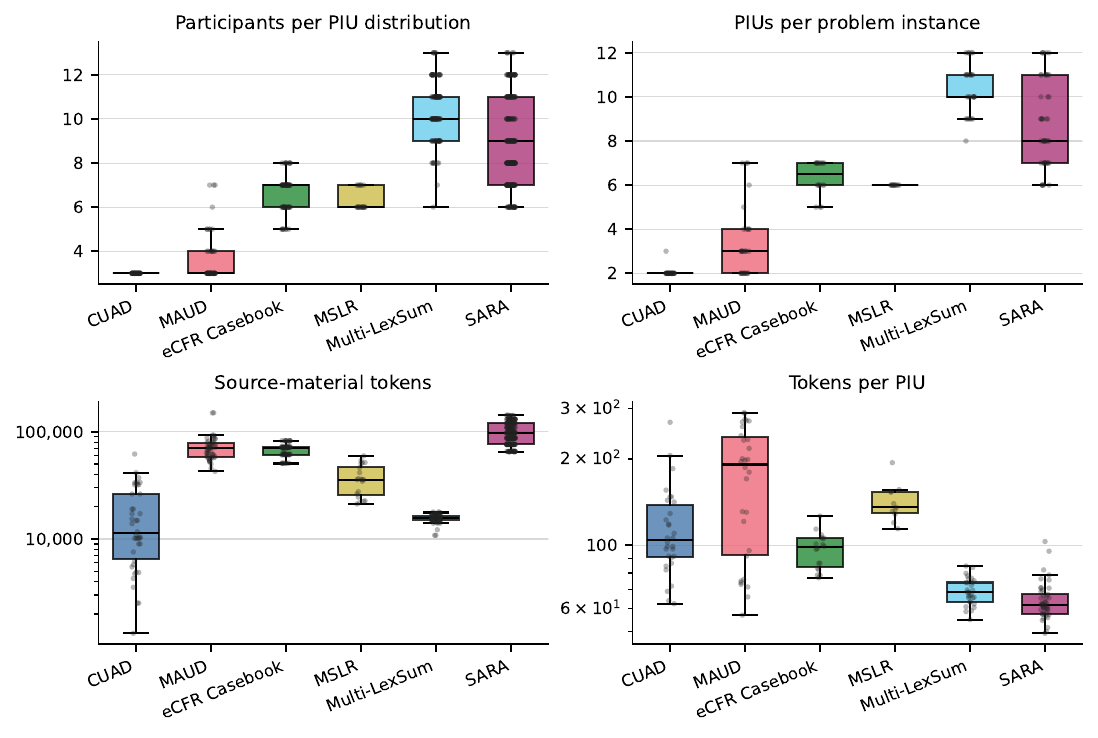}
    \caption{\textbf{Problem-instance and PIU-distribution statistics in the
    legal evaluation.} In the PIU-count and PIU-length panels, each point is one
    problem instance. In the participant-count and source-material-length
    panels, each point is one PIU distribution. Boxes show the median and
    interquartile range; token-length panels use logarithmic axes.}
    \label{fig:legal-problem-instance-structural-statistics}
\end{figure}

Figure~\ref{fig:legal-problem-instance-structural-statistics} reports the
PIU count and PIU length for the 163 problem instances in the legal evaluation.
It also reports participant count and the length of source material provided to
participants for the 514 PIU distributions under the five PIU-distribution
conditions in Figure~\ref{fig:piu-distribution-conditions}. For the first two
measures, each point is one problem instance; for the latter two, each point is
one PIU distribution.

Source-material length sums the material provided to each participant after
applying the initial-local-observation token budget of 40,000 tokens. Repeated
material is counted for each participant who receives it because it contributes
separately to that participant's retrieval burden. PIU length is measured from
the PIU text supplied in Multi-Agent PIU-Only. All lengths use the same
\texttt{cl100k\_base} tokenizer used for initial-local-observation budgeting.
Table~\ref{tab:legal-problem-instance-structural-statistics} reports the
corresponding means; measures that depend on PIU distribution are first averaged
within problem instance and then across problem instances.

\begin{table}[!h]
\centering
\scriptsize
\setlength{\tabcolsep}{2.5pt}
\renewcommand{\arraystretch}{1.08}
\caption{\textbf{Mean problem-instance and PIU-distribution statistics for the legal evaluation.} Problem-instance measures use all 163 problem instances. PIU-distribution measures use 514 distributions. For measures that depend on PIU distribution, we first average within problem instance and then across problem instances. Source-material lengths reflect the material provided after applying the initial-local-observation token budget of 40,000 tokens.}
\label{tab:legal-problem-instance-structural-statistics}
\begin{tabular*}{\textwidth}{@{\extracolsep{\fill}}lrrrrrrr@{}}
\toprule
Source corpus & \makecell{Problem\\instances} & \makecell{PIU\\distributions} & \makecell{Participants/\\distribution} & PIUs & \makecell{Source-material tok./\\distribution} & \makecell{Source-material tok./\\participant} & \makecell{PIU tok./\\unit} \\
\midrule
CUAD & 30 & 42 & 3.0 & 2.0 & 15,644.9 & 5,215.0 & 116.4 \\
MAUD & 30 & 53 & 3.5 & 3.4 & 70,859.8 & 20,759.0 & 174.8 \\
eCFR Casebook & 18 & 61 & 6.7 & 6.3 & 68,383.3 & 10,270.3 & 96.3 \\
MSLR & 11 & 20 & 6.3 & 6.0 & 35,948.0 & 5,688.9 & 141.7 \\
Multi-LexSum & 29 & 113 & 10.4 & 10.4 & 15,621.4 & 1,535.4 & 69.2 \\
SARA & 45 & 225 & 8.9 & 8.7 & 97,936.4 & 10,969.9 & 64.3 \\
\midrule
All source corpora & 163 & 514 & 6.7 & 6.4 & 55,744.2 & 9,555.2 & 103.9 \\
\bottomrule
\end{tabular*}
\end{table}

\subsubsection{CUAD Problem-Instance Construction}
\label{app:cuad-conversion-notes}

\paragraph{Source corpus and retained problem instances.}
CUAD \citep{hendrycks2021cuad} provides contract-level annotations across
commercial clause categories. We do not use the original CUAD label as
the final decision. A question such as ``does this contract contain source-code
escrow?'' does not require multiple PIUs because one annotated \texttt{span}
can determine the final decision. Instead, we use contract--label pairs with
two to five positive annotated \texttt{span}s to construct finite-choice
problem instances. Contract--label pairs with negative labels are excluded
because the decision objective requires combining information rather than
proving absence from a long contract.

The construction procedure produces 542 candidate problem instances. Of these,
539 admit at least one PIU distribution under the conditions in
Figure~\ref{fig:piu-distribution-conditions}, and 538 are retained after the common
legal checks and the CUAD-specific checks below. The 538 retained problem
instances, which are reported as the constructed CUAD problem instances in
Table~\ref{tab:problem-instance-counts}, are drawn from 289 distinct contracts.
Most retained problem instances (514) contain two PIUs; the remaining 24
contain at least three. For the legal model comparison, we use the first
retained problem instance from each previously unseen contract in the released
order until 30 contracts are represented.

\paragraph{Decision objective and PIU construction.}
For each eligible contract--label pair, a construction model selects positive
annotated \texttt{span}s and constructs a finite-choice problem instance whose
final decision depends on their content.

Each candidate unit is one CUAD-annotated \texttt{span}. If a \texttt{span}
contains several legal facts, they remain within one candidate unit; if two
\texttt{span}s express similar legal content, they remain separate candidate
units.

Using the selected candidate units, construction follows the common legal
procedure in Appendix~\ref{app:legal-generation-procedure} to construct the
question, decision set, and reference decision. The alternatives use neutral,
opaque identifiers, with their legal content placed in their definitions. The
model-based check in Section~\ref{sec:real-source-problems} determines which
candidate units form \(\mathcal{I}^{\min}_q\). Because CUAD problem instances use a finite-choice
decision format, each constructed problem instance must admit exactly one
reference decision. Ambiguous questions such as ``is this clause risky?'' or
``is this protection good enough?'' are rejected unless the alternatives are
specified so that the problem instance admits one reference decision.

Table~\ref{tab:cuad-positive-conversion} summarizes the CUAD label families
used in construction.

\begin{table}[!h]
\centering
\small
\setlength{\tabcolsep}{4pt}
\renewcommand{\arraystretch}{1.10}
\caption{\textbf{CUAD label families used in problem-instance construction.}
The decision objective specifies a legal profile supported by multiple
annotated spans rather than the original CUAD label.}
\label{tab:cuad-positive-conversion}
\begin{tabular}{@{}
    >{\centering\arraybackslash}m{0.19\linewidth}
    >{\raggedright\arraybackslash}m{0.43\linewidth}
    >{\raggedright\arraybackslash}m{0.30\linewidth}
    @{}}
\toprule
\multicolumn{1}{c}{CUAD label family} &
\multicolumn{1}{c}{Information in annotated spans} &
\multicolumn{1}{c}{Decision objective} \\
\midrule
Source Code Escrow &
deposit, release trigger, post-release use right, update duty, cost &
source-code continuity protection strength \\
ROFR/ROFO/ROFN &
covered opportunity, notice, exercise period, matching or negotiation right, expiration &
transaction restriction strength \\
Insurance &
coverage type, coverage limit, certificate, cancellation notice, additional insured, remedy &
insurance burden or protection level \\
Audit Rights &
recordkeeping, records review, facility or system inspection, notice, frequency, cost &
audit burden level \\
License Grant &
grant, field of use, territory, duration, sublicense, use restriction &
license sufficiency for intended use \\
Minimum Commitment &
amount, period, purchase or revenue metric, shortfall remedy, exceptions &
commercial commitment materiality \\
Cap on Liability &
cap amount or formula, excluded damages, carveouts, special warranty or IP rule &
liability protection profile \\
\bottomrule
\end{tabular}
\end{table}

\paragraph{Source material and PIU distribution.}
For CUAD, the underlying source material is the source contract, rather than
only the annotated \texttt{span}s that determine the reference decision. The
source contract is divided into contiguous units using legal section headings
first, then paragraphs, sentences, and words as progressively weaker split
points. A unit does not divide a PIU. Units containing PIUs are provided to the
participants assigned those PIUs under the PIU distribution.

Units that do not contain a PIU are distributed based on their overlap with
participant descriptions and PIUs, their proximity to PIUs in the source contract,
their legal section, and the lengths of participants' initial local
observations. A unit with overlap or proximity to a PIU is, when possible,
provided to a participant not assigned that PIU.

\paragraph{Example participant descriptions.}
The CUAD label is not provided to participants.
Table~\ref{tab:cuad-participant-description-examples} provides example participant descriptions
constructed under the common legal procedure.

\begin{table}[!h]
\centering
\scriptsize
\setlength{\tabcolsep}{4pt}
\renewcommand{\arraystretch}{1.10}
\caption{\textbf{Example participant descriptions for CUAD problem instances.}}
\label{tab:cuad-participant-description-examples}
\begin{tabular}{@{}
    >{\centering\arraybackslash}m{0.15\linewidth}
    >{\raggedright\arraybackslash}m{0.79\linewidth}
    @{}}
\toprule
\multicolumn{1}{c}{CUAD label family} &
\multicolumn{1}{c}{Example participant descriptions} \\
\midrule
Source Code Escrow &
deposit reviewer; release-condition reviewer; use-right reviewer; update/cost reviewer \\
ROFR/ROFO/ROFN &
covered-opportunity reviewer; notice reviewer; exercise-period reviewer; post-expiration reviewer \\
Insurance &
coverage-type reviewer; coverage-limit reviewer; certificate/notice reviewer; special-coverage reviewer \\
Audit Rights &
records reviewer; access reviewer; operations-inspection reviewer; process/cost reviewer \\
\bottomrule
\end{tabular}
\end{table}

\paragraph{CUAD-specific construction checks.}
In addition to the common legal checks in
Appendix~\ref{app:legal-generation-procedure}, CUAD requires each candidate unit
to be a positive annotated \texttt{span} copied exactly from the source contract
with a recoverable source location. Each problem instance must contain at least
two PIUs. The units used to divide the source contract must collectively cover
it without dividing a PIU. A problem instance is rejected if it fails any of
these checks.

\subsubsection{MAUD Problem-Instance Construction}
\label{app:maud-conversion-notes}

\paragraph{Source corpus and retained problem instances.}
MAUD \citep{wang2023maud} pairs merger agreements with expert-curated
annotations of their provisions. The source corpus contains 25,827 annotation
rows from 153 agreements. Construction uses 100 agreements and produces 557
retained problem instances from 318 agreement--category pairs. Of those pairs,
239 produce two problem instances and 79 produce one. Among the 557 retained
problem instances, 178 contain two PIUs and 379 contain at least three.
The legal evaluation uses 30 fixed problem instances from distinct agreements,
as described in Appendix~\ref{app:legal-model-comparison-problems}.

The source corpus organizes annotation rows under seven legal categories. The
retained problem instances use the four categories in
Table~\ref{tab:maud-source-categories}.

\begin{table}[!h]
\centering
\small
\setlength{\tabcolsep}{4pt}
\renewcommand{\arraystretch}{1.08}
\caption{\textbf{MAUD categories used in problem-instance construction.}
Each problem instance uses provision excerpts from one agreement and one
category.}
\label{tab:maud-source-categories}
\begin{tabular}{@{}
    >{\centering\arraybackslash}m{0.24\linewidth}
    >{\raggedright\arraybackslash}m{0.41\linewidth}
    >{\raggedright\arraybackslash}m{0.27\linewidth}
    @{}}
\toprule
\multicolumn{1}{c}{MAUD category} &
\multicolumn{1}{c}{Information in provision excerpts} &
\multicolumn{1}{c}{Decision objective} \\
\midrule
Deal protection and related provisions &
no-shop limits, fiduciary-out conditions, matching periods,
superior-proposal conditions, and termination-fee provisions &
board-process or termination-trigger profile \\
\arrayrulecolor{black!18}\specialrule{0.3pt}{0.8pt}{0.8pt}\arrayrulecolor{black}
Conditions to closing &
representation bring-down timing and standard, materiality treatment, and
fundamental-representation treatment &
closing-condition risk profile \\
\arrayrulecolor{black!18}\specialrule{0.3pt}{0.8pt}{0.8pt}\arrayrulecolor{black}
Material adverse effect &
base definition, carveouts, disproportionate-impact treatment, and pandemic
or market exceptions &
material-adverse-effect risk-allocation profile \\
\arrayrulecolor{black!18}\specialrule{0.3pt}{0.8pt}{0.8pt}\arrayrulecolor{black}
Operating and efforts covenant &
ordinary-course covenant, negative covenants, buyer consent rights,
regulatory-efforts language, and emergency exceptions &
pre-closing control profile \\
\bottomrule
\end{tabular}
\end{table}

\paragraph{Decision objective and PIU construction.}
Many MAUD row-level questions can be answered from one provision excerpt. We
therefore use the annotated questions and excerpts to construct a finite-choice
problem instance whose final decision depends on multiple provisions from the
same agreement and category. The alternatives describe the combined legal
profile implied by those provisions rather than reproduce the MAUD row-level
annotations.

A MAUD row may contain a curated provision excerpt rather than a complete
agreement section, and some rows join passages separated by omitted text.
Construction removes annotation artifacts, divides such rows at omitted-text
markers, and does not reconstruct omitted text. Each resulting provision
excerpt is a candidate unit. Construction then locates the candidate unit in
the agreement, using exact matching when possible and normalization when
necessary. The model-based check in Section~\ref{sec:real-source-problems}
determines which candidate units form \(\mathcal{I}^{\min}_q\). The MAUD row-level annotation is used
to organize candidate units but is not itself a PIU.

\paragraph{Source material and PIU distribution.}
For MAUD, the underlying source material is the merger agreement. The agreement
is divided into contiguous units without dividing a PIU. Units containing
PIUs are provided to the participants assigned those PIUs under the PIU
distribution. The remaining units are distributed based on their overlap with
participant descriptions and PIUs, their proximity to PIUs in the agreement, their
legal section, and the lengths of participants' initial local observations.
When an initial local observation is truncated under the
applicable resource budget, every PIU assigned to that participant remains
present.

\paragraph{MAUD-specific construction checks.}
In addition to the common legal checks in
Appendix~\ref{app:legal-generation-procedure}, MAUD requires every PIU to be a
provision excerpt from a source annotation row and to have a recoverable
location in the agreement. Each problem instance must contain at least two
PIUs from one agreement and one retained category. The units used to divide
the agreement must preserve every PIU. A problem instance is rejected if it
fails any of these checks.

\subsubsection{eCFR Casebook Problem-Instance Construction}
\label{app:long-context-casebook20}

\paragraph{Source corpus and retained problem instances.}
eCFR Casebook uses regulatory text from 12 C.F.R. Parts~1 and~3 together with
controlled transaction memos
\citep{ecfrTitle12Part1,ecfrTitle12Part3,ecfrApi}. The regulatory text covers
bank investment securities and related capital requirements. Each transaction
memo states the facts needed to apply one or more of these provisions.
Construction produces 20 candidate problem instances. Eighteen are retained
after the common legal checks and the eCFR Casebook-specific checks below. The
legal evaluation uses all 18 retained problem instances.

\paragraph{Decision objective and PIU construction.}
Each problem instance asks for a short written compliance conclusion about one
transaction memo. The final decision must state the controlling regulatory
status or rule, the relevant transaction facts, any required threshold,
exception, or calculation, and the resulting action. These requirements form
the answer components evaluated in the final decision. The evaluator for
free-form decisions in Appendix~\ref{app:policy-training-free-form-evaluation} compares the final
decision with the reference decision and these answer components.

Construction selects regulatory provisions whose application depends on more
than one condition. Candidate units consist of regulatory text copied from the
eCFR and transaction facts copied from the memo. The model-based check in
Section~\ref{sec:real-source-problems} determines which candidate units form
\(\mathcal{I}^{\min}_q\). The retained problem instances contain five to seven PIUs.

\paragraph{Source material and PIU distribution.}
For eCFR Casebook, the underlying source materials are the eCFR provisions and
the transaction memo. These materials are divided into long excerpts without
dividing a PIU. Each excerpt containing a PIU is provided to a participant
assigned that PIU under the PIU distribution, and PIUs assigned to other
participants are removed from that participant's source material. Each
participant's source material has a target length of 12,000 tokens; across the
retained problem instances, the resulting lengths range from 9,846 to 11,830
tokens.

\paragraph{eCFR Casebook-specific construction checks.}
In addition to the common legal checks in
Appendix~\ref{app:legal-generation-procedure}, eCFR Casebook requires every PIU
to occur in the regulatory text or transaction memo. Construction checks that
each PIU appears in the initial local observation of every participant assigned
that PIU and not in the initial local observations of participants not assigned
that PIU. A retained problem instance must also pass the model-based
solvability check with all PIUs centralized. A problem instance is rejected if
it fails any of these checks.

\subsubsection{MSLR Problem-Instance Construction}
\label{app:mslr-problem-instance-construction}

\paragraph{Source corpus and retained problem instances.}
MSLR \citep{yu2027mslr} provides anonymized Chinese
administrative-enforcement case files. The source corpus used in construction
contains 1,389 case files, of which 613 identify one target party, contain the
required source material, and state a final judgment. Construction selects 100
candidate problem instances. After the common legal checks and the
MSLR-specific checks below, 11 problem instances are retained. The legal
evaluation uses all 11 retained problem instances.

\paragraph{Decision objective and PIU construction.}
Each problem instance asks for a short written enforcement conclusion for one
target party. The reference decision combines the material nonpublic
information and its timing, the basis for the party's knowledge or access, the
party's trading conduct, the legal characterization, and the final
administrative handling. The final judgment supplies the final administrative
handling. These requirements form the answer components evaluated in the final
decision. The evaluator for free-form decisions in
Appendix~\ref{app:policy-training-free-form-evaluation} compares the final decision
with the reference decision and these answer components.

Construction identifies six exact, nonoverlapping excerpts from the case file.
The excerpts cover the event and timing, knowledge or access, trading conduct,
legal characterization, and final handling. Each candidate unit retains its
location in the case file. We apply the PIU construction requirements in
Section~\ref{sec:real-source-problems}; every retained MSLR problem instance
contains six PIUs.

\paragraph{Source material and PIU distribution.}
For MSLR, the underlying source material is the complete anonymized case file.
Each participant receives the case file with the exact excerpts corresponding
to the PIUs assigned to that participant. Passages corresponding to PIUs
not assigned to that participant are removed before the source material is
provided.

\paragraph{MSLR-specific construction checks.}
In addition to the common legal checks in
Appendix~\ref{app:legal-generation-procedure}, MSLR requires each problem
instance to identify one target party and to contain a final judgment. Every
candidate unit must be copied exactly from the case file, have a recoverable
source location, and not overlap another candidate unit. A retained problem
instance must also pass the model-based solvability check with all PIUs
centralized. A problem instance is rejected if it fails any of these checks.

\subsubsection{Multi-LexSum Problem-Instance Construction}
\label{app:multilexsum-problem-instance-construction}

\paragraph{Source corpus and retained problem instances.}
Multi-LexSum \citep{shen2022multilexsum} contains 4,539 civil-litigation
collections. Of these, 3,086 contain between three and 18 source documents and
have a Multi-LexSum summary within the length range used in construction. A
model-based check of source grounding retains 48 candidate problem instances,
and construction selects 30 for the remaining checks. After the common legal
checks and the Multi-LexSum-specific checks below, 29 problem instances are
retained. The legal evaluation uses all 29 retained problem instances.

\paragraph{Decision objective and PIU construction.}
Each problem instance asks for a written procedural or legal conclusion that
requires two to four interdependent conclusions from the documents in one
litigation collection. The question and reference decision are based on the
original documents rather than the Multi-LexSum summary. Construction removes
from the question the names and source-specific identifiers that would identify
the litigation collection. The evaluator for free-form decisions in
Appendix~\ref{app:policy-training-free-form-evaluation} compares the final decision
with the reference decision and the required answer components.

Construction identifies decision-relevant information in the original
documents and states each candidate unit concisely. Each candidate unit records
the sentence identifiers in the original documents that support it. Each
candidate unit must distinguish the reference decision from at least one
alternative considered during construction. We apply the PIU construction
requirements in Section~\ref{sec:real-source-problems}.

A model-based check compares each candidate unit with the
supporting sentences and requires the original documents to support the
complete reference decision. Each retained problem instance contains eight to
twelve PIUs.

\paragraph{Source material and PIU distribution.}
For Multi-LexSum, the underlying source materials are the original documents in
one litigation collection. The source material provided to a participant
contains the original-document sentences supporting each PIU assigned to that
participant, together with additional passages from the same litigation
collection. Units containing PIUs and units containing the additional passages
are distributed without dividing a PIU.

\paragraph{Multi-LexSum-specific construction checks.}
In addition to the common legal checks in
Appendix~\ref{app:legal-generation-procedure}, Multi-LexSum requires the
question to admit a determinate procedural or legal conclusion without
revealing that conclusion or the identity of the litigation collection. The
candidate units must use at least three original-document excerpts, with at
least two candidate units supported by each excerpt. Every candidate unit and
the reference decision must pass the model-based check of source grounding. A
retained problem instance must also pass the model-based solvability check with
all PIUs
centralized. A problem instance is rejected if it fails any of these checks.

\paragraph{Sensitivity to the ordering used for \PairwiseRIUOverlap{}.}
\label{app:multilexsum-pairwise-overlap-ordering}
The PIU-distribution conditions in
Figure~\ref{fig:piu-distribution-conditions} specify which participants initially
observe each PIU, but they do not prescribe an ordering based on similarity
between PIUs. We therefore conduct an exploratory analysis using two
additional \PairwiseRIUOverlap{} distributions for each of the 29 retained
Multi-LexSum problem instances. One uses a seeded random ordering. The other
uses an ordering that places PIUs with similar embedding representations next
to one another.

We represent each PIU using Cohere Embed v4 \citep{cohere2025embed4}, with the
clustering input type and 1,536-dimensional float embeddings, and normalize
each representation. For a PIU ordering $\pi$, we compute
\[
    \frac{1}{|\mathcal{I}^{\min}_q|-1}
    \sum_{j=1}^{|\mathcal{I}^{\min}_q|-1}
    \cos\!\left(
        z(i_{\pi_j}),
        z(i_{\pi_{j+1}})
    \right),
\]
where $z(\cdot)$ maps a PIU to its embedding representation. We find the
ordering that maximizes this quantity using exact dynamic programming with
deterministic tie breaking. The seeded comparison uses a deterministic random
permutation.

For either ordering, participant $j<|\mathcal{I}^{\min}_q|$ initially observes
$i_{\pi_j}$ and $i_{\pi_{j+1}}$, and the final participant initially observes
$i_{\pi_{|\mathcal{I}^{\min}_q|}}$. The two distributions therefore have the same problem
instance, pivotal information, number of participants, total number of PIU assignments,
and number of PIUs in each participant's initial local observation. They
differ only in which PIUs occupy these positions.

We evaluate both distributions in Multi-Agent PIU-Only using
Qwen3-235B-A22B, \CoordinatorSelectedSpeaker{}, 15 participant responses, and
the resource budgets used in the legal model comparison. We apply the
evaluation procedure for free-form decisions in
Appendix~\ref{app:policy-training-free-form-evaluation} using GPT-5.4 as the model
judge. Table~\ref{tab:multilexsum-pairwise-overlap-ordering} reports the paired
results.

\begin{table}[!h]
\centering
\small
\setlength{\tabcolsep}{5pt}
\caption{\textbf{Sensitivity to the ordering used to construct
\PairwiseRIUOverlap{} distributions for Multi-LexSum.}
Values average over the same 29 problem instances. Smaller $\pm$ terms are
standard errors across problem instances.}
\label{tab:multilexsum-pairwise-overlap-ordering}
\begin{tabular}{lrrrr}
\toprule
Ordering &
\makecell{Mean adjacent\\cosine similarity} &
\makecell{Episode\\success (\%)} &
\makecell{LLM-judged\\score (\%)} &
\makecell{Answer-component\\coverage (\%)} \\
\midrule
Seeded random
& .449 & 13.8\uncert{6.4} & 65.2\uncert{4.3} & 80.7\uncert{3.4} \\
Maximum similarity
& .598 & 27.6\uncert{8.3} & 71.9\uncert{3.5} & 87.4\uncert{2.8} \\
\bottomrule
\end{tabular}
\end{table}

The maximum-similarity ordering increases the intended construction
measurement: mean adjacent cosine similarity increases from .449 to .598. Its
episode success rate, LLM-judged score, and answer-component coverage are
descriptively higher than under the seeded random ordering. The paired
difference in the LLM-judged score is 6.7 percentage points, with a
problem-instance-bootstrap 95\% confidence interval of $[-3.1,16.3]$
percentage points.
Thus, this exploratory analysis does not establish a difference between the
two PIU distributions. We report it to document a construction based on PIU
similarity, not to rank PIU-distribution procedures or select an allocation
using evaluation outcomes.

\paragraph{Sensitivity to the assignment of participant descriptions.}
\label{app:multilexsum-participant-description-permutation}

Following Step~5 of the common legal construction procedure
(Appendix~\ref{app:legal-generation-procedure}), each participant receives a
participant description constructed from the broad subject or function of its assigned
PIU. The description must not disclose the PIU's specific content, values,
conclusions, or effect on the final decision. The 301 participant descriptions used
across the 29 retained Multi-LexSum problem instances pass the model-based
check specified in that procedure.

We construct two versions of each problem instance. In one version, each
participant description is attached to the participant assigned the PIU from
which the description was constructed. In the paired version, the same
participant descriptions are permuted across participants. The permutation
preserves every participant description, including repeated descriptions,
while ensuring that the description attached to every participant changes.
Apart from which participant description is attached to each participant, the paired versions have the same
problem instance, pivotal information, PIU distribution, initial local observations,
communication structure, and resource budgets.

We evaluate both versions in Multi-Agent PIU-Only using Qwen3-235B-A22B,
\CoordinatorSelectedSpeaker{}, 15 participant responses, and the resource
budgets used in the legal model comparison. We apply the evaluation procedure
for free-form decisions in Appendix~\ref{app:policy-training-free-form-evaluation} using
GPT-5.4 as the model judge.
Table~\ref{tab:multilexsum-participant-description-permutation} reports the paired
results.

\begin{table}[!h]
\centering
\small
\setlength{\tabcolsep}{5pt}
\caption{\textbf{Sensitivity to the assignment of participant descriptions for
Multi-LexSum.}
Values average over the same 29 problem instances. Smaller $\pm$ terms are
standard errors across problem instances.}
\label{tab:multilexsum-participant-description-permutation}
\begin{tabular}{lrrr}
\toprule
Participant descriptions &
\makecell{Episode success\\rate (\%)} &
\makecell{LLM-judged\\score (\%)} &
\makecell{Answer-component\\coverage (\%)} \\
\midrule
Constructed from participants' assigned PIUs
& 24.1\uncert{8.1}
& 77.6\uncert{1.9}
& 84.8\uncert{2.4} \\
Same descriptions permuted across participants
& 0.0\uncert{0.0}
& 46.9\uncert{4.8}
& 67.2\uncert{3.6} \\
\bottomrule
\end{tabular}
\end{table}

Permuting the participant descriptions decreases the mean LLM-judged score by
30.7 percentage points. The LLM-judged score is lower after permutation for
24 problem instances, higher for three, and unchanged for two; a two-sided
exact sign test gives \(p=4.9\times10^{-5}\). Every PIU appears in the
communication transcript in all 29 problem instances when participant descriptions
are constructed from participants' assigned PIUs, compared with 18 of 29
problem instances when the same descriptions are permuted across participants.
Thus, under \CoordinatorSelectedSpeaker{}, performance depends on whether the
participant descriptions correspond to the PIUs assigned to those participants.
This analysis isolates the assignment of participant descriptions; it does not
compare communication structures or procedures for constructing participant
descriptions.

\subsubsection{SARA Problem-Instance Construction}
\label{app:sara-problem-instance-construction}

\paragraph{Source corpus and retained problem instances.}
SARA \citep{holzenberger2020sara} provides 100 tax cases with case facts,
statutory text, and calculation and reasoning steps. Of these source cases, 83
yield a candidate problem instance and 63 satisfy the source and PIU
requirements described below. Forty-five pass the repeated model-based
solvability checks with all PIUs centralized and are retained. The retained
problem instances contain 393 PIUs, with six to twelve PIUs per problem
instance (mean 8.7), and four to eight answer components (mean 5.5). The legal
evaluation uses all 45 retained problem instances.

\paragraph{Decision objective and PIU construction.}
Construction executes the supplied calculation and reasoning steps and selects
an intermediate conclusion that depends on multiple steps. The selected
conclusion defines a numeric decision objective, such as a deduction, taxable
income, or a tax amount, rather than requiring every problem instance to
recover the original total-tax result. Construction traces the steps backward
from that conclusion to the facts and rules that determine it and fixes the
reference decision before writing the question.

Each candidate unit states one fact or rule needed for at least one of the
selected calculation and reasoning steps. Source-stated facts and statutory
rules retain their source locations. When a required relation appears only in
the supplied calculation and reasoning steps, construction states that
relation in ordinary legal language and records its basis. A calculated
intermediate result is not repeated as a candidate unit when the supplied facts
and rules already determine it. The model-based sufficiency and necessity
check determines which candidate units form \(\mathcal{I}^{\min}_q\).

The final decision must state the requested numeric result and a concise
derivation. The evaluator described in
Appendix~\ref{app:policy-training-free-form-evaluation} compares the requested numeric
result with the reference value and evaluates the answer components that form
the derivation.

\paragraph{Source material and PIU distribution.}
For SARA, the supplied calculation and reasoning steps are used during
problem-instance construction but are not provided to participants. Each
participant's source material contains the fact or rule stated by each PIU
assigned to that participant, together with the case facts and statutory text.
Information stating a PIU not assigned to that participant is removed before
the source material is provided.

\paragraph{SARA-specific construction checks.}
In addition to the common legal checks in
Appendix~\ref{app:legal-generation-procedure}, SARA requires every candidate
unit to have a recorded basis in the case facts, statutory text, or supplied
calculation and reasoning steps. A fact or rule must resolve each condition
needed to determine the selected conclusion, including a condition that rules
out an exception. We apply the PIU construction requirements in
Section~\ref{sec:real-source-problems}. Each problem instance then undergoes five model-based solvability
checks with all PIUs centralized and is retained when at least four succeed. A
problem instance is rejected if it fails any of these checks.

\subsection{Finance}
\label{app:finance}
We construct finance problem instances from FinDVer
\citep{zhao2024findver} and FinLongDocQA \citep{wang2026finlongdocqa}.
Each source corpus corresponds to one problem family, which we call the
FinDVer and FinLongDocQA problem families, respectively. The procedure below
reports the choices specific to these source corpora and problem families.
Table~\ref{tab:finance-source-corpus-summary} summarizes the finance problem
families and their source corpora.

\begin{table}[!h]
\centering
\small
\setlength{\tabcolsep}{4pt}
\renewcommand{\arraystretch}{1.03}
\caption{\textbf{Finance problem families and source corpora.}}
\label{tab:finance-source-corpus-summary}
\begin{tabular}{@{}
    >{\centering\arraybackslash}m{0.18\textwidth}
    >{\centering\arraybackslash}m{0.34\textwidth}
    >{\centering\arraybackslash}m{0.22\textwidth}
    >{\centering\arraybackslash}m{0.20\textwidth}
    @{}}
\toprule
Source corpus &
Underlying source materials &
Reference decision &
Problem family \\
\midrule
FinDVer & SEC filings with source questions and annotated contexts & Written filing-review answer & Filing-review reasoning \\
FinLongDocQA & Company-year annual reports with source questions and answers & Written annual-report answer & Annual-report reasoning \\
\bottomrule
\end{tabular}
\end{table}

\subsubsection{Common Finance Construction Procedure}
\label{app:finance-construction}

\paragraph{Step 1. Select the underlying source materials.}
A problem instance uses exactly one U.S. Securities and Exchange Commission
(SEC) filing or company-year annual report.
Native questions, answers, labels, source-text references, and available
calculations, code, and reasoning may be used during construction but do not
form part of the initial common observation.

\paragraph{Step 2. Divide the underlying source materials.}
The filing or report is divided into complete paragraphs and tables while
preserving their source locations. Adjacent fragments may be combined only
when a label in one fragment requires a value in the other to be complete. A
table is not divided into individual cells or values.

\paragraph{Step 3. Identify candidate units.}
For each source question, a model-based source-grounding check identifies every
paragraph, table, or combination of an adjacent label and value that contributes a
report-specific value, definition, period, unit, entity scope, condition,
exception, or contradiction needed to reconstruct the source annotation.
Before problem-instance construction, the procedure records whether candidate
units are exact duplicates, whether one fully contains another, whether one can
be derived from another, and whether they partially overlap. A candidate unit
is removed only when a retained candidate unit contains its entire
contribution; partial overlap alone does not support removal.

\paragraph{Step 4. Specify the decision objective and format.}
Construction combines one or more source questions and at least two candidate
units into a problem instance with a free-form decision format. It fixes the reference
decision, answer components, the candidate units supporting each answer
component, and the evaluation criteria for values, units, periods, signs,
scopes, and conclusions.

\paragraph{Step 5. Identify the pivotal information.}
Each candidate unit is one paragraph, complete table, or combined
adjacent label and value. We apply the PIU
construction and removal checks in Section~\ref{sec:real-source-problems},
using the answer components to assess completeness, correctness, and source
grounding. A redundant candidate unit is removed before the checks are repeated.

\paragraph{Step 6. Construct participant descriptions.}
Each participant receives a source-corpus-specific reviewer description. These
descriptions identify the participant as a reviewer of a filing or annual
report and include a participant number. Construction records broad information
features for PIUs, but participant descriptions and the participant directory do
not disclose PIU values, calculations, conclusions, or their effect on the
reference decision.

\paragraph{Step 7. Define the PIU distribution and distribute source material.}
Each finance problem instance uses the \OneRIUPerParticipant{} condition in
the two team evaluation conditions. The additional PIU distributions reported in
Table~\ref{tab:finance-allocation-counts} use the other conditions from
Figure~\ref{fig:piu-distribution-conditions}. After each PIU distribution is fixed,
the underlying source materials are distributed across participants. The
source material provided to each participant is checked for information that
contains or supports reconstruction of a PIU assigned to another participant.
Such material is removed while every PIU assigned to that participant is
retained.

This source-material check concerns meaning rather than exact wording. Exact
repetition of a PIU assigned to another participant is sufficient for removal
but is not required: a passage or collection of passages is also removed when
it supports reconstruction of the report-specific value, period, unit, entity
scope, condition, exception, or conclusion contributed by that PIU. Partial
topical overlap alone does not support removal. After revising the source
material, the procedure repeats the participant-level check while preserving
every PIU assigned to that participant.

\paragraph{Step 8. Complete problem-instance construction.}
A problem instance is retained only after the specified participants, pivotal information,
PIU distribution, reference decision, and distributed source material pass the
applicable checks.

\subsubsection{Problem Instances and PIU Distributions in the Finance Evaluation}
\label{app:finance-model-comparison-problems}

The construction pipeline produces 389 FinDVer problem instances and 249
FinLongDocQA problem instances. Following the same fixed-subset convention used
for CUAD and MAUD, the finance evaluation uses 30 problem instances
from each source corpus under the three evaluation conditions. The FinDVer
subset is selected deterministically in released order within source-material
strata, with 20 Form 10-K filings and 10 Form 10-Q filings. The FinLongDocQA
subset is selected deterministically in released order within year strata, with
9 reports from 2022, 10 from 2023, and 11 from 2024. For each language-model
policy, these conditions contain 90 FinDVer episodes and 90 FinLongDocQA
episodes. Each problem instance uses the \OneRIUPerParticipant{} condition in
the two team conditions.

We also construct PIU distributions under the other conditions in
Figure~\ref{fig:piu-distribution-conditions}. For each distribution, the initial local
observation of each participant is checked for information that contains or
supports reconstruction of a PIU assigned to another participant.
Table~\ref{tab:finance-allocation-counts} reports the number of evaluated
problem instances admitting each PIU-distribution condition.

\begin{table}[!h]
\centering
\small
\setlength{\tabcolsep}{4pt}
\renewcommand{\arraystretch}{1.08}
\caption{\textbf{Finance problem-instance counts by PIU-distribution condition.}}
\label{tab:finance-allocation-counts}
\begin{tabular}{@{}
    >{\centering\arraybackslash}p{0.46\linewidth}
    >{\centering\arraybackslash}p{0.14\linewidth}
    >{\centering\arraybackslash}p{0.17\linewidth}
    @{}}
\toprule
PIU-distribution condition & FinDVer & FinLongDocQA \\
\midrule
\OneRIUPerParticipant{} & 30 & 30 \\
\PairwiseRIUOverlap{} & 26 & 23 \\
\ExtraParticipantWithoutRIU{} & 30 & 28 \\
\OneCommonRIU{} & 28 & 23 \\
\AllButOneRIUConcentration{} & 30 & 29 \\
\midrule
PIU distributions, total & 144 & 133 \\
\bottomrule
\end{tabular}
\end{table}

\subsubsection{Finance Problem-Instance and PIU-Distribution Statistics}
\label{app:finance-statistics}

\paragraph{Source-corpus and problem-instance counts.}
Each problem instance uses underlying source materials from one filing or
company-year report. Underlying source materials from different filings or
reports are not combined. Table~\ref{tab:finance-source-counts} separates the
available source-material pools from the processed and retained construction
outputs.

\begin{table}[!h]
\centering
\small
\setlength{\tabcolsep}{4pt}
\renewcommand{\arraystretch}{1.08}
\caption{\textbf{Finance source-corpus and problem-instance counts.}}
\label{tab:finance-source-counts}
\begin{tabular}{@{}
    >{\centering\arraybackslash}p{0.58\linewidth}
    >{\centering\arraybackslash}p{0.14\linewidth}
    >{\centering\arraybackslash}p{0.22\linewidth}
    @{}}
\toprule
Statistic & FinDVer & FinLongDocQA \\
\midrule
Source questions in the complete release & 2,400 & 7,527 \\
Source questions considered during construction & 700 & 7,527 \\
Available underlying source materials & 255 filings & 1,329 company-years \\
Constructed problem instances & 389 & 249 \\
\bottomrule
\end{tabular}
\end{table}

The finance evaluation uses a fixed subset of 30 constructed problem instances
from each source corpus, as described in
Appendix~\ref{app:finance-model-comparison-problems}.

\paragraph{Problem-instance and PIU-distribution statistics.}
Each problem instance requires a source-grounded written final decision. Source
questions are used only during construction; the constructed question does not
provide the corpus-provided answer or its label, explanation, or calculation.
It also omits the PIU identity and hidden evaluation annotations. The fixed
FinDVer evaluation set records 231 answer components across 70 source questions
and uses 20 Form 10-K filings and 10 Form 10-Q filings. The fixed FinLongDocQA
evaluation set records 64 answer components across 63 source questions and uses
9 company-year reports from 2022, 10 from 2023, and 11 from 2024. Across all 60
evaluated problem instances, every referenced source-material unit and PIU has
a recorded source location, and no participant initially receives every PIU.

\begin{figure}[!h]
    \centering
    \includegraphics[width=\textwidth]{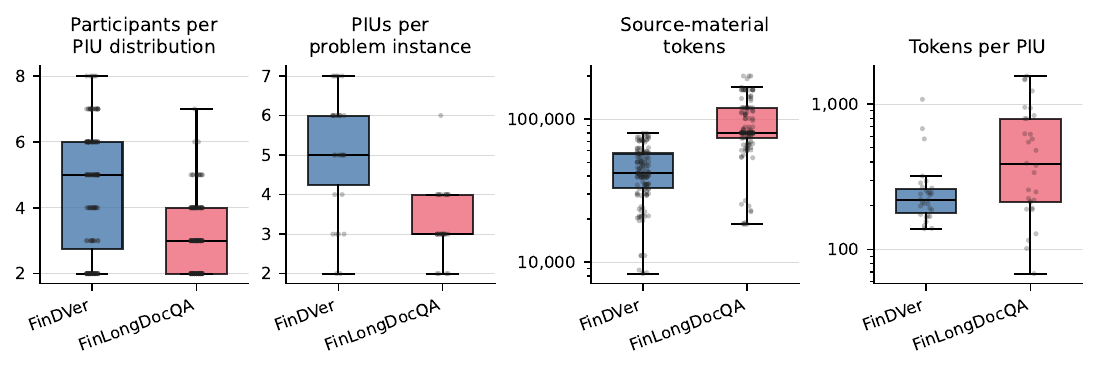}
    \caption{\textbf{Problem-instance and PIU-distribution statistics in the
    finance evaluation.} In the PIU-count and PIU-length panels, each point is
    one problem instance. In the participant-count and source-material-length
    panels, each point is one PIU distribution. Boxes show the median and
    interquartile range; token-length panels use logarithmic axes.}
    \label{fig:finance-riu-distribution}
\end{figure}

Figure~\ref{fig:finance-riu-distribution} reports the PIU count and PIU length
for the 60 problem instances in the finance evaluation. It also reports
participant count and the length of source material provided to participants
for the 277 PIU distributions under the five PIU-distribution conditions in
Figure~\ref{fig:piu-distribution-conditions}. For the PIU-count and PIU-length panels,
each point is one problem instance; for the participant-count and
source-material-length panels, each point is one PIU distribution.

Source-material length sums the material provided to each participant. Repeated
material is counted for each participant who receives it because it contributes
separately to that participant's initial local observation. The source-material
lengths are measured after the participant-level leakage-removal step and after
applying the 40,000-token initial-local-observation cap. PIU length is measured
from the PIU text supplied in Multi-Agent PIU-Only. All lengths use the same
\texttt{cl100k\_base} tokenizer used for the legal statistics.

\begin{table}[!h]
\centering
\scriptsize
\setlength{\tabcolsep}{2.5pt}
\renewcommand{\arraystretch}{1.08}
\caption{\textbf{Mean problem-instance and PIU-distribution statistics for the finance evaluation.} Problem-instance measures use the 60 fixed evaluation problem instances. PIU-distribution measures use 277 distributions. For measures that depend on PIU distribution, we first average within problem instance and then across problem instances. Source-material lengths reflect the material provided after participant-level leakage removal and the 40,000-token initial-local-observation cap.}
\label{tab:finance-problem-instance-structural-statistics}
\begin{tabular*}{\textwidth}{@{\extracolsep{\fill}}lrrrrrrr@{}}
\toprule
Source corpus & \makecell{Problem\\instances} & \makecell{PIU\\distributions} & \makecell{Participants/\\distribution} & PIUs & \makecell{Source-material tok./\\distribution} & \makecell{Source-material tok./\\participant} & \makecell{PIU tok./\\unit} \\
\midrule
FinDVer & 30 & 144 & 4.5 & 5.1 & 44,038.1 & 11,887.5 & 271.3 \\
FinLongDocQA & 30 & 133 & 3.1 & 3.3 & 91,711.8 & 31,168.6 & 546.0 \\
\midrule
All source corpora & 60 & 277 & 3.8 & 4.2 & 67,874.9 & 21,528.1 & 408.6 \\
\bottomrule
\end{tabular*}
\end{table}

\paragraph{Evaluation criteria.}
In both finance source corpora, a PIU may support several answer components,
and one answer component may require several PIUs. Written final decisions are
graded using problem-specific criteria for values, units, periods, signs,
scopes, calculations, and qualifying conclusions.

\subsubsection{FinDVer Problem-Instance Construction}

\paragraph{Source corpus and constructed problem instances.}
FinDVer provides entailment, numeric, and knowledge-oriented source questions
linked to SEC filings. Construction uses the FinDVer source corpus's
\texttt{testmini} split and
retains the source location for each Form 10-K or Form 10-Q filing. The
procedure constructs 389 problem instances. The finance
evaluation uses the fixed 30-instance subset described in
Appendix~\ref{app:finance-model-comparison-problems}.

\paragraph{Decision objective and PIU construction.}
Construction does not use the original FinDVer question as the decision
objective. Instead, it constructs a free-response filing-review task that asks
for related filing-specific findings, calculations, comparisons, or
classifications without providing the source values or expected conclusions.
The reference decision is a source-grounded written final decision. Hidden
evaluation annotations record the answer components, the links between answer
components and PIUs, and problem-specific evaluation criteria. For numeric
answer components, the criteria record the value, unit, sign, period, and
tolerance. For textual and classification answer components, the criteria
record the accepted conclusion and required scope.

Construction reconstructs each source annotation from complete filing
paragraphs or tables and records overlap, containment, and derived equivalence
among the candidate units before forming one filing-review problem instance.
The task-generation step selects complete candidate units. It prefers a broad
coherent cluster when distinct candidate units concern the same company,
period, transaction, segment, liquidity topic, accounting policy, operating
trend, or filing-review objective. Candidate units concerning unrelated
information are excluded, and a candidate unit may be retained with another
partially overlapping unit only when the task and evaluation criteria use a
distinct contribution from each. Every PIU must be needed for at least one
answer component, and every selected PIU is linked to the component or
reasoning step it supports.

\paragraph{Source material and PIU distribution.}
Complete filing material is distributed across participants. Surrounding
passages or tables that directly repeat, tabulate, or clearly paraphrase a PIU
assigned to another participant are removed. BM25 lexical scores provide
candidate pointers for review, after which a model-based check inspects the
complete packet and determines whether any context contains or supports
reconstruction of such a PIU. The procedure performs one
removal pass without redistributing the remaining source material. Every
assigned PIU remains in the corresponding participant's initial local
observation.

\paragraph{Participant descriptions.}
Participant descriptions are finance-specific reviewer labels. They may
describe a broad filing or report topic, but they do not disclose PIU values,
calculations, conclusions, or their effect on the reference decision.

\paragraph{FinDVer-specific construction checks.}
In addition to the common finance checks in
Appendix~\ref{app:finance-construction}, a model-based source-grounding check
verifies the calculations, comparisons, scopes, and links between PIUs and
answer components. Additional checks require the pivotal information to support a
complete and correct final decision, require public information without PIUs to
be insufficient, and reject candidate PIUs that are redundant under the
answer-component links. A problem instance is rejected if it fails these checks
or the common finance checks.

\subsubsection{FinLongDocQA Problem-Instance Construction}

\paragraph{Source corpus and constructed problem instances.}
Each FinLongDocQA source question provides its answer type and annotated page
numbers. Its reasoning and code are not used in construction. The procedure
constructs 249 problem instances. The
finance evaluation uses the fixed 30-instance subset described in
Appendix~\ref{app:finance-model-comparison-problems}.

\paragraph{Decision objective and PIU construction.}
Construction does not use a source question verbatim as the decision objective.
It constructs a free-response annual-report task that asks for a written memo,
calculation, explanation, or numbered review. The task may use one source
question or combine several source questions from the same company-year report.
When the selected source questions do not share one analytical purpose,
construction asks a concise multi-part review rather than inventing a causal
relationship or comparison. It records a reference decision, answer components,
links between PIUs and answer components, evaluation criteria, and errors that
make a final decision incorrect. Written final decisions are evaluated using
the recorded problem-specific evaluation criteria.

For each source question, construction uses the annotated pages to identify
every paragraph, table, or combined adjacent label and value needed for the answer's
values, definitions, periods, units, scope, and exceptions. These candidate
units then follow the division and overlap requirements in
Appendix~\ref{app:finance-construction}. Each selected evidence unit is one
complete mapped block. A block is not duplicated into multiple PIUs for
separate values, rows, or claims. Every selected PIU must support at least one
report-specific claim, and removing that PIU must leave at least one linked
claim not fully grounded even when all other PIUs remain available. A PIU may
support several claims, and one claim may require several PIUs.

\paragraph{Source material and PIU distribution.}
Each PIU is provided to the participant assigned that PIU. Other report
passages and tables are provided to the participant with the shortest initial
local observation, using distance from the page containing that participant's
PIU to resolve a tie. A model-based check determines whether one passage or
table, or several in combination, supports reconstruction of a PIU assigned
to another participant. The source material is revised until every PIU is
preserved and such information is removed.

When unrelated rows of a large table independently state a PIU assigned to
another participant, the table is restricted to the rows supporting the answer
component associated with the PIU assigned to that participant.

\paragraph{Participant descriptions.}
Participant descriptions are finance-specific reviewer labels. They may
describe a broad filing or report topic, such as liquidity and capital
resources or deferred tax assets and liabilities, but they do not disclose PIU
values, calculations, conclusions, or their effect on the reference decision.

\paragraph{FinLongDocQA-specific construction checks.}
In addition to the common finance checks in
Appendix~\ref{app:finance-construction}, a problem instance is rejected if a
participant does not receive a PIU assigned to that participant or if source
material still supports reconstruction of a PIU assigned to another
participant. Additional checks require the reference answer to satisfy every
recorded claim, preserve material units and periods, and use no outside facts.
A problem instance is rejected if it fails these checks or the common finance
checks.

\subsection{Medical}
\label{app:medical-construction}

We construct medical problem instances from MLCR
\citep{wisedocs2026mlcrbenchmark} and EvidenceBench
\citep{wang2025evidencebench}. Each source corpus corresponds to one problem
family: MLCR medical-record reasoning and EvidenceBench biomedical-evidence
reasoning. The procedure below reports the choices specific to these source
corpora and problem families. Table~\ref{tab:medical-source-corpus-summary}
summarizes the medical problem families and their source corpora.

\begin{table}[!h]
\centering
\small
\setlength{\tabcolsep}{4pt}
\renewcommand{\arraystretch}{1.03}
\caption{\textbf{Medical problem families and source corpora.}}
\label{tab:medical-source-corpus-summary}
\begin{tabular}{@{}
    >{\centering\arraybackslash}m{0.18\textwidth}
    >{\centering\arraybackslash}m{0.34\textwidth}
    >{\centering\arraybackslash}m{0.22\textwidth}
    >{\centering\arraybackslash}m{0.20\textwidth}
    @{}}
\toprule
Source corpus &
Underlying source materials &
Reference decision &
Problem family \\
\midrule
MLCR & Case-level visit-summary documents with source questions and answers & Written medical-record review & Medical-record reasoning \\
EvidenceBench & Biomedical-paper sentence pools with hypotheses and evidence annotations & Written biomedical-evidence answer & Biomedical-evidence reasoning \\
\bottomrule
\end{tabular}
\end{table}

\subsubsection{Common Medical Construction Procedure}
\label{app:medical-common-construction}

\paragraph{Step 1. Select the underlying source materials.}
A problem instance uses source materials from one medical source record. For
EvidenceBench, this is one biomedical paper represented as an ordered sentence
pool. For MLCR, this is one case whose source material may contain many visit
summary documents. Source materials from different papers or cases are not
combined.

\paragraph{Step 2. Divide the underlying source materials.}
EvidenceBench source materials are divided into paper sentences, preserving
sentence indices and sentence types. MLCR source materials are divided into
exact sentence-like atoms and paragraph spans with raw character offsets,
document identifiers, and checksums. Adjacent atoms or sentences may be
combined only when needed to preserve a patient, date, encounter, negation,
measurement unit, uncertainty, abbreviation, comparison, or clinical scope.

\paragraph{Step 3. Identify candidate units.}
Construction uses source annotations to identify candidate evidence units.
EvidenceBench starts from annotated optimal-evidence anchors. MLCR first maps
source question-answer pairs to exact source atoms in the case record. In both
families, candidate units are source-faithful evidence spans rather than final
PIUs: they become PIUs only after task construction and the boundary checks
show that they are answer-relevant and necessary.

\paragraph{Step 4. Specify the decision objective and format.}
Construction creates a free-response medical task with a written reference
decision and a hidden claim-level rubric. The public task does not reveal PIU
identities, candidate-unit identifiers, source annotation labels, hidden rubric
claims, reference decisions, or expected values and conclusions. Required
claims record medical scope, patient or study population, dates or periods,
comparisons, outcomes, direction, uncertainty, units, tolerances, and causal or
associational strength when applicable.

\paragraph{Step 5. Identify the pivotal information.}
We apply the PIU construction requirements in
Section~\ref{sec:real-source-problems}. The pivotal information must support a
complete and grounded answer to the public task, while a missing PIU must leave
at least one required claim unsupported or incorrect. A model-based solver and
separate judge check the all-PIU condition, a no-PIU condition, and
leave-one-out conditions. The solver sees only the public task and displayed
source material; the judge uses the hidden rubric to evaluate claim correctness
and grounding. Correct guesses that are not grounded in displayed material do
not satisfy the grounding check.

\paragraph{Step 6. Define the PIU distribution and distribute source material.}
Each medical problem instance uses the \OneRIUPerParticipant{} condition in the
two team evaluation conditions. The additional PIU distributions reported in
Table~\ref{tab:medical-allocation-counts} use the other conditions from
Figure~\ref{fig:piu-distribution-conditions}. After each PIU distribution is fixed,
source material is distributed across participants. The material provided to
each participant is checked for information that contains or supports
reconstruction of a PIU assigned to another participant. Such material is
removed while every PIU assigned to that participant is retained.

\paragraph{Step 7. Complete problem-instance construction.}
A problem instance is retained only after the public task, pivotal information, PIU
distribution, reference decision, hidden rubric, and distributed source material
pass the applicable checks.

\subsubsection{Problem Instances and PIU Distributions in the Medical Evaluation}
\label{app:medical-model-comparison-problems}

{The final medical evaluation uses 29 MLCR problem instances
and 45 EvidenceBench problem instances under the three evaluation
conditions. For each language-model policy, these conditions contain
87 MLCR episodes and 135 EvidenceBench episodes.}

We also construct PIU distributions under the other conditions in
Figure~\ref{fig:piu-distribution-conditions}. For each distribution, the initial local
observation of each participant is checked for information that contains or
supports reconstruction of a PIU assigned to another participant.
Table~\ref{tab:medical-allocation-counts} reports the number of evaluated
problem instances admitting each PIU-distribution condition.

\begin{table}[!h]
\centering
\small
\setlength{\tabcolsep}{4pt}
\renewcommand{\arraystretch}{1.08}
\caption{\textbf{Medical problem-instance counts by PIU-distribution condition.}}
\label{tab:medical-allocation-counts}
\begin{tabular}{@{}
    >{\centering\arraybackslash}p{0.46\linewidth}
    >{\centering\arraybackslash}p{0.14\linewidth}
    >{\centering\arraybackslash}p{0.17\linewidth}
    @{}}
\toprule
PIU-distribution condition & MLCR & EvidenceBench \\
\midrule
\OneRIUPerParticipant{} & 29 & 45 \\
\PairwiseRIUOverlap{} & 19 & 36 \\
\ExtraParticipantWithoutRIU{} & 28 & 45 \\
\OneCommonRIU{} & 19 & 36 \\
\AllButOneRIUConcentration{} & 19 & 45 \\
\midrule
PIU distributions, total & 114 & 207 \\
\bottomrule
\end{tabular}
\end{table}

\subsubsection{Medical Problem-Instance and PIU-Distribution Statistics}
\label{app:medical-statistics}

\paragraph{Source-corpus and problem-instance counts.}
Each problem instance uses one EvidenceBench paper or one MLCR case. For MLCR,
the case may contain many underlying visit-summary source documents, but cases
are not combined. Table~\ref{tab:medical-source-counts} separates the available
source-material pools from the fixed evaluation problem instances.

\begin{table}[!h]
\centering
\small
\setlength{\tabcolsep}{4pt}
\renewcommand{\arraystretch}{1.08}
\caption{\textbf{Medical source-corpus and problem-instance counts.}}
\label{tab:medical-source-counts}
\begin{tabular}{@{}
    >{\centering\arraybackslash}m{0.54\linewidth}
    >{\centering\arraybackslash}m{0.14\linewidth}
    >{\centering\arraybackslash}m{0.17\linewidth}
    @{}}
\toprule
Statistic & MLCR & EvidenceBench \\
\midrule
Source questions or hypotheses in the construction pool & 205 source questions & 293 test-split hypotheses \\
Available underlying source materials & 1,071 visit-summary documents & 293 biomedical-paper sentence pools \\
Evaluated problem instances & 29 & 45 \\
Evaluated PIU distributions & 114 & 207 \\
\bottomrule
\end{tabular}
\end{table}

\paragraph{Problem-instance and PIU-distribution statistics.}
Each medical problem instance requires a source-grounded written final decision.
The constructed public question omits the source-provided answer, expected
conclusion, PIU identities, and hidden claim rubric. Across the 74 evaluated
problem instances, every PIU has a source location, and no participant
initially receives every PIU in the team evaluation conditions.

\begin{figure}[!h]
    \centering
    \includegraphics[width=\textwidth]{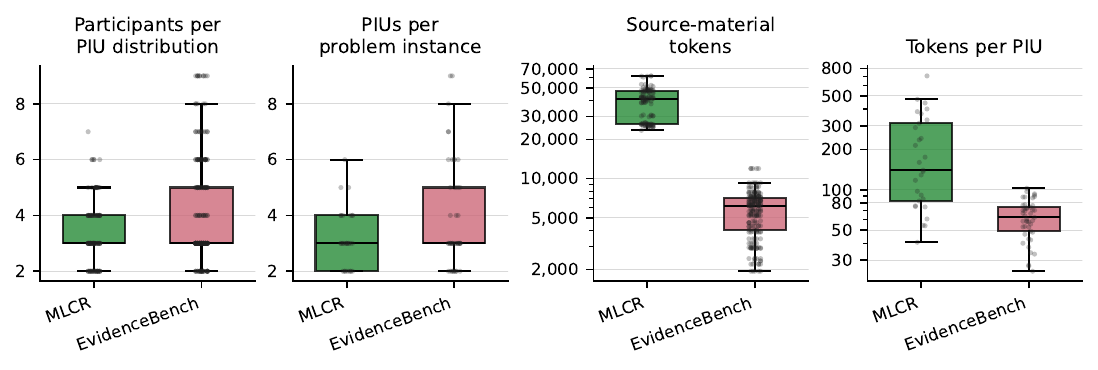}
    \caption{\textbf{Problem-instance and PIU-distribution statistics in the
    medical evaluation.} In the PIU-count and PIU-length panels, each point is
    one problem instance. In the participant-count and source-material-length
    panels, each point is one PIU distribution. Boxes show the median and
    interquartile range; token-length panels use logarithmic axes.}
    \label{fig:medical-riu-distribution}
\end{figure}

Figure~\ref{fig:medical-riu-distribution} reports the PIU count and
PIU length for the 74 problem instances in the final medical evaluation. It
also reports
participant count and the length of source material provided to participants
for the 321 PIU distributions under the five PIU-distribution conditions in
Figure~\ref{fig:piu-distribution-conditions}. For the PIU-count and PIU-length panels,
each point is one problem instance; for the participant-count and
source-material-length panels, each point is one PIU distribution.

Source-material length sums the material provided to each participant. Repeated
material is counted for each participant who receives it because it contributes
separately to that participant's initial local observation. The source-material
lengths are measured after the participant-level leakage-removal step and after
applying the 40,000-token initial-local-observation cap. PIU length is measured
from the PIU text supplied in Multi-Agent PIU-Only. All lengths use the same
\texttt{cl100k\_base} tokenizer used for the legal statistics.

\begin{table}[!h]
\centering
\scriptsize
\setlength{\tabcolsep}{2.5pt}
\renewcommand{\arraystretch}{1.08}
\caption{\textbf{Mean problem-instance and PIU-distribution statistics for the medical evaluation.} Problem-instance measures use the 74 fixed evaluation problem instances. PIU-distribution measures use 321 distributions. For measures that depend on PIU distribution, we first average within problem instance and then across problem instances. Source-material lengths reflect the material provided after participant-level leakage removal and the 40,000-token initial-local-observation cap.}
\label{tab:medical-problem-instance-structural-statistics}
\begin{tabular*}{\textwidth}{@{\extracolsep{\fill}}lrrrrrrr@{}}
\toprule
Source corpus & \makecell{Problem\\instances} & \makecell{PIU\\distributions} & \makecell{Participants/\\distribution} & PIUs & \makecell{Source-material tok./\\distribution} & \makecell{Source-material tok./\\participant} & \makecell{PIU tok./\\unit} \\
\midrule
MLCR & 29 & 114 & 3.2 & 3.2 & 38,119.9 & 13,080.7 & 208.5 \\
EvidenceBench & 45 & 207 & 4.0 & 4.0 & 5,678.0 & 1,727.4 & 62.7 \\
\midrule
All source corpora & 74 & 321 & 3.7 & 3.7 & 18,391.7 & 6,176.7 & 119.8 \\
\bottomrule
\end{tabular*}
\end{table}

\paragraph{Evaluation criteria.}
In both medical source corpora, a PIU may support several required claims, and
one required claim may require several PIUs. Written final decisions are graded
using problem-specific criteria for clinical scope, study or patient
population, dates or periods, measurements, units, direction, uncertainty,
comparison, and causal or associational strength.

\subsubsection{MLCR Problem-Instance Construction}

\paragraph{Source corpus and retained problem instances.}
MLCR provides long medical-record cases with source questions, reference
answers, and visit-summary source documents \citep{wisedocs2026mlcrbenchmark}.
The evaluated construction uses 29 problem instances. Because an MLCR case may
contain many visit summaries, a single problem instance can draw source
material from multiple source documents within the same case, but it never
combines source material across cases.

\paragraph{Decision objective, reference decision, and evaluation criteria.}
Construction first maps each eligible source question-answer pair to exact
source atoms that materially reconstruct the answer. It then groups one or
more source question-answer pairs from the same case into a clinically coherent
free-response task. When selected source questions do not share one clinical or
claims-review purpose, construction uses a concise multi-part review rather
than inventing a causal relationship or common theme. The public task asks for
the medical-record review and does not reveal source question identifiers,
candidate identifiers, PIU identifiers, or case-specific expected values and
conclusions.

The hidden rubric records required claims, the selected evidence units that
support each claim, and the criteria for clinical scope, dates, encounters,
measurements, units, direction, negation, uncertainty, and causal strength. The
reference decision must satisfy all required claims using only the selected
source evidence. Generic medical knowledge may connect observations but cannot
replace missing patient-specific evidence.

\paragraph{Candidate-unit and PIU construction.}
MLCR source documents are divided into exact source atoms while preserving raw
offsets and document identifiers. Each candidate unit is a contiguous span of
atoms from one source document. Adjacent atoms are included only when needed to
preserve patient, date, encounter, negation, measurement unit, uncertainty, or
clinical scope. Non-adjacent evidence and repeated equivalents are kept
separate rather than merged.

A model-based grounding check verifies that every required claim and every
material statement in the reference decision are grounded in the selected exact
source evidence, and that each selected unit makes a unique necessary
contribution. Boundary checks then test the complete selected set, a no-PIU
condition, and leave-one-out conditions. A selected candidate becomes a PIU
only after these checks show that the full set is sufficient and that removing
the candidate breaks at least one claim it supports.

\paragraph{Source material and PIU distribution.}
MLCR source material is distributed across participants under the PIU
distribution. The participant assigned a PIU receives the canonical source span
for that PIU. Additional source atoms from the same case are assigned to
participants subject to the 40,000-token local-observation budget. Multi-Agent
PIU-Only presents only the assigned PIU text. Multi-Agent Full-Context presents
the participant's medical-record source material with neutral source-file
structure. Single-Agent All-PIU presents all pivotal information. All three
conditions use the same public task, and hidden source mappings, candidate
identifiers, PIU identifiers, rubric mappings, and reference answers are not
shown to decision makers.

\paragraph{MLCR-specific construction checks.}
For each participant and each PIU assigned to another participant, a
model-based packet-leakage check determines whether the participant's complete
full-context packet directly contains, duplicates, paraphrases, or permits
reconstruction of that foreign PIU, including by combining multiple packet
passages. Source material containing the participant's assigned PIU is not
removed. If noncanonical source material reconstructs a foreign PIU, removable
contexts are deleted and the packet is checked again. If the participant's own
canonical source span itself reveals a foreign PIU, construction records this
as canonical overlap and either merges ownership as allowed by the admissible
allocation or rejects the allocation when strict one-each ownership is
required. A problem instance or allocation is rejected if leakage remains
without removable source material or the check does not converge.

\subsubsection{EvidenceBench Problem-Instance Construction}

\paragraph{Source corpus and retained problem instances.}
EvidenceBench provides biomedical hypotheses, full papers represented as
ordered sentence pools, sentence types, and sentence-level annotations of the
minimum evidence required for each hypothesis \citep{wang2025evidencebench}.
We use the test split. Forty-five problem instances pass the construction
checks, and all 45 are used in the medical evaluation.

We treat the sentences identified by these annotations as source locations
used during construction rather than as PIUs. For every source record,
construction records the instance and paper identifiers, original hypothesis,
ordered sentence indices, sentence types, paper-sentence count, split,
and source-file hash as hidden source annotations. Up to two neighboring
sentences may be added only to restore an incomplete referent, comparison,
abbreviation, or measurement context; the complete neighborhood is not
automatically included in a candidate unit.

\paragraph{Decision objective, reference decision, and evaluation criteria.}
Construction turns the source hypothesis and grounded evidence annotations
into a biomedical question, a reference decision, and separately evaluated
answer components. The decision objective omits the annotated sentence indices,
PIU identities, candidate-unit identifiers, reference decision, and hidden
construction information. The hypothesis guides the question but is not
necessarily reused verbatim: the question may instead ask for the population,
exposure or intervention, comparator, outcome, direction, evidentiary strength,
or required limitation established by the paper evidence. The reference
decision uses only selected exact evidence and preserves medical scope and the
source's causal or associational strength.

Answer components and candidate units are distinct. An answer component may
require several candidate units that make complementary contributions, and one
candidate unit may support several answer components. The hidden evaluation
criteria therefore record the links between answer components and candidate
units. For some answer components, every linked candidate unit is
required; for others, any one of several linked candidate units is sufficient.
The recorded links state which part of an answer component each candidate unit
supports. Construction also records, for every candidate unit, whether it is
retained and, if it is excluded, the reason for exclusion.

\paragraph{Candidate-unit construction and checks.}
Construction preserves a one-to-one relation between the sentences identified
by the source annotations and candidate units. Each such sentence produces
exactly one candidate unit, and distinct source sentences are not
merged even when they jointly support the same answer component. The candidate
unit contains the source sentence verbatim and may add only the minimal
contiguous neighboring context needed to make it interpretable. It may not
include another sentence identified by the source annotations as context.
Construction checks complete one-to-one coverage of the source annotations,
unique identifiers, contiguous and nonoverlapping sentence spans, exact
agreement with the paper text, and absence of another sentence identified by
the source annotations. Relations among candidate units are represented by
their links to answer components rather than by merging their source locations.

A separate model-based check receives the decision objective, reference
decision, hidden evaluation criteria, and the complete candidate-unit set,
including excluded candidate units. For each retained candidate unit, it checks
relevance to the decision objective, direct support, and necessity under the
recorded links to answer components; for each excluded candidate unit, it
checks that exclusion is justified. It also checks that the decision objective
is natural rather than artificially expanded to accommodate unrelated evidence,
that every answer component and the reference decision are source-grounded, and
that population, comparison, outcome, direction, uncertainty, and causal scope
are preserved. A failed check permits bounded revision of the decision
objective, answer components, and links while the exact candidate units remain
fixed. At least two annotated source sentences and two retained candidate units
are required, but construction imposes no fixed upper bound.

\paragraph{PIU construction and checks.}
Construction then applies the model-based removal check described in
Section~\ref{sec:real-source-problems}, using a model and a separate model
judge. The model receives only the decision objective, the
paper excerpts available in the current check, and neutral evidence numbers. It
does not receive candidate-unit, PIU, annotated-sentence, answer-component, or
omitted-unit identifiers. The model judge receives the hidden evaluation
criteria and reference decision and separately records whether each answer
component is correct and grounded in the material provided to the model. A
correct guess is not treated as grounded.

With all retained candidate units, every answer component must be correct and
grounded, the final decision must be complete, and it must contain no critical
error. For each removal required by Section~\ref{sec:real-source-problems}, at
least one answer component linked to the removed candidate unit must cease to
be satisfied. The recorded links are also checked against the source references in
the model's response: every linked unit
is required when the answer component requires all of them, whereas at least
one linked unit is required when any one is sufficient. After every check
passes, each retained candidate unit forms one PIU; candidate units that jointly
support one answer component remain separate PIUs.

\paragraph{Source material and PIU distribution.}
The retained construction uses \OneRIUPerParticipant{}. Each source sentence
containing a PIU is reserved for the participant assigned that PIU. Overlap
between the source sentences containing distinct PIUs is rejected. Every
remaining sentence from the EvidenceBench paper sentence pool is then assigned
to exactly one participant, with seeded assignment balancing the estimated
lengths of the initial local observations; the resulting division is exhaustive
and disjoint. Multi-Agent PIU-Only presents only the exact text of the assigned
PIU. Multi-Agent Full-Context presents the participant's paper sentences with
neutral source identifiers. Single-Agent All-PIU presents all pivotal information.
All three conditions use the same decision objective, and hidden candidate-unit
identifiers, PIU identifiers, annotated source locations, links to answer
components, reference source locations, and source-material check information
are removed from the decision makers' observations.

\paragraph{EvidenceBench-specific construction checks.}
For each participant and each PIU assigned to another participant, a model-based
check considers the participant's complete initial local observation rather
than isolated lexical matches. The initial local observation contains a PIU
assigned to another participant only when an exact duplicate, clear paraphrase,
equivalent reported result, or combination of source passages supports that
PIU's complete contribution at the same medical scope. Shared topics, partial
overlap in population or outcome, a different measurement, a general
mechanistic association, or a plausible inference requiring external knowledge
do not suffice.

When such reconstruction is possible, the removable additional source passages
that support the reconstruction are deleted, and every participant and PIU
assigned to another participant are checked again. Source material containing
the participant's assigned PIU is never deleted. If that material alone
supports a PIU assigned to another participant, the problem instance is
rejected because a candidate unit is incorrectly defined. The procedure permits at most
three revisions and rejects the problem instance if the initial local
observation still contains another participant's PIU, no removable source
material exists, final-decision validation fails, or the checks do not
converge.

\subsection{Software Engineering}
\label{app:swe-construction}

We construct software-engineering problem instances from SWE Atlas Codebase
QnA \citep{raghavendra2026sweatlas} and DeepSWE
\citep{huang2026deepswe}. Table~\ref{tab:swe-source-status}
summarizes the source corpora, problem families, evaluation criteria, and
evaluation scope. The two source corpora use separate
construction procedures, described in the following subsections.

\subsubsection{Source Corpora and Evaluation Scope}

\begin{table}[!h]
\centering
\small
\setlength{\tabcolsep}{4pt}
\caption{\textbf{Software-engineering source corpus and evaluation status.}}
\label{tab:swe-source-status}
\begin{tabular}{@{}
    >{\centering\arraybackslash}m{0.20\linewidth}
    >{\centering\arraybackslash}m{0.27\linewidth}
    >{\centering\arraybackslash}m{0.31\linewidth}
    >{\centering\arraybackslash}m{0.12\linewidth}
    @{}}
\toprule
Source corpus & Problem family & Evaluation criteria & \makecell{Evaluated\\problem\\instances} \\
\midrule
SWE Atlas Codebase QnA & Multi-file and runtime code comprehension & Every answer component is complete and correct & 32 \\
DeepSWE & Long-horizon repository-level code repair & Official repository verifier passes & 12 \\
\bottomrule
\end{tabular}
\end{table}

\paragraph{Shared construction conventions.}
Both source corpora fix a repository and base commit, identify exact
source-grounded PIUs, assign each PIU to one participant in collaborative
conditions, and withhold reference answers and hidden qualification records
from the evaluated agents. Source-specific procedures determine how code,
tests, runtime observations, and additional context are collected, grouped,
distributed, and checked for leakage and format salience.

The SWE Atlas evaluation uses all 32 retained instances and eight
language-model policies across the three evaluation conditions. For DeepSWE,
we archive intermediate construction and allocation artifacts for 21 screened
candidate tasks, then complete construction for a frozen 12-task cohort selected
to preserve programming-language coverage and variation in original benchmark
difficulty. These 12 tasks constitute both the constructed and evaluated
DeepSWE problem instances reported in Table~\ref{tab:problem-instance-counts}.
The source-specific construction and evaluation procedures are described below.

\subsubsection{Software-Engineering Problem-Instance Statistics}

\begin{figure}[!h]
    \centering
    \includegraphics[width=\textwidth]{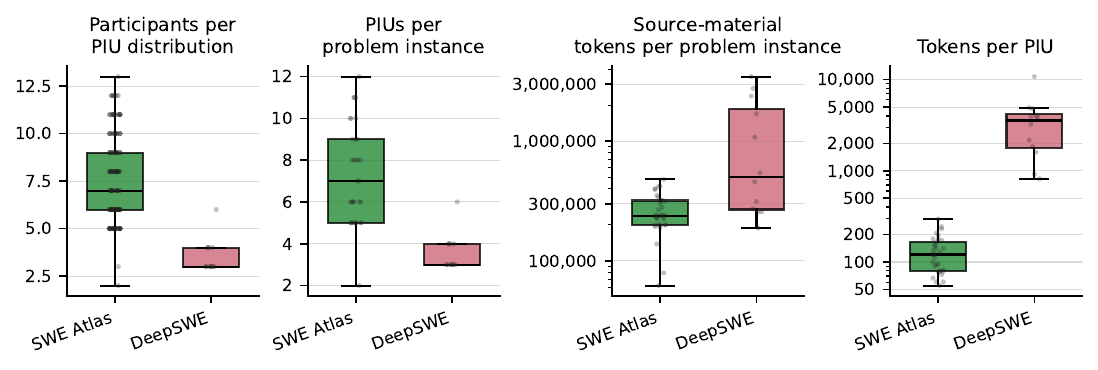}
    \caption{\textbf{Software-engineering problem-instance and PIU-distribution statistics.}
    In the PIU-count and PIU-length panels, each point is one problem instance.
    In the participant-count panel, each point is one PIU distribution; in the
    source-material-length panel, each point is one problem instance. SWE Atlas
    includes its five valid allocation modes; DeepSWE includes the 12 selected
    functional one-each allocations. Boxes show the median and interquartile
    range; token-length panels use logarithmic axes.}
    \label{fig:swe-problem-instance-structural-statistics}
\end{figure}

Figure~\ref{fig:swe-problem-instance-structural-statistics} and
Table~\ref{tab:swe-problem-instance-structural-statistics} report structural
statistics for both software-engineering source corpora. For SWE Atlas, the
problem-instance measures use the 32 one-each instances, while distribution
measures use the 157 valid distributions across the five allocation modes. For
DeepSWE, both sets of measures use the fixed 12-task evaluation cohort and its
functional one-each allocations. Tasks outside this fixed cohort are not
included in these descriptive statistics.

The code/source-unit column counts Atlas context units and DeepSWE code
fragments. The runtime-observation column counts explicitly recorded execution
observations. PIU length is measured from the evidence assigned to each PIU.
Full source-material length uses the underlying Atlas source units or the
DeepSWE code archive, observations, and public-check catalog. The distribution
column reports assigned participant packets; for DeepSWE these are the
available PIU evidence packets before finalized full-context distractor
packing. All lengths use the \texttt{cl100k\_base} tokenizer.

\begin{table}[!h]
\centering
\scriptsize
\setlength{\tabcolsep}{3pt}
\renewcommand{\arraystretch}{1.08}
\caption{\textbf{Software-engineering structural statistics.}}
\label{tab:swe-problem-instance-structural-statistics}
\textbf{Problem-instance measures.}\par\smallskip
\begin{tabular*}{\textwidth}{@{\extracolsep{\fill}}lrrrrrr@{}}
\toprule
Source corpus & \makecell{Problem\\instances} & \makecell{PIUs/\\instance} & \makecell{Code/source\\units/instance} & \makecell{Runtime\\observations/instance} & \makecell{Full source-material\\tok./instance} & \makecell{PIU tok./\\unit} \\
\midrule
SWE Atlas & 32 & 7.3 & 236.5 & 0.0 & 261,951.8 & 130.2 \\
DeepSWE & 12 & 3.6 & 16.8 & 7.2 & 1,140,635.7 & 3570.6 \\
\bottomrule
\end{tabular*}
\par\medskip
\textbf{PIU-distribution measures.}\par\smallskip
\begin{tabular*}{\textwidth}{@{\extracolsep{\fill}}lrrr@{}}
\toprule
Source corpus & \makecell{PIU\\distributions} & \makecell{Participants/\\distribution} & \makecell{Source-material tok./\\distribution} \\
\midrule
SWE Atlas & 157 & 7.6 & 245,444.0 \\
DeepSWE & 12 & 3.6 & 59,474.2 \\
\bottomrule
\end{tabular*}
\end{table}

\subsubsection{SWE Atlas Codebase QnA Problem-Instance Construction}
\label{app:swe-atlas-conversion}

\paragraph{Source corpus and decision objective.}
SWE Atlas Codebase QnA asks for a source-grounded technical decision about a
repository fixed at a recorded base commit. Construction loads the task
question, positive and negative evaluation criteria, corpus-provided answer,
Docker image, and base commit. The reference decision and evaluation criteria
are used for hidden construction checks but are not shown to the agents that
search the repository or receive the released evidence. Thirty-two of the 124
official entries pass the completed construction procedure and are retained.

\paragraph{Evidence collection and candidate units.}
Construction collects exact source, test, configuration, documentation, and,
when a criterion requires observed behavior, runtime evidence from the
unchanged Docker checkout. Runtime commands use the installed repository tools,
do not use the network or install packages, and record their output, exit
status, commit, and repository state. A runtime observation is included only
with the source or test evidence needed to interpret it; runtime output alone
does not form a PIU.

Candidate units are exact contiguous source-material spans that establish one
answer contribution. Source, test, and runtime spans are grouped when they
jointly establish one indivisible proposition; duplicates and fully contained
contributions are merged, while independent contributions remain separate.
Navigation results and model-written summaries are excluded. The construction
does not target a fixed PIU count and retains at least two PIUs per
collaborative problem instance.

\paragraph{Grounding, sufficiency, and necessity checks.}
Construction evaluates the complete candidate set, the decision objective with
no candidate units, and each leave-one-out removal. The solver receives only
the exact evidence available in the current check, while a separate judge
assesses correctness and source grounding. A candidate is retained only when
the complete set supports every required positive and negative criterion, the
no-unit condition fails, and removing that candidate causes the linked answer
component to fail. Candidate grouping or evidence collection is revised and
the checks are repeated when these conditions do not hold.

\paragraph{PIU distribution and evaluation conditions.}
The evaluated distribution uses \OneRIUPerParticipant{}. Source or test units
and runtime output containing a PIU are reserved for its owner; additional
context is assigned without duplication unless the allocation condition
explicitly permits overlap. Multi-Agent PIU-Only provides the assigned
PIU evidence, Multi-Agent Full-Context adds neutral additional repository
material and actual runtime output, and Single-Agent All-PIU provides all
pivotal information in a fixed order. The common observation identifies the
task, repository, base commit, and response format without exposing PIU
ownership or reference answers.

\paragraph{Atlas-specific leakage and salience checks.}
For every participant, a model-based check determines whether the initial local
observation contains or supports a PIU assigned to another participant,
including information reconstructable by combining multiple passages.
Topical overlap alone does not count. Revisions may remove or reassign
additional context but may not remove the owner's PIU evidence. The procedure
also checks that PIU-bearing material is not identifiable from position,
wrapper, annotation, or conspicuous length, and excludes instances that still
contain foreign PIU information after bounded revision.

\subsubsection{DeepSWE Problem-Instance Construction}
\label{app:deepswe-construction}

DeepSWE provides long-horizon repository-level code-repair tasks, each defined
by a pinned base repository, a public task instruction, and an official hidden
verifier. A task may admit multiple verifier-passing patches, and repeated
attempts by the same model may follow different search and implementation
paths. The original DeepSWE evaluation accounts for this variation through
repeated rollouts: its analysis uses three independent attempts by the same
model on each task, while its final evaluation uses approximately four
rollouts per task and model configuration. Successful GPT-5.5 repair
trajectories take roughly 80 agent steps on average
\citep{huang2026deepswe}.

Our goal is not to reproduce this unrestricted long-horizon search, but to
control the evidence required for collaboration under a fixed execution
budget. We therefore define a DeepSWE PIU as a unit of functional evidence
needed to support a successful repair under a fixed, bounded protocol. The
complete PIU set is constructed so that, together with the public task
instruction, it supports a verifier-passing patch using a restricted action
space and five atomic edit turns.

\paragraph{Instruction-only screening.}
We begin with 40 DeepSWE tasks. For each task, the model receives the public
task instruction but no PIUs or additional code evidence. We give the model
the same five-turn repair budget used in the subsequent evidence-bundle checks
and run three independent attempts. We continue PIU construction only for
tasks that fail all three instruction-only attempts. This screen removes tasks
whose instructions already provide enough information to locate and implement
a repair within the bounded protocol. The instruction-only runs therefore
serve as the no-evidence, or ``none,'' condition for construction.

\paragraph{Constructor-only materials and evidence collection.}
For each candidate task, the constructor may inspect the reference patch,
hidden-test patch and identifiers, public-check definitions and results, and
the pinned base repository to locate implementation boundaries and audit
evidence coverage. These materials are construction-only: the reference patch,
hidden tests, verifier internals, hidden expected values, and constructor
rationales are never copied into solver-visible evidence or exposed to
evaluated agents. The public task instruction is common input to every
condition and is not itself a PIU.

Construction collects exact task-relevant fragments from the unmodified base
commit and deterministic runtime observations produced from the same checkout.
Each observation records the executed probe, its output and exit status, and
the repository state needed to interpret it. Observations may describe public
base-repository behavior but do not reveal hidden assertions, hidden expected
values, or solution code. Solver-visible evidence is therefore composed of
base-commit code and reproducible base-repository behavior, not pieces of the
reference patch.

\paragraph{Functional PIU construction.}
A DeepSWE task may admit multiple implementations that pass the official
verifier. Searching for a minimal evidence set relative to one reference patch
could therefore steer the agent toward only one implementation path despite
the existence of other valid repairs. We instead define each DeepSWE PIU to
represent one coherent functional behavior or contract required by the
verifier, rather than a fragment of a particular reference implementation.

A functional PIU may combine several exact code fragments with deterministic
runtime observations when they jointly specify an implementation boundary,
such as parsing and validation, state accumulation, lifecycle and cleanup,
persistence, public-API wiring, or error propagation. Supporting types,
callers, interfaces, and exports are included when they are required to
implement that behavior. Each selected code fragment or observation is
assigned to exactly one PIU. Functional labels, evidence-selection rationales,
and hidden failure information remain constructor metadata and are not included
in participant packets.

The purpose of these PIUs is not to reproduce the reference patch. Each PIU is
grounded in base-repository code and reproducible runtime behavior, and the
complete PIU set must support a verifier-passing repair under the bounded
protocol.

\paragraph{Bounded evidence-bundle checks.}
We inspect candidate PIU bundles with the same source-restricted repair
interface used in evaluation. The patch author has five atomic edit turns and
may invoke checks from a frozen public-check catalog. Repository mutation is
restricted to \texttt{replace}, \texttt{write}, and \texttt{delete}
operations, and the agent may terminate before using all five turns. After
termination, the official hidden verifier is run once. Success is determined
solely by whether the final patch passes this verifier; matching the reference
patch is not required.

Each complete PIU bundle is evaluated in three independent qualification runs.
These runs form the complete-evidence, or ``all,'' condition for construction.
When all three instruction-only runs fail but the complete bundle enables a
successful repair within the same five-turn protocol, the bundle provides
operational evidence of sufficiency under that bounded protocol.

Repeated qualification outcomes and targeted leave-one-PIU-out checks are
used diagnostically to refine the evidence bundle. We inspect whether removing
a PIU reduces repair success or removes a functional boundary needed for the
repair, and use the result to identify missing, irrelevant, or redundant
evidence. Because repair trajectories are stochastic and multiple patches may
pass the verifier, these leave-one-out checks are not treated as a formal proof
of necessity over every possible repair procedure. They diagnose the
contribution of each functional PIU under the fixed five-turn protocol.

\paragraph{Final evaluation cohort.}
After screening, evidence-bundle construction, and qualification, we freeze a
12-task evaluation cohort from the initial 40 tasks. The cohort contains four
Go, four Python, two JavaScript, and two TypeScript tasks and preserves both
programming-language coverage and variation in original benchmark difficulty.
These 12 tasks are the constructed and evaluated DeepSWE problem instances
reported in Table~\ref{tab:problem-instance-counts}.

\paragraph{PIU allocation and evaluation conditions.}
The final functional PIUs are assigned one per participant. In Single-Agent
All-PIU, one patch author receives the ordered union of all PIU evidence. In
Multi-Agent PIU-Only, each participant receives only its assigned functional
PIU. In Multi-Agent Full-Context, each participant receives its assigned PIU
together with non-PIU base-repository code and additional runtime context.

Non-PIU code fragments in the Full-Context packets are distributed exclusively
across participants. Additional runtime context uses only previously collected
observations; selected non-owned observations may be duplicated as distractors
across participant packets. No participant receives the entire repository.
The collaborative transcript is sequential and coordinator-selected, and the
final patch author acts on the public transcript using the same five-turn
repair interface and public-check catalog. No condition exposes PIU
identifiers, the reference patch, hidden-test information, hidden expected
values, or constructor rationales.

\paragraph{Repeated-run aggregation and failure handling.}
To account for stochastic repair trajectories, we schedule three independent
runs for each task--model--condition combination. Within each task and
condition, we first average the available verifier outcomes across repetitions
and then average across the 12 tasks, giving each task equal weight.

A completed verifier failure, an ordinary unsuccessful repair, or a confirmed
context-window failure receives a score of zero and remains in the denominator.
A run is marked unavailable and excluded only when no valid task outcome can be
obtained because of provider-capacity, API or serving, container or Docker, or
missing-verifier-result failures. Availability exclusions are recorded
separately from completed task failures.

\subsection{Independent Validation of PIU Construction}
\label{app:piu-construction-validation}

All problem instances undergo checks during PIU construction, and the
participant-level source-material check is performed after source-material
distribution. Centralized solver success additionally establishes operational
answerability, but does not by itself validate leave-one-out pivotality or the
participant-level source-material check. We therefore conducted an additional
post-construction audit of 20 instances from the fixed 30-instance MAUD
evaluation set and 20 instances from the fixed 30-instance FinLongDocQA
evaluation set. We used GPT-5.4, which was not used for instance construction,
as the primary evaluator. The sample was frozen before evaluation. The audit
covered every selected instance under every PIU leave-one-out removal and each
participant's initial local observation together with the initial common
observation. Solver outputs and answer accuracy were not used as evidence in
these judgments.

For the leave-one-out pivotality check, a removal passes only when the linked
answer component becomes unsupported or genuinely unresolved. The
participant-level source-material check passes at the instance level only when
no participant's initial local observation together with the initial common
observation grounds the complete reference decision.

\begin{table*}[t]
\centering
\small
\setlength{\tabcolsep}{6pt}
\renewcommand{\arraystretch}{1.08}
\caption{\textbf{Independent validation of two representative PIU-construction
regimes.} Values are passing judgments or instances over the corresponding
validation denominator, with percentages and 95\% intervals in brackets. Instance-
level intervals are Wilson intervals; leave-one-out intervals use an
instance-cluster bootstrap with 10,000 resamples.}
\label{tab:piu-construction-validation}
\begin{tabular}{@{}lcc@{}}
\toprule
Source corpus &
\makecell{Leave-one-out\\pivotality} &
\makecell{Participant-level\\source-material check} \\
\midrule
MAUD &
64/67 (95.5\%) [89.5, 100.0] &
18/20 (90.0\%) [69.9, 97.2] \\
FinLongDocQA &
65/67 (97.0\%) [92.3, 100.0] &
20/20 (100.0\%) [83.9, 100.0] \\
\bottomrule
\end{tabular}

\vspace{2pt}
\parbox{0.96\textwidth}{\footnotesize
This is targeted independent validation of two representative construction regimes and is
not an independent revalidation of every retained instance or source corpus.
An unclear, invalid, failed, or missing judgment is counted as non-passing.}
\end{table*}

Across the two validated construction regimes, the leave-one-out results provide
the strongest evidence that retained PIUs are individually pivotal. The
participant-level source-material check finds that no participant's initial
observation is sufficient for a complete grounded decision in any selected
FinLongDocQA instance and identifies two MAUD instances in which one
participant's initial observation together with the initial common observation
is sufficient.

\revised{%
\FloatBarrier
\subsection{\revised{Human Validation of PIU Construction}}
\label{app:piu-human-validation}
\begingroup

Across the four domains, human raters agreed on 90.0\% of complete-set
sufficiency judgments and 93.8\% of unit-level necessity judgments. The evaluation
covered 40 problem instances containing 209 information units, with two raters
per domain. These overall rates pool matched judgments across domains. Agreement on whether a set
was sufficient with every unit necessary was 80.0\%. Table~\ref{tab:human-piu-agreement}
reports the domain-level results before adjudication.

\begin{table}[H]
\centering
\small
\setlength{\tabcolsep}{5pt}
\renewcommand{\arraystretch}{1.08}
\caption{\revised{\textbf{Inter-rater agreement before adjudication.} Sufficiency is evaluated per instance and necessity per unit. Percentages are raw agreement; the overall row pools matched judgments. Cohen's $\kappa$ is reported within each domain.}}
\label{tab:human-piu-agreement}
\revised{%
\begin{tabular}{@{}lccccc@{}}
\toprule
Domain & \makecell{Instances /\\units} & \makecell{Sufficiency\\agreement} & $\kappa$ & \makecell{Necessity\\agreement} & $\kappa$ \\
\midrule
Legal & 10 / 61 & 80.0\% & 0.412 & 83.6\% & 0.000 \\
Medical & 10 / 33 & 80.0\% & 0.412 & 100.0\% & 1.000 \\
\makecell[l]{Software\\engineering} & 10 / 67 & 100.0\% & 1.000 & 95.5\% & 0.816 \\
Finance & 10 / 48 & 100.0\% & 1.000 & 100.0\% & 1.000 \\
\midrule
Overall (pooled) & 40 / 209 & 90.0\% & --- & 93.8\% & --- \\
\bottomrule
\end{tabular}}
\end{table}

The legal necessity $\kappa$ is zero because one rater labeled all 61 units
necessary, while the other labeled 51 necessary and 10 not needed; observed and
expected agreement therefore coincide despite 83.6\% raw agreement.

\paragraph{Protocol.} Eight annotators with academic backgrounds relevant to their
assigned domains participated, with two annotators per domain. They were
offered compensation for their time. No demographic or sensitive personal
information was requested, and responses were indexed by rater codes.
In each domain, raters evaluated five constructed PIU sets and five other
candidate-unit sets judged by the model to be insufficient or to contain
unnecessary units. Group membership and model-based judgments were not shown
to raters, who were instructed to judge without sharing responses.

Sufficiency concerns whether the complete set supports the reference decision
$d^{\mathrm{ref}}$. Necessity concerns whether removing one unit while keeping
the others fixed leaves a required answer component or reasoning step
unsupported or unresolved. For constructed PIUs, this tests
$\mathcal{I}^{\min}_q\setminus\{i_j\}$ for each unit $i_j$. Raters also assessed
the unique contribution of individual units when the complete set was
insufficient. The sufficiency interface offered \emph{Sufficient},
\emph{Insufficient}, and \emph{Cannot judge}; the necessity interface offered
\emph{Necessary}, \emph{Not needed}, and \emph{Cannot judge}. The sufficiency
analysis labels the third category ambiguous. Ambiguous and cannot-judge
responses were retained as distinct categories in the corresponding
agreement calculations. Both requirements were confirmed only when the set was
judged sufficient and every unit necessary.

\paragraph{Confirmation of constructed PIUs.} Both raters judged all 20 constructed
PIU sets sufficient. In 18 sets, both also confirmed the necessity of every
unit. The remaining two were software-engineering instances in which one rater
considered one unit unnecessary while still affirming complete-set sufficiency.
Thus, the sampled constructed PIUs received unanimous confirmation of
sufficiency, with localized differences over unit-level necessity.

\paragraph{Judgments on other candidate-unit sets.} In four legal and two medical
instances, one rater accepted the set as sufficient with every unit necessary,
whereas the other identified a sufficiency or necessity failure. For all other
candidate-unit sets, both raters identified a failure. In software engineering
and finance, both raters identified a failure in every candidate-unit set.

These judgments separated missing information from unnecessary information.
In medical, software engineering, and finance, both raters judged the two sets
containing unnecessary information sufficient while marking the identified
units unnecessary. Both software-engineering raters and both finance raters
also judged all three sets with missing information insufficient. Medical
raters differed on two such instances. In one, the excerpts covered four days
of nanoparticle exposure but omitted support for the reference answer's
six-hours-per-day schedule. Another supplied cohort sizes and chromium
measurements for hard chromeplating but did not establish the claimed exposure
contrast with bright chromeplating. One rater accepted these sets, whereas the
other and the model judged them insufficient.

\paragraph{Interpretation.} These results support the sufficiency of the
sampled constructed PIUs. Disagreements on the other candidate-unit sets
suggest that the model-based checks applied a more conservative evidence
requirement in some cases: a set could be acceptable to a human rater while
failing the model's sufficiency or necessity check. Figure~\ref{fig:human-annotation-interface} shows the annotation interface.

\begin{figure}[tbp]
\centering
\includegraphics[width=0.82\textwidth]{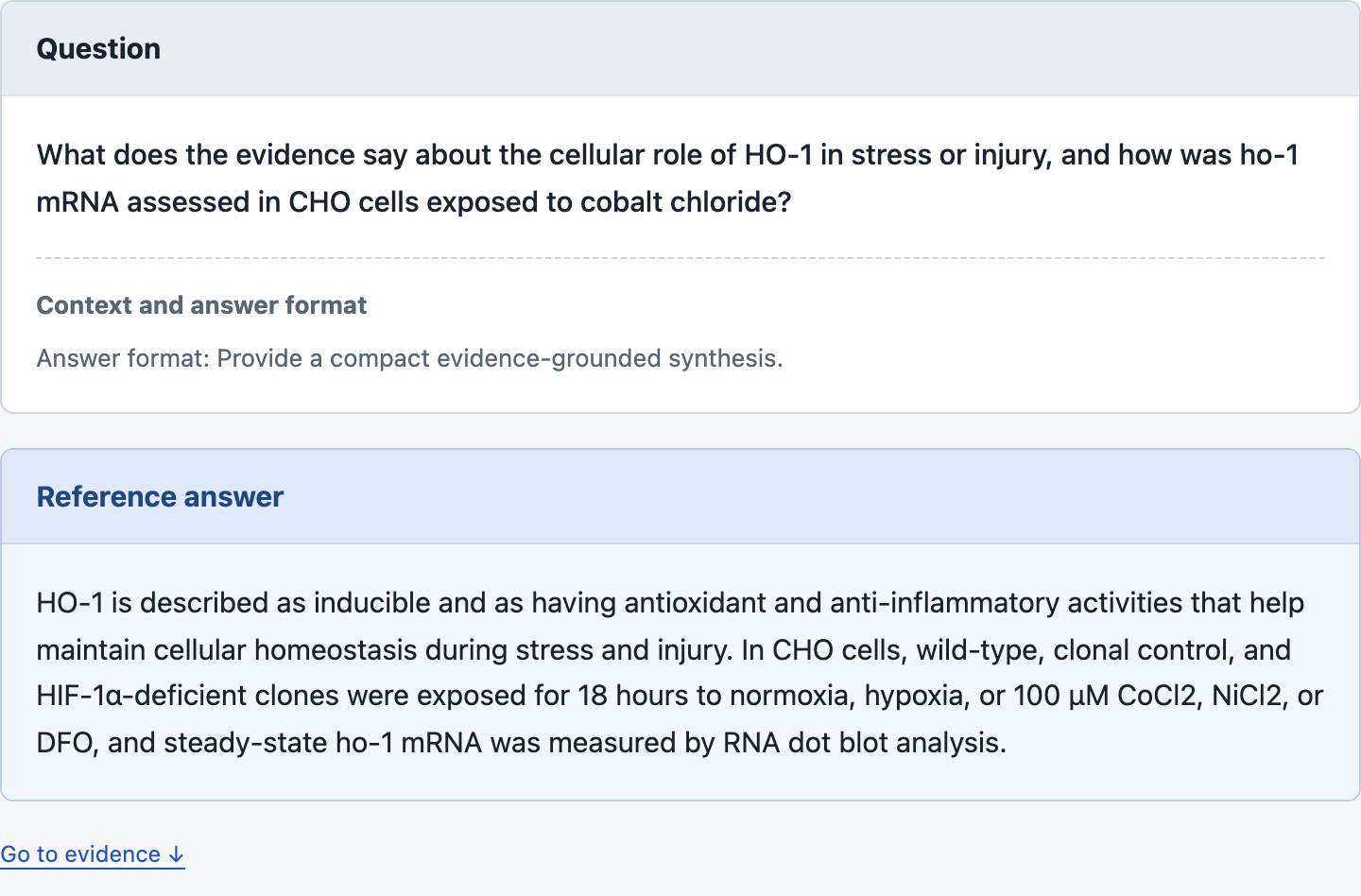}\\[5pt]
\includegraphics[width=0.82\textwidth]{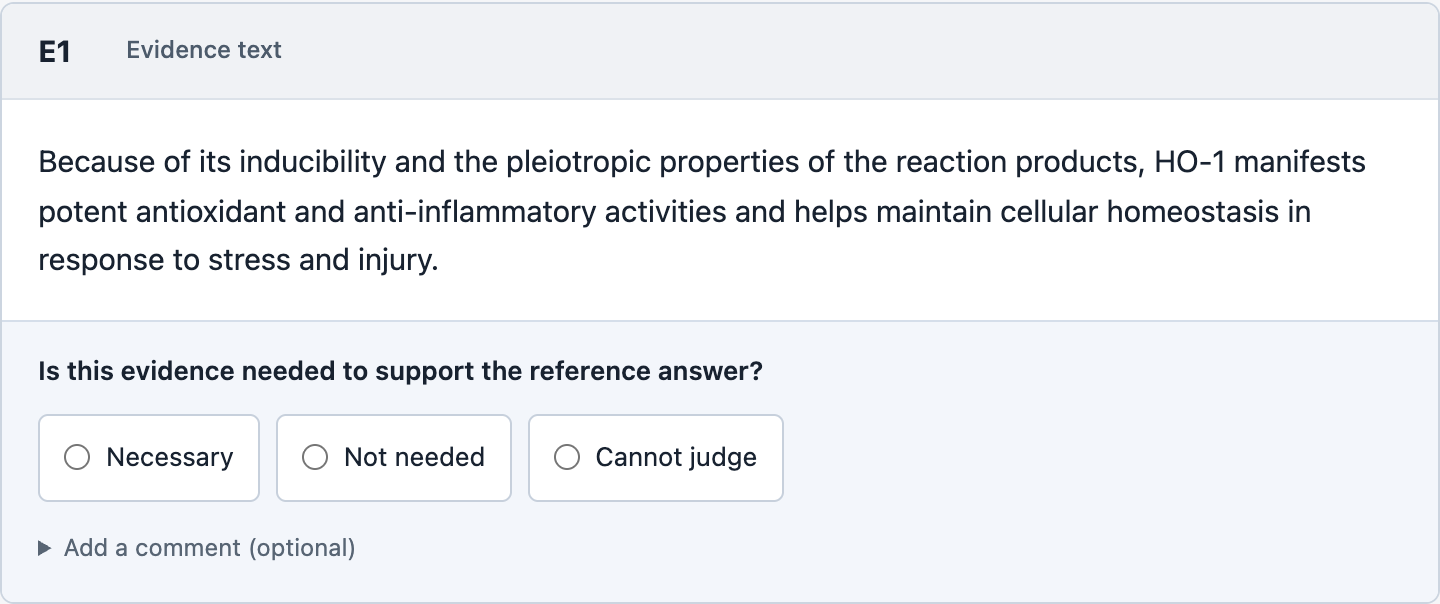}\\[5pt]
\includegraphics[width=0.82\textwidth]{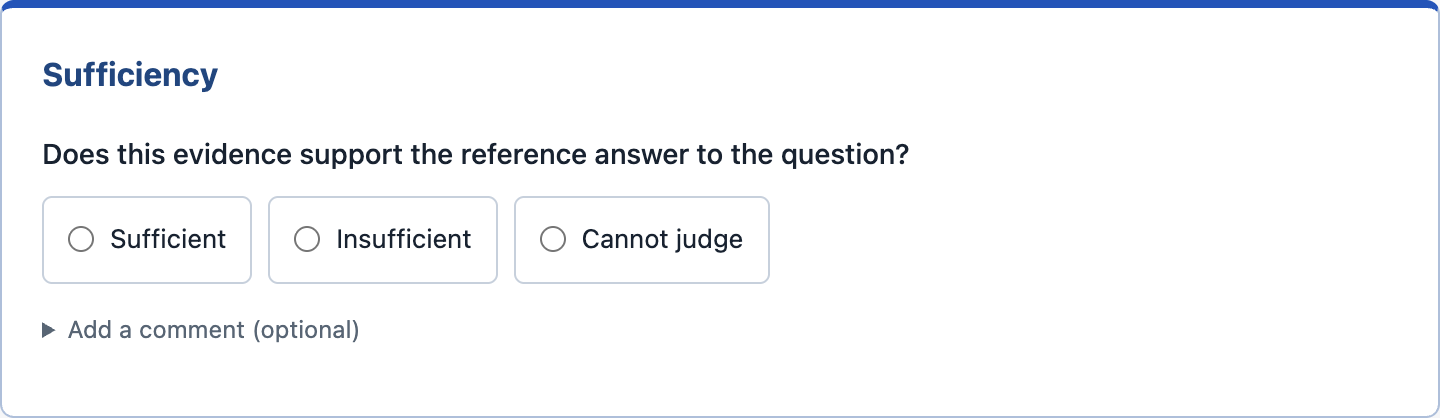}
\caption{\revised{\textbf{Human-annotation interface.} Selected regions from an unanswered medical instance (MED-10): the question and reference answer (top), necessity judgment for one evidence unit (middle), and sufficiency judgment for the complete evidence set (bottom).}}
\label{fig:human-annotation-interface}
\end{figure}

\FloatBarrier
\endgroup
}

\section{Evaluation Results Across Domains and Source Corpora}
\label{app:cross-domain-three-evaluation-conditions-results}

The first three subsections disaggregate the model--domain results in
Figure~\ref{fig:cross-domain-three-evaluation-conditions-heatmap}. Every reported
comparison uses the same problem instances, their pivotal information, and the
problem-specific evaluator across the three evaluation conditions within a source corpus. The
legal, finance, medical, and software-engineering evaluations use
\OneRIUPerParticipant{}. All
team episodes use \CoordinatorSelectedSpeaker{} with 15 participant responses
before the final decision. Within each model--source-corpus
comparison, the two multi-agent conditions hold this PIU distribution,
communication structure, and the source-corpus-specific resource budgets
fixed. The results do not pool different communication structures or
PIU-distribution conditions within an episode.
For the three evaluation conditions in
Figure~\ref{fig:cross-domain-three-evaluation-conditions-heatmap}, confirmed
context-window failures count as unsuccessful episodes, whereas episodes
without a valid outcome because of provider-capacity or other infrastructure
failures are unavailable.

\subsection{Source-Corpus Results}

Table~\ref{tab:cross-domain-source-corpus-summary} reports episode success rate
for each source corpus and gives equal weight to the models. The model--domain
results in Figure~\ref{fig:cross-domain-three-evaluation-conditions-heatmap} pool problem
instances across the source corpora within each model.

\begin{table*}[!h]
\centering
\scriptsize
\setlength{\tabcolsep}{2.4pt}
\renewcommand{\arraystretch}{1.05}
\caption{\textbf{Evaluation results by source corpus.} Values are episode success rates in percent. Models receive equal weight within each source corpus. The Models column gives scheduled coverage; unavailable estimates are omitted condition by condition. Smaller $\pm$ terms are binomial standard errors propagated through each fixed-model average.}
\label{tab:cross-domain-source-corpus-summary}
\begin{tabular*}{\textwidth}{@{\extracolsep{\fill}}llrrrrr@{}}
\toprule
Domain & Source corpus & \makecell{Problem\\instances} & Models & \makecell{Single-Agent\\All-PIU} & \makecell{Multi-Agent\\PIU-Only} & \makecell{Multi-Agent\\Full-Context} \\
\midrule
Legal & CUAD & 30 & 8 & 90.4\uncert{1.7} & 72.5\uncert{2.7} & 62.9\uncert{2.8} \\
 & MAUD & 30 & 8 & 88.3\uncert{1.8} & 67.9\uncert{2.2} & 60.0\uncert{2.7} \\
 & eCFR Casebook & 18 & 8 & 88.9\uncert{2.2} & 47.9\uncert{3.5} & 18.1\uncert{2.5} \\
 & MSLR & 11 & 8 & 57.9\uncert{4.0} & 45.5\uncert{4.6} & 8.0\uncert{2.6} \\
 & Multi-LexSum & 29 & 8 & 62.5\uncert{2.3} & 30.2\uncert{2.3} & 20.2\uncert{2.1} \\
 & SARA & 45 & 8 & 63.6\uncert{1.9} & 37.2\uncert{2.0} & 8.1\uncert{1.4} \\
\addlinespace[2pt]
Finance & FinDVer & 30 & 8 & 59.2\uncert{2.6} & 25.4\uncert{1.9} & 17.1\uncert{1.8} \\
 & FinLongDocQA & 30 & 8 & 62.5\uncert{2.4} & 46.6\uncert{2.3} & 22.9\uncert{1.6} \\
\addlinespace[2pt]
Medical & MLCR & 29 & 8 & 79.3\uncert{2.5} & 44.8\uncert{3.0} & 12.5\uncert{2.0} \\
 & EvidenceBench & 45 & 8 & 49.4\uncert{2.4} & 30.8\uncert{2.0} & 6.9\uncert{1.3} \\
\addlinespace[2pt]
Software Engineering & SWE Atlas Codebase QnA & 32 & 8 & 49.2\uncert{2.6} & 34.8\uncert{2.8} & 8.2\uncert{1.6} \\
 & DeepSWE & 12 & 8 & 9.0\uncert{1.6} & 2.4\uncert{1.4} & 2.1\uncert{1.3} \\
\bottomrule
\end{tabular*}
\end{table*}

\subsection{Model and Source-Corpus Results}
\label{app:cross-domain-model-source-corpus-results}

Table~\ref{tab:cross-domain-model-source-corpus-results} reports the complete
three-condition model and source-corpus results underlying the domain and
source-corpus summaries, together with available Single-Agent Full-Context
estimates. In that condition, one model
receives the ordered union of the source material assigned under the same fixed
PIU distribution. Each row averages episode success over the fixed problem
instances for one model and source corpus. Within a row, the two multi-agent
conditions use the same fixed PIU distributions, communication structure, and
resource budgets. The table does not pool the additional PIU-distribution
conditions or communication structures.
For legal Single-Agent Full-Context, the episode-success denominator includes
evaluated inputs and structured-interface failures at the final-decision stage,
with the latter counted as unsuccessful episodes. Serving-context rejections
and other evaluation errors are excluded. When fewer than the fixed inputs
contribute to an estimate, the table shows its denominator as \(n/N\). An em
dash in the Single-Agent Full-Context column means that the current archive
does not provide a comparable success estimate; it does not convert an
unavailable result into a failure.

\begingroup
\raggedbottom
\scriptsize
\setlength{\tabcolsep}{1.4pt}
\renewcommand{\arraystretch}{1.04}
\begin{longtable}{@{}lllr@{\hspace{4pt}}rrrr@{}}
\caption{\textbf{Evaluation results by model and source corpus.} Values are episode success rates in percent. Smaller $\pm$ terms are binomial standard errors over each condition's episode-success denominator. For the three primary conditions, confirmed context-window failures are counted as unsuccessful episodes, while provider-capacity and other infrastructure failures remain unavailable. Single-Agent Full-Context excludes context-window, infrastructure, and judge failures, while output failures are counted as unsuccessful episodes. For legal Single-Agent Full-Context, serving-context rejections and other evaluation errors are excluded from this denominator, while structured-interface failures at the final-decision stage are included as unsuccessful episodes. When fewer than the fixed inputs contribute to an estimate, its denominator is shown as $n/N$. An em dash denotes that no comparable estimate is available. Appendix~\ref{app:cross-domain-single-full-context} reports the corresponding coverage and failure counts.}\label{tab:cross-domain-model-source-corpus-results}\\
\toprule
Domain & Source corpus & Model & \makecell{Problem\\instances} & \makecell{Single-Agent\\All-PIU} & \makecell{Single-Agent\\Full-Context} & \makecell{Multi-Agent\\PIU-Only} & \makecell{Multi-Agent\\Full-Context} \\
\midrule
\endfirsthead
\multicolumn{8}{c}{\tablename~\thetable{} continued} \\
\toprule
Domain & Source corpus & Model & \makecell{Problem\\instances} & \makecell{Single-Agent\\All-PIU} & \makecell{Single-Agent\\Full-Context} & \makecell{Multi-Agent\\PIU-Only} & \makecell{Multi-Agent\\Full-Context} \\
\midrule
\endhead
\midrule
\multicolumn{8}{r}{Continued on next page} \\
\endfoot
\bottomrule
\endlastfoot
Legal & CUAD & GPT-5.5 & 30 & 96.7\uncert{3.3} & 86.7\uncert{6.2} & 93.3\uncert{4.6} & 90.0\uncert{5.5} \\
 &  & Claude Sonnet 4.5 & 30 & 100.0\uncert{0.0} & 90.0\uncert{5.5} & 53.3\uncert{9.1} & 43.3\uncert{9.0} \\
 &  & Claude Haiku 4.5 & 30 & 90.0\uncert{5.5} & 93.3\uncert{4.6} & 70.0\uncert{8.4} & 60.0\uncert{8.9} \\
 &  & Gemma 3 4B IT & 30 & 60.0\uncert{8.9} & 30.0\uncert{8.4} & 50.0\uncert{9.1} & 40.0\uncert{8.9} \\
 &  & Gemma 3 12B IT & 30 & 90.0\uncert{5.5} & 53.3\uncert{9.1} & 70.0\uncert{8.4} & 36.7\uncert{8.8} \\
 &  & Gemma 3 27B IT & 30 & 93.3\uncert{4.6} & 50.0\uncert{9.1} & 80.0\uncert{7.3} & 70.0\uncert{8.4} \\
 &  & Qwen3-32B & 30 & 93.3\uncert{4.6} & \makecell{83.3\uncert{7.6}\\{\tiny $n=24/30$}} & 73.3\uncert{8.1} & 76.7\uncert{7.7} \\
 &  & Qwen3-235B-A22B & 30 & 100.0\uncert{0.0} & 93.3\uncert{4.6} & 90.0\uncert{5.5} & 86.7\uncert{6.2} \\
\midrule
Legal & MAUD & GPT-5.5 & 30 & 100.0\uncert{0.0} & 96.7\uncert{3.3} & 100.0\uncert{0.0} & 100.0\uncert{0.0} \\
 &  & Claude Sonnet 4.5 & 30 & 96.7\uncert{3.3} & 80.0\uncert{7.3} & 3.3\uncert{3.3} & 30.0\uncert{8.4} \\
 &  & Claude Haiku 4.5 & 30 & 90.0\uncert{5.5} & 93.3\uncert{4.6} & 73.3\uncert{8.1} & 56.7\uncert{9.0} \\
 &  & Gemma 3 4B IT & 30 & 46.7\uncert{9.1} & \makecell{17.2\uncert{7.0}\\{\tiny $n=29/30$}} & 33.3\uncert{8.6} & 40.0\uncert{8.9} \\
 &  & Gemma 3 12B IT & 30 & 93.3\uncert{4.6} & \makecell{72.4\uncert{8.3}\\{\tiny $n=29/30$}} & 73.3\uncert{8.1} & 63.3\uncert{8.8} \\
 &  & Gemma 3 27B IT & 30 & 86.7\uncert{6.2} & \makecell{58.6\uncert{9.1}\\{\tiny $n=29/30$}} & 76.7\uncert{7.7} & 63.3\uncert{8.8} \\
 &  & Qwen3-32B & 30 & 96.7\uncert{3.3} & --- & 86.7\uncert{6.2} & 33.3\uncert{8.6} \\
 &  & Qwen3-235B-A22B & 30 & 96.7\uncert{3.3} & 96.7\uncert{3.3} & 96.7\uncert{3.3} & 93.3\uncert{4.6} \\
\midrule
Legal & eCFR Casebook & GPT-5.5 & 18 & 100.0\uncert{0.0} & 88.9\uncert{7.4} & 83.3\uncert{8.8} & 55.6\uncert{11.7} \\
 &  & Claude Sonnet 4.5 & 18 & 100.0\uncert{0.0} & 100.0\uncert{0.0} & 55.6\uncert{11.7} & 55.6\uncert{11.7} \\
 &  & Claude Haiku 4.5 & 18 & 100.0\uncert{0.0} & 88.9\uncert{7.4} & 72.2\uncert{10.6} & 27.8\uncert{10.6} \\
 &  & Gemma 3 4B IT & 18 & 50.0\uncert{11.8} & 0.0\uncert{0.0} & 0.0\uncert{0.0} & 0.0\uncert{0.0} \\
 &  & Gemma 3 12B IT & 18 & 83.3\uncert{8.8} & 16.7\uncert{8.8} & 27.8\uncert{10.6} & 0.0\uncert{0.0} \\
 &  & Gemma 3 27B IT & 18 & 83.3\uncert{8.8} & 11.1\uncert{7.4} & 22.2\uncert{9.8} & 0.0\uncert{0.0} \\
 &  & Qwen3-32B & 18 & 94.4\uncert{5.4} & --- & 66.7\uncert{11.1} & 0.0\uncert{0.0} \\
 &  & Qwen3-235B-A22B & 18 & 100.0\uncert{0.0} & 94.4\uncert{5.4} & 55.6\uncert{11.7} & 5.6\uncert{5.4} \\
\midrule
Legal & MSLR & GPT-5.5 & 11 & 100.0\uncert{0.0} & 27.3\uncert{13.4} & 81.8\uncert{11.6} & 36.4\uncert{14.5} \\
 &  & Claude Sonnet 4.5 & 11 & 81.8\uncert{11.6} & 9.1\uncert{8.7} & 63.6\uncert{14.5} & 9.1\uncert{8.7} \\
 &  & Claude Haiku 4.5 & 11 & 81.8\uncert{11.6} & 0.0\uncert{0.0} & 27.3\uncert{13.4} & 0.0\uncert{0.0} \\
 &  & Gemma 3 4B IT & 11 & 9.1\uncert{8.7} & 0.0\uncert{0.0} & 0.0\uncert{0.0} & 9.1\uncert{8.7} \\
 &  & Gemma 3 12B IT & 11 & 27.3\uncert{13.4} & 0.0\uncert{0.0} & 63.6\uncert{14.5} & 0.0\uncert{0.0} \\
 &  & Gemma 3 27B IT & 11 & 18.2\uncert{11.6} & 9.1\uncert{8.7} & 36.4\uncert{14.5} & 0.0\uncert{0.0} \\
 &  & Qwen3-32B & 11 & 63.6\uncert{14.5} & \makecell{28.6\uncert{17.1}\\{\tiny $n=7/11$}} & 36.4\uncert{14.5} & 9.1\uncert{8.7} \\
 &  & Qwen3-235B-A22B & 11 & 81.8\uncert{11.6} & 45.5\uncert{15.0} & 54.5\uncert{15.0} & 0.0\uncert{0.0} \\
\midrule
Legal & Multi-LexSum & GPT-5.5 & 29 & 100.0\uncert{0.0} & 72.4\uncert{8.3} & 93.1\uncert{4.7} & 72.4\uncert{8.3} \\
 &  & Claude Sonnet 4.5 & 29 & 93.1\uncert{4.7} & 41.4\uncert{9.1} & 24.1\uncert{7.9} & 0.0\uncert{0.0} \\
 &  & Claude Haiku 4.5 & 29 & 82.8\uncert{7.0} & 31.0\uncert{8.6} & 48.3\uncert{9.3} & 31.0\uncert{8.6} \\
 &  & Gemma 3 4B IT & 29 & 6.9\uncert{4.7} & 0.0\uncert{0.0} & 0.0\uncert{0.0} & 0.0\uncert{0.0} \\
 &  & Gemma 3 12B IT & 29 & 31.0\uncert{8.6} & 3.4\uncert{3.4} & 3.4\uncert{3.4} & 0.0\uncert{0.0} \\
 &  & Gemma 3 27B IT & 29 & 37.9\uncert{9.0} & 6.9\uncert{4.7} & 6.9\uncert{4.7} & 3.4\uncert{3.4} \\
 &  & Qwen3-32B & 29 & 48.3\uncert{9.3} & 24.1\uncert{7.9} & 24.1\uncert{7.9} & 31.0\uncert{8.6} \\
 &  & Qwen3-235B-A22B & 29 & 100.0\uncert{0.0} & 20.7\uncert{7.5} & 41.4\uncert{9.1} & 24.1\uncert{7.9} \\
\midrule
Legal & SARA & GPT-5.5 & 45 & 93.3\uncert{3.7} & 40.0\uncert{7.3} & 91.1\uncert{4.2} & 22.2\uncert{6.2} \\
 &  & Claude Sonnet 4.5 & 45 & 93.3\uncert{3.7} & 13.3\uncert{5.1} & 42.2\uncert{7.4} & 4.4\uncert{3.1} \\
 &  & Claude Haiku 4.5 & 45 & 86.7\uncert{5.1} & 0.0\uncert{0.0} & 75.6\uncert{6.4} & 8.9\uncert{4.2} \\
 &  & Gemma 3 4B IT & 45 & 2.2\uncert{2.2} & \makecell{0.0\uncert{0.0}\\{\tiny $n=30/45$}} & 0.0\uncert{0.0} & 0.0\uncert{0.0} \\
 &  & Gemma 3 12B IT & 45 & 46.7\uncert{7.4} & \makecell{3.3\uncert{3.3}\\{\tiny $n=30/45$}} & 6.7\uncert{3.7} & 6.7\uncert{3.7} \\
 &  & Gemma 3 27B IT & 45 & 48.9\uncert{7.5} & \makecell{0.0\uncert{0.0}\\{\tiny $n=30/45$}} & 15.6\uncert{5.4} & 11.1\uncert{4.7} \\
 &  & Qwen3-32B & 45 & 46.7\uncert{7.4} & --- & 24.4\uncert{6.4} & 4.4\uncert{3.1} \\
 &  & Qwen3-235B-A22B & 45 & 91.1\uncert{4.2} & 13.3\uncert{5.1} & 42.2\uncert{7.4} & 6.7\uncert{3.7} \\
\midrule
Finance & FinDVer & GPT-5.5 & 30 & 86.7\uncert{6.2} & 36.7\uncert{8.8} & 86.7\uncert{6.2} & 63.3\uncert{8.8} \\
 &  & Claude Sonnet 4.5 & 30 & 86.7\uncert{6.2} & 53.3\uncert{9.1} & 70.0\uncert{8.4} & 56.7\uncert{9.0} \\
 &  & Claude Haiku 4.5 & 30 & 73.3\uncert{8.1} & 36.7\uncert{8.8} & 0.0\uncert{0.0} & 0.0\uncert{0.0} \\
 &  & Gemma 3 4B IT & 30 & 3.3\uncert{3.3} & 0.0\uncert{0.0} & 0.0\uncert{0.0} & 0.0\uncert{0.0} \\
 &  & Gemma 3 12B IT & 30 & 33.3\uncert{8.6} & 0.0\uncert{0.0} & 3.3\uncert{3.3} & 3.3\uncert{3.3} \\
 &  & Gemma 3 27B IT & 30 & 43.3\uncert{9.0} & 6.7\uncert{4.6} & 23.3\uncert{7.7} & 3.3\uncert{3.3} \\
 &  & Qwen3-32B & 30 & 70.0\uncert{8.4} & \makecell{40.0\uncert{21.9}\\{\tiny $n=5/30$}} & 0.0\uncert{0.0} & 0.0\uncert{0.0} \\
 &  & Qwen3-235B-A22B & 30 & 76.7\uncert{7.7} & 16.7\uncert{6.8} & 20.0\uncert{7.3} & 10.0\uncert{5.5} \\
\midrule
Finance & FinLongDocQA & GPT-5.5 & 30 & 100.0\uncert{0.0} & 73.3\uncert{8.1} & 93.3\uncert{4.6} & 90.0\uncert{5.5} \\
 &  & Claude Sonnet 4.5 & 30 & 96.7\uncert{3.3} & \makecell{61.5\uncert{9.5}\\{\tiny $n=26/30$}} & 83.3\uncert{6.8} & 70.0\uncert{8.4} \\
 &  & Claude Haiku 4.5 & 30 & 56.7\uncert{9.0} & \makecell{53.8\uncert{9.8}\\{\tiny $n=26/30$}} & 3.3\uncert{3.3} & 0.0\uncert{0.0} \\
 &  & Gemma 3 4B IT & 30 & 0.0\uncert{0.0} & \makecell{4.8\uncert{4.6}\\{\tiny $n=21/30$}} & 0.0\uncert{0.0} & 0.0\uncert{0.0} \\
 &  & Gemma 3 12B IT & 30 & 33.3\uncert{8.6} & \makecell{9.5\uncert{6.4}\\{\tiny $n=21/30$}} & 10.0\uncert{5.5} & 0.0\uncert{0.0} \\
 &  & Gemma 3 27B IT & 30 & 53.3\uncert{9.1} & \makecell{23.8\uncert{9.3}\\{\tiny $n=21/30$}} & 53.3\uncert{9.1} & 6.7\uncert{4.6} \\
 &  & Qwen3-32B & 30 & 80.0\uncert{7.3} & \makecell{50.0\uncert{35.4}\\{\tiny $n=2/30$}} & 63.3\uncert{8.8} & 0.0\uncert{0.0} \\
 &  & Qwen3-235B-A22B & 30 & 80.0\uncert{7.3} & \makecell{21.4\uncert{7.8}\\{\tiny $n=28/30$}} & 66.7\uncert{8.6} & 16.7\uncert{6.8} \\
\midrule
Medical & MLCR & GPT-5.5 & 29 & 96.6\uncert{3.4} & 31.0\uncert{8.6} & 65.5\uncert{8.8} & 37.9\uncert{9.0} \\
 &  & Claude Sonnet 4.5 & 29 & 86.2\uncert{6.4} & 37.9\uncert{9.0} & 72.4\uncert{8.3} & 17.2\uncert{7.0} \\
 &  & Claude Haiku 4.5 & 29 & 82.8\uncert{7.0} & 31.0\uncert{8.6} & 58.6\uncert{9.1} & 17.2\uncert{7.0} \\
 &  & Gemma 3 4B IT & 29 & 51.7\uncert{9.3} & 0.0\uncert{0.0} & 6.9\uncert{4.7} & 0.0\uncert{0.0} \\
 &  & Gemma 3 12B IT & 29 & 79.3\uncert{7.5} & 3.4\uncert{3.4} & 24.1\uncert{7.9} & 0.0\uncert{0.0} \\
 &  & Gemma 3 27B IT & 29 & 79.3\uncert{7.5} & 6.9\uncert{4.7} & 34.5\uncert{8.8} & 3.4\uncert{3.4} \\
 &  & Qwen3-32B & 29 & 75.9\uncert{7.9} & \makecell{0.0\uncert{0.0}\\{\tiny $n=12/29$}} & 41.4\uncert{9.1} & 6.9\uncert{4.7} \\
 &  & Qwen3-235B-A22B & 29 & 82.8\uncert{7.0} & 31.0\uncert{8.6} & 55.2\uncert{9.2} & 17.2\uncert{7.0} \\
\midrule
Medical & EvidenceBench & GPT-5.5 & 45 & 93.3\uncert{3.7} & 31.1\uncert{6.9} & 88.9\uncert{4.7} & 20.0\uncert{6.0} \\
 &  & Claude Sonnet 4.5 & 45 & 57.8\uncert{7.4} & 24.4\uncert{6.4} & 48.9\uncert{7.5} & 13.3\uncert{5.1} \\
 &  & Claude Haiku 4.5 & 45 & 53.3\uncert{7.4} & 20.0\uncert{6.0} & 44.4\uncert{7.4} & 4.4\uncert{3.1} \\
 &  & Gemma 3 4B IT & 45 & 13.3\uncert{5.1} & 0.0\uncert{0.0} & 0.0\uncert{0.0} & 0.0\uncert{0.0} \\
 &  & Gemma 3 12B IT & 45 & 33.3\uncert{7.0} & 0.0\uncert{0.0} & 13.3\uncert{5.1} & 2.2\uncert{2.2} \\
 &  & Gemma 3 27B IT & 45 & 44.4\uncert{7.4} & 8.9\uncert{4.2} & 6.7\uncert{3.7} & 2.2\uncert{2.2} \\
 &  & Qwen3-32B & 45 & 37.8\uncert{7.2} & 8.9\uncert{4.2} & 20.0\uncert{6.0} & 6.7\uncert{3.7} \\
 &  & Qwen3-235B-A22B & 45 & 62.2\uncert{7.2} & 8.9\uncert{4.2} & 24.4\uncert{6.4} & 6.7\uncert{3.7} \\
\midrule
Software Engineering & SWE Atlas Codebase QnA & GPT-5.5 & 32 & 84.4\uncert{6.4} & \makecell{61.9\uncert{10.6}\\{\tiny $n=21/32$}} & 43.8\uncert{8.8} & 25.0\uncert{7.7} \\
 &  & Claude Sonnet 4.5 & 32 & 81.2\uncert{6.9} & \makecell{0.0\uncert{0.0}\\{\tiny $n=1/32$}} & 50.0\uncert{8.8} & 25.0\uncert{7.7} \\
 &  & Claude Haiku 4.5 & 32 & 75.0\uncert{7.7} & \makecell{100.0\uncert{0.0}\\{\tiny $n=1/32$}} & 50.0\uncert{8.8} & 9.4\uncert{5.2} \\
 &  & Gemma 3 4B IT & 32 & 3.1\uncert{3.1} & \makecell{0.0\uncert{0.0}\\{\tiny $n=2/32$}} & 3.1\uncert{3.1} & 0.0\uncert{0.0} \\
 &  & Gemma 3 12B IT & 32 & 31.2\uncert{8.2} & \makecell{0.0\uncert{0.0}\\{\tiny $n=2/32$}} & 28.1\uncert{7.9} & 0.0\uncert{0.0} \\
 &  & Gemma 3 27B IT & 32 & 43.8\uncert{8.8} & \makecell{0.0\uncert{0.0}\\{\tiny $n=2/32$}} & 40.6\uncert{8.7} & 0.0\uncert{0.0} \\
 &  & Qwen3-32B & 32 & 31.2\uncert{8.2} & --- & 25.0\uncert{7.7} & 3.1\uncert{3.1} \\
 &  & Qwen3-235B-A22B & 32 & 43.8\uncert{8.8} & \makecell{0.0\uncert{0.0}\\{\tiny $n=2/32$}} & 37.5\uncert{8.6} & 3.1\uncert{3.1} \\
\midrule
Software Engineering & DeepSWE & GPT-5.5 & 12 & 72.2\uncert{12.9} & --- & 19.4\uncert{11.4} & 16.7\uncert{10.8} \\
 &  & Claude Sonnet 4.5 & 12 & 0.0\uncert{0.0} & --- & 0.0\uncert{0.0} & 0.0\uncert{0.0} \\
 &  & Claude Haiku 4.5 & 12 & 0.0\uncert{0.0} & --- & 0.0\uncert{0.0} & 0.0\uncert{0.0} \\
 &  & Gemma 3 4B IT & 12 & 0.0\uncert{0.0} & --- & 0.0\uncert{0.0} & 0.0\uncert{0.0} \\
 &  & Gemma 3 12B IT & 12 & 0.0\uncert{0.0} & --- & 0.0\uncert{0.0} & 0.0\uncert{0.0} \\
 &  & Gemma 3 27B IT & 12 & 0.0\uncert{0.0} & --- & 0.0\uncert{0.0} & 0.0\uncert{0.0} \\
 &  & Qwen3-32B & 12 & 0.0\uncert{0.0} & --- & 0.0\uncert{0.0} & 0.0\uncert{0.0} \\
 &  & Qwen3-235B-A22B & 12 & 0.0\uncert{0.0} & --- & 0.0\uncert{0.0} & 0.0\uncert{0.0} \\
\end{longtable}
\endgroup

\paragraph{Legal structured-interface failures.}
\label{app:legal-structured-interface-failures}
The reported episode success rate is a strict end-to-end measure. We define a
\textit{structured-interface failure} as an episode that does not produce a
valid final decision because a participant response or final-decision response
cannot be parsed under the required structured response format. A valid final
decision is one that is successfully parsed and passed to the problem-specific
evaluator. Structured-interface failures remain unsuccessful episodes.

Across the primary legal evaluation, each of the three evaluation conditions
contains 1,304 episodes. In 276 multi-agent episodes, a participant response
cannot be parsed and the episode terminates before the final-decision stage.
Four additional episodes reach the final-decision stage but do not produce a
valid final-decision response. Of these 280 structured-interface failures, 254
are produced by the two Claude models: 250 occur during participant interaction
and four at the final-decision stage.

The failure records do not support reliable semantic rescoring. For failed
responses, the archive retains only the first 500 characters and does not
record the complete raw response, provider finish reason, or output-token
count. We therefore cannot reliably distinguish schema noncompliance from
generation truncation or determine what a model would have produced had
generation continued.

\begin{table}[!h]
\centering
\small
\setlength{\tabcolsep}{2pt}
\renewcommand{\arraystretch}{1.08}
\caption{\textbf{Outcome decomposition for the primary legal
evaluation.} Structured-interface failures remain unsuccessful episodes in
the reported episode success rate. The final column is a diagnostic conditional
rate computed only over episodes that produce a valid final decision.}
\label{tab:legal-structured-interface-failures}
\begin{tabular}{@{}lrrrr@{}}
\toprule
Evaluation condition &
\makecell{Strict\\success} &
\makecell{Valid final decision,\\unsuccessful} &
\makecell{Structured-interface\\failure} &
\makecell{Success given a valid\\final decision} \\
\midrule
Single-Agent All-PIU
& 982/1,304 (75.3\%)
& 318/1,304 (24.4\%)
& 4/1,304 (0.3\%)
& 75.5\% \\
Multi-Agent PIU-Only
& 650/1,304 (49.8\%)
& 531/1,304 (40.7\%)
& 123/1,304 (9.4\%)
& 55.0\% \\
Multi-Agent Full-Context
& 404/1,304 (31.0\%)
& 747/1,304 (57.3\%)
& 153/1,304 (11.7\%)
& 35.1\% \\
\bottomrule
\end{tabular}
\end{table}

Among episodes that produce a valid final decision, the corresponding
diagnostic success rates are 75.5\%, 55.0\%, and 35.1\%. The communication and
retrieval gaps therefore remain 20.5 and 19.9 percentage points, respectively.
These conditional rates are descriptive rather than alternative estimates of
end-to-end performance: producing a valid final decision is itself an outcome
that can depend on the model and evaluation condition.

As an additional sensitivity analysis, excluding both Claude models gives
strict aggregate legal episode success rates of 69.8\%, 49.4\%, and 32.3\% in
Single-Agent All-PIU, Multi-Agent PIU-Only, and Multi-Agent Full-Context,
respectively. Conditional on producing a valid final decision, the corresponding
rates are 69.8\%, 49.7\%, and 33.0\%. The strict gaps are 20.4 and 17.1
percentage points, and the conditional gaps are 20.1 and 16.7 points.

In the additional Single-Agent Full-Context evaluation, Claude Sonnet 4.5 and
Claude Haiku 4.5 produce seven and two structured-interface failures,
respectively, at the final-decision stage. These evaluated inputs remain
unsuccessful episodes; serving-context rejections and other evaluation errors
remain excluded from the episode-success denominator as described above.

\subsection{Single-Agent Full-Context Evaluation Counts}
\label{app:cross-domain-single-full-context}

Some Single-Agent Full-Context model inputs exceed the serving endpoint's
context limit or otherwise fail before evaluation. The current coverage audit
spans the 11 source corpora other than DeepSWE; Single-Agent Full-Context was
not run for the DeepSWE cohort.
Table~\ref{tab:cross-domain-single-full-context-coverage} separately reports
evaluated problem-instance--model pairs, context-window failures, output
failures, and other unavailable failures across domains. The percentages use
all fixed problem-instance--model pairs for each domain--model row as the
denominator. Episode-success rates include output failures as unsuccessful
episodes but exclude context-window, infrastructure, and judge failures.
Table~\ref{tab:cross-domain-model-source-corpus-results} reports the resulting
failed/evaluable count and uses an em dash only when no comparable success
estimate is available.

\begin{table}[!h]
\centering
\scriptsize
\setlength{\tabcolsep}{2pt}
\renewcommand{\arraystretch}{1.06}
\caption{\textbf{Single-Agent Full-Context evaluation counts by domain and model.} The table covers the 11 source corpora other than DeepSWE, for which this condition was not run. Percentages in parentheses use all fixed problem-instance--model pairs in each domain--model row. Context-window, infrastructure, and judge failures are excluded from the episode-success denominator; output failures are included as unsuccessful episodes.}
\label{tab:cross-domain-single-full-context-coverage}
\begin{tabular*}{\textwidth}{@{\extracolsep{\fill}}llrrrrr@{}}
\toprule
Domain & Model & Planned & \makecell{Evaluated,\\$n$ (\%)} & \makecell{Context unavailable,\\$n$ (\%)} & \makecell{Output failures,\\$n$ (\%)} & \makecell{Other\\failures, $n$ (\%)} \\
\midrule
Legal & GPT-5.5 & 163 & 163 (100.0) & 0 (0.0) & 0 (0.0) & 0 (0.0) \\
Legal & Claude Sonnet 4.5 & 163 & 156 (95.7) & 0 (0.0) & 7 (4.3) & 0 (0.0) \\
Legal & Claude Haiku 4.5 & 163 & 161 (98.8) & 0 (0.0) & 2 (1.2) & 0 (0.0) \\
Legal & Gemma 3 4B IT & 163 & 147 (90.2) & 16 (9.8) & 0 (0.0) & 0 (0.0) \\
Legal & Gemma 3 12B IT & 163 & 147 (90.2) & 16 (9.8) & 0 (0.0) & 0 (0.0) \\
Legal & Gemma 3 27B IT & 163 & 147 (90.2) & 16 (9.8) & 0 (0.0) & 0 (0.0) \\
Legal & Qwen3-32B & 163 & 60 (36.8) & 100 (61.3) & 0 (0.0) & 3 (1.8) \\
Legal & Qwen3-235B-A22B & 163 & 163 (100.0) & 0 (0.0) & 0 (0.0) & 0 (0.0) \\
\midrule
Finance & GPT-5.5 & 60 & 42 (70.0) & 0 (0.0) & 18 (30.0) & 0 (0.0) \\
Finance & Claude Sonnet 4.5 & 60 & 53 (88.3) & 4 (6.7) & 3 (5.0) & 0 (0.0) \\
Finance & Claude Haiku 4.5 & 60 & 55 (91.7) & 4 (6.7) & 1 (1.7) & 0 (0.0) \\
Finance & Gemma 3 4B IT & 60 & 48 (80.0) & 9 (15.0) & 3 (5.0) & 0 (0.0) \\
Finance & Gemma 3 12B IT & 60 & 51 (85.0) & 9 (15.0) & 0 (0.0) & 0 (0.0) \\
Finance & Gemma 3 27B IT & 60 & 51 (85.0) & 9 (15.0) & 0 (0.0) & 0 (0.0) \\
Finance & Qwen3-32B & 60 & 7 (11.7) & 53 (88.3) & 0 (0.0) & 0 (0.0) \\
Finance & Qwen3-235B-A22B & 60 & 57 (95.0) & 2 (3.3) & 1 (1.7) & 0 (0.0) \\
\midrule
Medical & GPT-5.5 & 74 & 67 (90.5) & 0 (0.0) & 7 (9.5) & 0 (0.0) \\
Medical & Claude Sonnet 4.5 & 74 & 71 (95.9) & 0 (0.0) & 3 (4.1) & 0 (0.0) \\
Medical & Claude Haiku 4.5 & 74 & 72 (97.3) & 0 (0.0) & 2 (2.7) & 0 (0.0) \\
Medical & Gemma 3 4B IT & 74 & 72 (97.3) & 0 (0.0) & 2 (2.7) & 0 (0.0) \\
Medical & Gemma 3 12B IT & 74 & 73 (98.6) & 0 (0.0) & 1 (1.4) & 0 (0.0) \\
Medical & Gemma 3 27B IT & 74 & 74 (100.0) & 0 (0.0) & 0 (0.0) & 0 (0.0) \\
Medical & Qwen3-32B & 74 & 57 (77.0) & 17 (23.0) & 0 (0.0) & 0 (0.0) \\
Medical & Qwen3-235B-A22B & 74 & 74 (100.0) & 0 (0.0) & 0 (0.0) & 0 (0.0) \\
\midrule
SWE & GPT-5.5 & 32 & 21 (65.6) & 11 (34.4) & 0 (0.0) & 0 (0.0) \\
SWE & Claude Sonnet 4.5 & 32 & 1 (3.1) & 22 (68.8) & 0 (0.0) & 9 (28.1) \\
SWE & Claude Haiku 4.5 & 32 & 1 (3.1) & 22 (68.8) & 0 (0.0) & 9 (28.1) \\
SWE & Gemma 3 4B IT & 32 & 2 (6.2) & 30 (93.8) & 0 (0.0) & 0 (0.0) \\
SWE & Gemma 3 12B IT & 32 & 2 (6.2) & 30 (93.8) & 0 (0.0) & 0 (0.0) \\
SWE & Gemma 3 27B IT & 32 & 2 (6.2) & 30 (93.8) & 0 (0.0) & 0 (0.0) \\
SWE & Qwen3-32B & 32 & 0 (0.0) & 30 (93.8) & 0 (0.0) & 2 (6.2) \\
SWE & Qwen3-235B-A22B & 32 & 2 (6.2) & 30 (93.8) & 0 (0.0) & 0 (0.0) \\
\midrule
All domains & All models & 2632 & 2099 (79.7) & 460 (17.5) & 50 (1.9) & 23 (0.9) \\
\bottomrule
\end{tabular*}
\end{table}

Across these 11 source corpora and all models, 2,099 of 2,632 scheduled inputs
(79.7\%) were normally evaluated, 50 (1.9\%) had output failures that remain
in the episode-success denominator, 460 (17.5\%) exceeded a context window,
and 23 (0.9\%) had other unavailable failures. Thus, 2,149 inputs contribute
to Single-Agent Full-Context episode-success estimates.
The Full-Context values reported in
Table~\ref{tab:cross-domain-model-source-corpus-results} therefore distinguish
episode success from evaluation coverage rather than treating unavailable
inputs as incorrect final decisions.

The nine legal Claude entries labeled Output failures in the table, seven for
Sonnet and two for Haiku, are structured-interface failures at the
final-decision stage, as described in
Appendix~\ref{app:legal-structured-interface-failures}, and are included as
unsuccessful episodes rather than treated as serving-context rejections.

\begingroup
\subsection{Synthesis and Error-Uptake Analysis}
\label{app:selected-mechanism-results}

This appendix reports the labeling criteria and transcript interventions for
correct synthesis and error uptake summarized in
Section~\ref{sec:selected-mechanism-results}.

\subsubsection{Definitions and Design}
\label{app:selected-mechanism-design}

A PIU is \emph{faithfully communicated} when a participant message preserves
its meaning without changing its value, entity, scope, qualification, or
effect. An \emph{error} is an incorrect or unsupported sentence or component
that could affect the final decision. \emph{Error uptake} occurs when a later
participant uses an error from a previous visible message to produce a new
sentence; rejecting or correcting the earlier error does not count.
The intended \emph{synthesis} criterion is an inference that combines at least
two distinct PIUs; repeating one PIU or merely proposing an answer does not
count. A synthesis inference is \emph{correct} when it is supported by the
reference PIUs and reference decision. An \emph{explicit answer proposal}
directly states the final choice or requested answer.

The analysis uses 474 frozen Multi-Agent PIU-Only trajectories: 79 problem
instances crossed with \RoundRobin{}, \CurrentSpeakerSelection{}, and
\SimultaneousResponses{}, for Qwen3-32B and Qwen3-235B-A22B participants;
Section~\ref{sec:selected-mechanism-results} abbreviates the latter as
Qwen3-235B. Every participant receives exactly one PIU. A separate GPT-5.5
auditor labels sentence-level components. Participants are not rerun. Instead,
we reconstruct the transcript supplied to the final decision maker and
regenerate the final decision. The intervention results pool \RoundRobin{} and
\CurrentSpeakerSelection{}, giving 158 original trajectories for each
participant-model--final-decision-maker combination; intervention effects are
paired within trajectory, and confidence intervals are paired 95\% bootstrap
intervals.

\subsubsection{Correct-Synthesis Intervention}
\label{app:selected-mechanism-synthesis}

For each original transcript $T$, let $E(T)$ contain the reference versions
of only the PIUs faithfully communicated in $T$. PIUs that are absent or
communicated incorrectly are excluded. Let $S(T)$ contain the correct
synthesis claims generated in $T$, after removing explicit answer proposals.
We compare $E(T)$ with $E(T)+S(T)$. The two inputs therefore contain the same
PIUs and differ only in whether they include correct synthesis.

\begin{table}[H]
\centering
\small
\setlength{\tabcolsep}{4pt}
\renewcommand{\arraystretch}{1.08}
\caption{\textbf{Effect of correct synthesis on episode success.} Effects and
paired 95\% bootstrap confidence intervals are in percentage points.}
\label{tab:selected-mechanism-synthesis}
\begin{tabular}{llrrrr}
\toprule
Participant model & Final decision maker & $E(T)$ & $E(T)+S(T)$ & Effect & 95\% CI \\
\midrule
Qwen3-32B & Qwen3-32B & 54.4\% & 69.0\% & $+14.6$ & $[7.0,22.2]$ \\
Qwen3-32B & GPT-5.5 & 69.6\% & 75.3\% & $+5.7$ & $[1.9,10.1]$ \\
Qwen3-235B-A22B & Qwen3-235B-A22B & 60.1\% & 77.8\% & $+17.7$ & $[10.1,25.3]$ \\
Qwen3-235B-A22B & GPT-5.5 & 78.5\% & 86.7\% & $+8.2$ & $[3.2,13.3]$ \\
\bottomrule
\end{tabular}
\end{table}

Correct synthesis improves episode success in all four comparisons. The effect
is larger for the Qwen3 final decision maker than for GPT-5.5 for both
participant models. Because $S(T)$ excludes explicit answer proposals, this
effect cannot be attributed to direct answer copying.

\paragraph{Scope qualification.}
The intended synthesis definition requires at least two distinct PIUs. The
auditor-labeled set used here also includes some inferences that combine one
PIU with information from the publicly visible decision options. Accordingly,
the reported effects apply to this broader auditor-labeled set and should not
be interpreted as isolating synthesis across distinct PIUs.

\subsubsection{Error-Uptake Intervention}
\label{app:selected-mechanism-error-uptake}

The control is the reconstructed original transcript. In the intervention, we
remove later sentences produced using an error from a previous visible message,
while retaining the sentence containing the original error. This isolates the
additional effect of later participants taking up the error. A positive effect
means that removing error-uptake sentences improves episode success.

\begin{table}[H]
\centering
\small
\setlength{\tabcolsep}{5pt}
\renewcommand{\arraystretch}{1.08}
\caption{\textbf{Effect of removing error uptake.} Effects and paired 95\%
bootstrap confidence intervals are in percentage points.}
\label{tab:selected-mechanism-error-uptake}
\begin{tabular}{llrr}
\toprule
Participant model & Final decision maker & Effect & 95\% CI \\
\midrule
Qwen3-32B & Qwen3-32B & $+2.5$ & $[-1.3,6.3]$ \\
Qwen3-32B & GPT-5.5 & $+1.3$ & $[-2.5,5.1]$ \\
Qwen3-235B-A22B & Qwen3-235B-A22B & $+5.7$ & $[1.9,10.1]$ \\
Qwen3-235B-A22B & GPT-5.5 & $+7.6$ & $[3.2,12.7]$ \\
\bottomrule
\end{tabular}
\end{table}

Removing error uptake improves episode success for Qwen3-235B-A22B
participants under both final decision makers. The Qwen3-32B estimates have the
same sign, but their confidence intervals include zero. These effects are
smaller than the corresponding correct-synthesis effects in all four
comparisons and do not by themselves explain the communication-structure
ordering. We do not apply the error-uptake intervention to
\SimultaneousResponses{} because participants cannot observe one another;
incorrect or inconsistent simultaneous messages are independent rather than
propagated errors.

\subsubsection{Explicit Answer Proposals}
\label{app:selected-mechanism-answer-proposals}

As a separate check, we remove explicit answer proposals from the original
\RoundRobin{} and \CurrentSpeakerSelection{} transcripts. The effect is $-5.7$
percentage points for Qwen3-32B participants and final decision makers and
$-10.8$ points for Qwen3-235B-A22B participants and final decision makers.
Explicit answer proposals therefore also help the Qwen3 final decision makers.
They cannot, however, explain the correct-synthesis results in
Table~\ref{tab:selected-mechanism-synthesis}, where such proposals are excluded
from $S(T)$.

Overall, adding correct synthesis improves episode success in all four
comparisons. Removing error uptake has smaller effects, with confidence
intervals excluding zero only for Qwen3-235B-A22B participants. The synthesis
effects are smaller with GPT-5.5 than with the Qwen3 final decision makers.
\FloatBarrier
\endgroup

\FloatBarrier
\subsection{PIU-Distribution--Communication-Structure Interaction Estimates}
\label{app:cross-domain-riu-communication-interaction-estimates}

Table~\ref{tab:cross-domain-riu-communication-interactions}
reports the complete probability-scale interaction estimates from the
problem-family and model fixed-effects specifications described in
Section~\ref{sec:cross-domain-riu-distribution-communication-structure-results}.

For PIU-distribution condition $\ell$ and communication structure $c$, the
reported value is the average difference-in-differences in predicted episode
success probability:
\[
100\,\mathbb{E}\!\left[
  \{p_e(\ell,c)-p_e(\ell,c_0)\}
  -\{p_e(\ell_0,c)-p_e(\ell_0,c_0)\}
\right],
\]
where $p_e(\ell,c)$ is episode $e$'s predicted success probability under
$(\ell,c)$, $\ell_0$ and $c_0$ are the PIU-distribution and
communication-structure reference categories, and the expectation averages
over the evaluated models and problem instances.

\begin{table}[H]
\centering
\footnotesize
\setlength{\tabcolsep}{6pt}
\renewcommand{\arraystretch}{1.06}
\caption{\textbf{PIU-distribution--communication-structure interactions across the evaluated problem families.}
Entries are average interaction estimates in episode-success percentage points; smaller $\pm$ terms are problem-instance-clustered standard errors. Rows and columns are relative to \OneRIUPerParticipant{} and \CoordinatorSelectedSpeaker{}, respectively. \revised{Asterisks mark estimates whose 95\% confidence intervals exclude zero.}}
\label{tab:cross-domain-riu-communication-interactions}
\begin{tabular}{@{}cccccc@{}}
\toprule
\makecell{Evaluation\\condition} & \makecell{PIU-distribution\\condition} & \BroadcastQueryCell{} & \RoundRobinCell{} & \CurrentSpeakerSelectionCell{} & \SimultaneousResponsesCell{} \\
\midrule
\multirow{4}{*}{\makecell{Multi-Agent\\PIU-Only\\{\scriptsize $N=50,438$}\\{\scriptsize joint $p<.001$}}} & \PairwiseRIUOverlapCell & $4.5$\uncert{1.4}\textsuperscript{*} & $4.4$\uncert{1.5}\textsuperscript{*} & $4.5$\uncert{1.9}\textsuperscript{*} & $1.2$\uncert{1.5} \\
 & \ExtraParticipantWithoutRIUCell & $0.6$\uncert{1.5} & $1.8$\uncert{1.3} & $-0.3$\uncert{1.6} & $-1.1$\uncert{1.6} \\
 & \OneCommonRIU{} & $-0.6$\uncert{1.5} & $1.3$\uncert{1.5} & $2.7$\uncert{1.8} & $-3.3$\uncert{1.5}\textsuperscript{*} \\
 & \AllButOneRIUConcentrationCell & $-1.1$\uncert{1.6} & $4.3$\uncert{1.5}\textsuperscript{*} & $6.5$\uncert{2.1}\textsuperscript{*} & $-0.4$\uncert{1.6} \\
\midrule
\multirow{4}{*}{\makecell{Multi-Agent\\Full-Context\\{\scriptsize $N=48,499$}\\{\scriptsize joint $p<.001$}}} & \PairwiseRIUOverlapCell & $0.4$\uncert{1.2} & $2.3$\uncert{1.2} & $-1.5$\uncert{1.5} & $0.9$\uncert{1.3} \\
 & \ExtraParticipantWithoutRIUCell & $-1.7$\uncert{1.3} & $-0.2$\uncert{1.2} & $-2.3$\uncert{1.3} & $-0.6$\uncert{1.1} \\
 & \OneCommonRIU{} & $-0.4$\uncert{1.4} & $0.6$\uncert{1.2} & $-3.6$\uncert{1.5}\textsuperscript{*} & $-1.4$\uncert{1.4} \\
 & \AllButOneRIUConcentrationCell & $-0.9$\uncert{1.5} & $-0.9$\uncert{1.3} & $-0.8$\uncert{1.6} & $-4.0$\uncert{1.5}\textsuperscript{*} \\
\bottomrule
\end{tabular}
\end{table}

\FloatBarrier
\subsection{Robustness Analysis with Problem-Instance Fixed Effects}
\label{app:problem-instance-riu-distribution-communication-structure-results}

The main analysis in
Section~\ref{sec:cross-domain-riu-distribution-communication-structure-results}
includes problem-family and model fixed effects. As a stricter
within-problem-instance comparison, we replace the problem-family fixed effects
with problem-instance fixed effects:
\[
y_e \sim \operatorname{Bernoulli}(p_e), \qquad
\operatorname{logit}(p_e)
= \beta_0 + \alpha_{m_e} + \delta_{q_e} + \pi_{\ell_e}
  + \lambda_{c_e} + \omega_{\ell_e,c_e}.
\]
Because each problem instance belongs to exactly one problem family, the
problem-instance fixed effects absorb the problem-family fixed effects rather
than entering the model alongside them. In both analyses, we assign episode
success of zero to 3,448 attempted episodes that failed and exclude unavailable cells. The
reference categories, weighting, and problem-instance-clustered standard errors
are otherwise the same as in the main analysis.

\begin{table}[H]
\centering
\footnotesize
\setlength{\tabcolsep}{6pt}
\renewcommand{\arraystretch}{1.06}
\caption{\textbf{Robustness analysis with problem-instance fixed effects.} Entries are average interaction estimates in episode-success percentage points; smaller $\pm$ terms are problem-instance-clustered standard errors. Columns replace \CoordinatorSelectedSpeaker{}, and the four displayed PIU-distribution rows replace \OneRIUPerParticipant{}. \revised{Asterisks mark estimates whose 95\% confidence intervals exclude zero.}}
\label{tab:problem-instance-riu-communication-interactions}
\begin{tabular}{@{}cccccc@{}}
\toprule
\makecell{Evaluation\\condition} & \makecell{PIU-distribution\\condition} & \BroadcastQueryCell{} & \RoundRobinCell{} & \CurrentSpeakerSelectionCell{} & \SimultaneousResponsesCell{} \\
\midrule
\multirow{4}{*}{\makecell{Multi-Agent\\PIU-Only\\{\scriptsize $N=50,438$}\\{\scriptsize joint $p<.001$}}} & \PairwiseRIUOverlapCell & $4.8$\uncert{1.5}\textsuperscript{*} & $4.7$\uncert{1.5}\textsuperscript{*} & $4.6$\uncert{1.9}\textsuperscript{*} & $1.6$\uncert{1.5} \\
 & \ExtraParticipantWithoutRIUCell & $0.6$\uncert{1.5} & $1.9$\uncert{1.3} & $-0.3$\uncert{1.6} & $-1.0$\uncert{1.6} \\
 & \OneCommonRIU{} & $-0.5$\uncert{1.5} & $1.5$\uncert{1.5} & $2.7$\uncert{1.8} & $-2.9$\uncert{1.6} \\
 & \AllButOneRIUConcentrationCell & $-1.1$\uncert{1.6} & $4.2$\uncert{1.5}\textsuperscript{*} & $6.3$\uncert{2.1}\textsuperscript{*} & $-0.3$\uncert{1.6} \\
\midrule
\multirow{4}{*}{\makecell{Multi-Agent\\Full-Context\\{\scriptsize $N=48,499$}\\{\scriptsize joint $p<.001$}}} & \PairwiseRIUOverlapCell & $0.4$\uncert{1.2} & $2.4$\uncert{1.2}\textsuperscript{*} & $-1.5$\uncert{1.6} & $1.0$\uncert{1.3} \\
 & \ExtraParticipantWithoutRIUCell & $-1.8$\uncert{1.3} & $-0.2$\uncert{1.2} & $-2.3$\uncert{1.2} & $-0.7$\uncert{1.1} \\
 & \OneCommonRIU{} & $-0.4$\uncert{1.4} & $0.6$\uncert{1.2} & $-3.6$\uncert{1.5}\textsuperscript{*} & $-1.3$\uncert{1.4} \\
 & \AllButOneRIUConcentrationCell & $-1.0$\uncert{1.5} & $-0.9$\uncert{1.3} & $-0.8$\uncert{1.6} & $-3.9$\uncert{1.5}\textsuperscript{*} \\
\bottomrule
\end{tabular}
\end{table}

The problem-instance fixed effects absorb differences in absolute difficulty
among the evaluated problem instances. The reported interactions therefore
compare changes across PIU distributions and communication structures within
the same problem instances. The same five positive Multi-Agent PIU-Only
interactions and the same two negative Multi-Agent Full-Context interactions have
95\% confidence intervals that exclude zero as in the main analysis, with
similar point estimates. The positive \PairwiseRIUOverlap{}--\RoundRobin{}
interaction in Multi-Agent Full-Context has a 95\% confidence interval that
excludes zero only with problem-instance fixed effects. The negative \OneCommonRIU{}--\SimultaneousResponses{}
Multi-Agent PIU-Only interaction has a 95\% confidence interval that includes
zero with problem-instance fixed effects.

\begingroup
\captionsetup{labelfont={color=black},textfont={color=black}}
\FloatBarrier
\subsection{Interpreting PIU-Distribution--Communication-Structure Interactions}
\label{app:riu-distribution-communication-structure-results-interpretation}

This appendix reports the matched comparisons and interventions accompanying
Section~\ref{sec:cross-domain-riu-distribution-communication-structure-results}.
The cross-domain interaction estimates and problem-instance fixed-effects
analysis remain in Appendix~\ref{app:cross-domain-riu-communication-interaction-estimates}
and Appendix~\ref{app:problem-instance-riu-distribution-communication-structure-results}.
The finance experiments vary what a coordinator query asks the selected
participant to communicate, including by appending public PIU category labels
for its assigned PIUs and an instruction to report evidence for each.
Speaker selection is held fixed in these interventions. The SARA experiments vary who receives private
source material and which participant messages the final decision maker sees.
These experiments use separate problem-instance sets and do not estimate the
fraction of the cross-domain regression interaction attributable to either
mechanism. All effects below concern episode success unless stated otherwise;
percentage-point estimates are calculated before rounding the displayed rates.

\subsubsection{Finance: Design and Query Interventions}
\label{app:piu-interaction-finance-design}

The matched communication comparison uses 59 finance problem instances:
30 FinDVer and 29 FinLongDocQA instances. The primary comparison pools Qwen3-32B and
Qwen3-235B-A22B, with the same model used for the participants
and the coordinator or separate final decision maker within an episode.
The design crosses \OneRIUPerParticipant{} and
\AllButOneRIUConcentration{} with \CoordinatorSelectedSpeaker{} and
\RoundRobin{}, giving 472 primary episodes and 118
problem-instance--model pairs per condition. Participants receive private
source material. The interaction
compares the difference in episode success between
\RoundRobin{} and \CoordinatorSelectedSpeaker{} under
\AllButOneRIUConcentration{} with the same difference under
\OneRIUPerParticipant{}.

\paragraph{Participant descriptions.}
The original public participant descriptions summarize a participant's
subject matter or function, without enumerating the public information
categories corresponding to its assigned PIUs. These categories describe
what the evidence concerns; they do not disclose the evidence itself.
Across the 59 problem instances, their mean
lengths are 4.41 words under \OneRIUPerParticipant{} and 5.10 words under
\AllButOneRIUConcentration{}. The 54 participants assigned multiple PIUs
have six-word descriptions but a mean of 3.43 public information categories
(range 2--6). A brief description thus need not indicate every category of evidence
available from a participant assigned multiple PIUs.

\paragraph{Experimental logic.}
The observed communication difference permits two explanations: the
coordinator may fail to select a participant assigned a needed PIU, or its
query may limit the evidence that the selected participant reports.
Near-complete coverage of required participants motivates the second account.
Under \CoordinatorSelectedSpeaker{}, the selected participant receives a
specific query from the coordinator; under \RoundRobin{}, there is no
coordinator query and the participant responds to the whole decision problem.
A query covering one PIU may suffice when the participant holds only that PIU,
but may omit other assigned PIUs when several are concentrated with it.
If this account is correct, broadening the requested evidence should improve episode
success more when one selected participant holds multiple PIUs. We therefore
change query content while retaining the source episode's speaker selection.

\paragraph{Fixed speaker selection.}
All query interventions operate within \CoordinatorSelectedSpeaker{}.
The original coordinator query is the task-specific question recorded in
the source episode.
For each source episode, we replay the recorded sequence of selected
participants and the number of responses. The source material, model,
PIU distribution, resource budgets, and final-decision procedure are held
fixed across query conditions; participant responses and final decisions are
regenerated. Original coordinator queries are not regenerated in response to the new
participant messages. The resulting effects apply to this fixed sequence,
rather than to a coordinator that adapts its later selections or queries.

\paragraph{Public information categories in the query.}
After speaker selection, one intervention appends a numbered list of
the public PIU category labels corresponding to the selected participant's
assigned PIUs and an instruction to report
the supporting facts or numbers for each category. The added list contains
short public PIU category labels, without private source text, detailed evidence,
values, answers, entity scope, or period scope. It augments the original query;
it does not replace the participant description or alter speaker selection.
For example, an original description is ``service proceeds and cybersecurity
operations reviewer.'' The added list includes service agreement proceeds details,
cybersecurity operations exposure, liquidity going-concern assessment,
cryptocurrency market exposure, earnings profitability outlook, and
share-based compensation expense. These labels specify the topics of the
assigned evidence, without supplying the proceeds, expenses, or other private
facts needed to answer the problem. Categories need not correspond one-to-one
to PIUs: several PIUs may concern the same category.

The added block averages 50.36 words for participants assigned one PIU and
61.00 words for participants assigned multiple PIUs. It is visible to the
selected participant and remains in the public transcript. The treatment
jointly changes the available category labels and the explicit elicitation
instruction; their separate effects are not identified.

The primary comparison with and without these additions contains
59 instances $\times$ 2 Qwen3 models $\times$ 2 PIU distributions $\times$
2 query conditions $\times$ 3 repetitions, or 1,416 episodes.
This analysis plan was frozen after an earlier generic-instruction experiment,
rather than preregistered before all related evidence was observed.

\paragraph{Removing the task-specific query.}
To test whether the additions are necessary or removing the narrow query
suffices, we replace each original coordinator query with ``It is your turn
to contribute,'' both with and without the public PIU category labels and
instruction. This produces 1,416 additional primary episodes. Together with
the 1,416 episodes retaining the original queries, the $2\times2$ design
crosses query retention versus removal with presence versus absence of the
additions under each PIU distribution, giving 2,832 episodes. The
original-query comparators are reused from the preceding experiment;
they are not generated again for the query-removal comparison.

\paragraph{Communication-structure interaction with Round robin held fixed.}
For each of the four coordinator query conditions, we retain the 236 original
Qwen3 \RoundRobin{} episodes and replace only the
\CoordinatorSelectedSpeaker{} outcomes with the corresponding
repetition-averaged replay outcomes. The intervention operates under both
PIU distributions; \OneRIUPerParticipant{} controls for a general benefit
from changing the query. Let $G_R$ be the difference in episode success
between \AllButOneRIUConcentration{} and \OneRIUPerParticipant{} under
\RoundRobin{}, and let $G_C(k)$ be that difference under
\CoordinatorSelectedSpeaker{} with query condition $k$. In percentage points,
the communication-structure interaction is
\[
I(k)=G_R-G_C(k).
\]
The fixed $G_R$ is $-12.7$ points. Consequently, a selective improvement in
the concentrated coordinator condition reduces $I(k)$ by the same amount.
This calculation uses the same primary Qwen3 population and adds no episodes.
It distinguishes the original single-generation comparison from the
three-repetition coordinator replay; their baseline estimates need not be
identical. The original Round robin and replay coordinator outcomes are resampled
together for inference on each interaction and its change.

\paragraph{Budgets, outcomes, and inference.}
Follow-up generation uses temperature 0.2, a 40,000-token private-context
budget per participant, 600-token participant and coordinator utterance caps,
and a 1,200-token final rationale budget. Complete queries including the
public PIU category labels and instruction satisfy the existing 600-token cap. A temperature-0 GPT-5.5
evaluator scores episode success and a continuous score on a 0--1 scale.
A separate temperature-0 GPT-5.5 semantic evaluation compares each PIU's
reference evidence with its assigned participant's public messages.
Substantive communication requires accurate content-bearing evidence;
complete communication additionally requires all key task-relevant elements.
Repeating a category label alone does not count.

For both follow-up experiments, three repetitions are averaged within each
problem-instance--model--PIU-distribution--query-condition combination.
All primary confidence intervals use 20,000 percentile bootstrap draws over
the 59 problem-instance clusters, retaining both models and all matched
conditions together. For the fixed-Round-robin analysis, each draw recomputes
$I(k)$ and the difference between query conditions. The original communication
comparison uses the same cluster-bootstrap procedure. No instance or
repetition is excluded based on its observed outcome.

\subsubsection{Finance: Results and Process Evidence}
\label{app:piu-interaction-finance-results}

The primary communication comparison and query-intervention analyses pool
Qwen3-32B and Qwen3-235B-A22B across the same 59 finance problem instances.
The original comparison uses one generation per condition; the query
interventions average three repetitions. Table~\ref{tab:piu-interaction-finance-original}
reports the original comparison, and the subsequent analysis recalculates its
interaction while holding the original Round robin outcomes fixed.

\begin{table}[H]
\centering
\small
\setlength{\tabcolsep}{5pt}
\caption{\textbf{Matched finance communication comparison.} Episode success
rates pool the two Qwen3 models across 59 problem instances. Effects are in
percentage points.}
\label{tab:piu-interaction-finance-original}
\begin{tabular}{@{}lrrr@{}}
\toprule
PIU distribution & \makecell{Coordinator-selected\\speaker} & \RoundRobinCell{} & Difference \\
\midrule
\OneRIUPerParticipant{} & 62.7\% & 80.5\% & $+17.8$ \\
\AllButOneRIUConcentration{} & 36.4\% & 67.8\% & $+31.4$ \\
\midrule
\multicolumn{3}{l}{Interaction (95\% CI)} & \makecell{$+13.6$\\$[0.8,26.3]$} \\
\bottomrule
\end{tabular}
\end{table}

Required participants speak in 97.2\% of cases under
\OneRIUPerParticipant{} and 99.2\% under \AllButOneRIUConcentration{} with
\CoordinatorSelectedSpeaker{}; both \RoundRobin{} rates are 100\%.
This weakens an explanation based solely on failing to select a participant
assigned a needed PIU.

\paragraph{Communication-structure interaction with Round robin held fixed.}
Table~\ref{tab:piu-interaction-finance-reconstructed} reports the interaction
for each coordinator query condition and its change from the original-query
replay. The replay baseline is $+15.5$ points (95\% CI $[3.4,27.4]$), compared
with $+13.6$ ($[0.8,26.3]$) in the original comparison. Adding the labels and
instruction reduces it by $9.6$ points; query removal reduces it by $20.9$
without additions and $24.9$ with them relative to that same baseline.
All three changes exclude zero, while all three residual interaction intervals
include zero. The latter does not establish equivalence to zero or a reversal
of the communication-structure ordering.

\begin{table}[H]
\centering
\small
\setlength{\tabcolsep}{4pt}
\caption{\textbf{Finance communication-structure interactions with Round robin held fixed.}
The original Qwen3 Round robin allocation difference is $-12.7$ points.
Interactions subtract the coordinator allocation difference from this fixed
reference. Changes compare each condition with the original query without
additions. Entries are percentage points with 95\% paired problem-instance
bootstrap intervals. \revised{Asterisks mark statistically significant effects.}}
\label{tab:piu-interaction-finance-reconstructed}
\begin{tabular}{@{}lrr@{}}
\toprule
Coordinator query condition & Interaction (95\% CI) & Change (95\% CI) \\
\midrule
\makecell[l]{Original query\\without additions} & \makecell{$+15.5^{*}$\\$[3.4,27.4]$} & Reference \\
\arrayrulecolor{black!18}\specialrule{0.3pt}{0.4pt}{0.4pt}\arrayrulecolor{black}
\makecell[l]{Original query\\with additions} & \makecell{$+5.9$\\$[-5.1,16.4]$} & \makecell{$-9.6^{*}$\\$[-16.9,-2.3]$} \\
\arrayrulecolor{black!18}\specialrule{0.3pt}{0.4pt}{0.4pt}\arrayrulecolor{black}
\makecell[l]{Query removed\\without additions} & \makecell{$-5.4$\\$[-14.7,3.7]$} & \makecell{$-20.9^{*}$\\$[-30.2,-11.3]$} \\
\arrayrulecolor{black!18}\specialrule{0.3pt}{0.4pt}{0.4pt}\arrayrulecolor{black}
\makecell[l]{Query removed\\with additions} & \makecell{$-9.3$\\$[-20.1,0.8]$} & \makecell{$-24.9^{*}$\\$[-34.7,-14.7]$} \\
\bottomrule
\end{tabular}
\end{table}

\paragraph{Intervention effects within Coordinator-selected speaker.}
Table~\ref{tab:piu-interaction-finance-query} reports success rates in the
four query conditions. Table~\ref{tab:piu-interaction-finance-intervention-contrasts}
reports PIU-distribution--intervention interactions: each intervention's
success effect under \AllButOneRIUConcentration{} minus its effect under
\OneRIUPerParticipant{}. With the Round robin outcomes held fixed, these
effects reduce the communication-structure interaction by the same amount.
For query removal with additions, the comparison retains the additions in
both conditions; it therefore differs from the combined intervention's change
relative to the original query without additions.

\begin{table}[H]
\centering
\small
\setlength{\tabcolsep}{5pt}
\caption{\textbf{Finance query interventions with fixed speaker selection.}
Episode success rates pool the two Qwen3 models, averaging three repetitions.
``With additions'' appends public PIU category labels corresponding to assigned PIUs
and an instruction to report supporting evidence for each category. ``Coordinator query removed'' replaces the original task-specific question
with ``It is your turn to contribute,'' while retaining speaker selection.}
\label{tab:piu-interaction-finance-query}
\begin{tabular}{@{}lrrrr@{}}
\toprule
& \multicolumn{2}{c}{\makecell{Original coordinator\\query}} & \multicolumn{2}{c}{\makecell{Coordinator query\\removed}} \\
\cmidrule(lr){2-3}\cmidrule(lr){4-5}
PIU distribution & \makecell{Without\\additions} & \makecell{With\\additions} & \makecell{Without\\additions} & \makecell{With\\additions} \\
\midrule
\OneRIUPerParticipant{} & 65.5\% & 69.2\% & 71.5\% & 74.3\% \\
\AllButOneRIUConcentration{} & 37.3\% & 50.6\% & 64.1\% & 70.9\% \\
\bottomrule
\end{tabular}
\end{table}

\begin{table}[H]
\centering
\small
\setlength{\tabcolsep}{5pt}
\caption{\textbf{Finance intervention interactions across PIU distributions.}
Each interaction subtracts the intervention effect under \OneRIUPerParticipant{}
from that under \AllButOneRIUConcentration{}, pooling the two Qwen3 models.
Effects and 95\% problem-instance bootstrap confidence intervals are in
percentage points. \revised{Asterisks mark statistically significant interactions.} Estimates use unrounded rates.}
\label{tab:piu-interaction-finance-intervention-contrasts}
\begin{tabular}{@{}lrr@{}}
\toprule
Comparison & Interaction & 95\% CI \\
\midrule
Add labels and instruction to original coordinator query & $+9.6^{*}$ & $[2.3,16.9]$ \\
\arrayrulecolor{black!18}\specialrule{0.3pt}{0.4pt}{0.4pt}\arrayrulecolor{black}
Remove coordinator query without additions & $+20.9^{*}$ & $[11.6,30.2]$ \\
\arrayrulecolor{black!18}\specialrule{0.3pt}{0.4pt}{0.4pt}\arrayrulecolor{black}
Remove coordinator query with additions & $+15.3^{*}$ & $[7.9,22.6]$ \\
\bottomrule
\end{tabular}
\end{table}

\paragraph{Removing the task-specific query.}
Query removal improves success more under \AllButOneRIUConcentration{}
both without and with additions (Table~\ref{tab:piu-interaction-finance-intervention-contrasts}).
This direction also holds within both source corpora: the intervention
interactions without additions are $+23.9$ points in FinDVer
(95\% CI $[11.7,36.7]$) and $+17.8$ in FinLongDocQA ($[4.0,31.6]$);
with additions, they are $+16.7$ ($[6.1,26.7]$) and $+13.8$
($[3.4,23.6]$), respectively.

Without the labels and instruction, the pooled-Qwen3 allocation gap decreases from
28.2 points (95\% CI $[18.1,38.4]$) with original coordinator queries to 7.3 points
($[-0.6,15.3]$) with the coordinator query removed. With the labels and instruction, it decreases
from 18.6 points ($[10.7,26.8]$) to 3.4 points ($[-3.1,9.3]$).
These point estimates correspond to 74\% and 82\% attenuation of the
respective original-query allocation gaps. They are not proportions of the
original $+13.6$-point communication interaction explained. The residual
intervals include zero, which does not establish statistical equivalence.

\paragraph{Combining the interventions.}
The query-removal interaction is $+20.9$ points without additions and $+15.3$
with them. Their difference, the PIU-distribution--query-removal--additions
interaction, is $+5.6$ points (95\% CI $[-4.2,15.5]$). Adding labels and an
instruction under query removal has an interaction of $+4.0$ points
($[-4.5,12.4]$). Neither interval excludes zero. The combined condition has
the smallest allocation gap descriptively, but these comparisons do not
establish separate, additive, or synergistic effects of the two interventions.

\paragraph{Communicated evidence.}
Adding the public PIU category labels and instruction selectively increases substantive PIU communication under
\AllButOneRIUConcentration{} relative to \OneRIUPerParticipant{}, with an
interaction of $+6.3$ points (95\% CI $[4.0,8.8]$); the complete-communication
interaction is $+3.6$ points ($[0.3,7.1]$). Query removal yields substantive
communication interactions of $+9.5$ points without the labels and instruction
($[5.7,13.5]$) and $+8.5$ with the labels and instruction ($[5.8,11.3]$).
The corresponding participant-word-count interactions are $+4.3$ words per
response ($[-0.1,8.6]$) and $-6.3$ ($[-11.7,-1.3]$).
Thus, selective improvement in substantive communication does not require
selectively longer responses in both comparisons.

\paragraph{Supporting qualitative audit.}
An outcome-selected audit covers all three actor models, including GPT-5.5.
It examines the 47 original problem-instance--model pairs in which
\CoordinatorSelectedSpeaker{} failed and \RoundRobin{} succeeded under
\AllButOneRIUConcentration{}. A temperature-0 GPT-5.5 auditor saw both
concentrated transcripts and the corresponding \OneRIUPerParticipant{}
transcripts. Condition names and success labels were withheld, while
transcript structure remained visible. The auditor attributed the primary
difference to queries limiting response scope in 44 pairs and to a mixture
of mechanisms in three. These selected frequencies are supporting process
evidence, rather than population mechanism frequencies or causal mediation
estimates.

\paragraph{Selected episode.}
For an HLI 2024 10-Q problem instance with Qwen3-235B-A22B, the original coordinator query
asked about Financial Restructuring fee recognition. The selected participant
also held evidence about acquisition consideration, income-tax factors, and
the quarterly compensation ratio. Its five PIUs corresponded to four public
information categories, since two concerned fee recognition. The added list
named these categories and the instruction requested evidence for each; it did not supply
the facts reported below. Without the public PIU category labels and instruction, two of five PIUs
were substantively communicated in all three repetitions and all final
decisions failed. With the public PIU category labels and instruction, all five were communicated, including
the 63.4\% compensation ratio, the tax-rate change from 20.2\% to 26.9\%
and its explanation, and acquisition consideration of \$14.5M plus \$4.0M.
All three final decisions succeeded. This selected episode illustrates the
intervention; the effect estimates use all 59 problem instances.

\paragraph{Scope.}
\label{app:piu-interaction-finance-scope}
The fixed-selection interventions identify effects of query content on
participant communication and episode success. They do not show that richer
participant descriptions improve the coordinator's selection or its initially
generated query. The public PIU category labels and instruction are supplied after selection
and their effects are not separated. The design does not separate
their effect on the selected participant from their effect on later readers of
the transcript. Holding the original Round robin outcomes fixed permits direct estimation
of changes in the communication-structure interaction within the same Qwen3
population. The appropriate baseline is the original-query replay, rather
than the original single-generation coordinator outcomes. These results
support a role for query content in the interaction, but do not establish
complete causal mediation, equivalence of the residual interaction to zero,
or generalization to an adaptive coordinator or the cross-domain regression.

\FloatBarrier
\subsubsection{SARA: Source-Material and Transcript Interventions}
\label{app:piu-interaction-sara-design}

Three experiments use the same 45 SARA problem instances and Qwen3-32B,
Qwen3-235B-A22B, and GPT-5.5. The longest instance,
\texttt{sara\_subproof\_tax\_case\_90\_taxable\_income}, was designated for
development checks and excluded from all primary analyses, leaving 44
problem instances. Each configuration is repeated three times.
Within an episode, participants and the coordinator or separate final
decision maker use the same model. All actors see public participant
descriptions generated from public information categories that identify
the subject matter without revealing private facts or answers. The
descriptions contain 3--6 words (mean 3.72). Their availability is held
fixed; the experiments do not estimate their causal effect or whether every
actor uses them correctly.

Under \OneRIUPerParticipant{}, every participant is assigned one PIU.
Under \AllButOneRIUConcentration{}, two participants receive all PIUs;
the others receive none. The participant assigned all but one receives
5--11 PIUs (mean 7.73), and 4--10 participants per instance receive none
(mean 6.71). Participant messages have a 600-token cap and private contexts
have a 24,000-token cap per participant. Actor temperature is 0.2; episode-success and
semantic evaluation use GPT-5.5 at temperature 0. Speaker order and initial
speaker are fixed within problem instance, PIU distribution, and repetition,
and reused across models and evidence conditions. The participant assigned
all but one PIU speaks first in only 16 of 132 primary concentrated-allocation
instance--repetition combinations.

\paragraph{Exact PIUs with or without source material.}
Source material may impair the processing of participants assigned PIUs,
or generate additional messages from participants assigned no PIU that impair
the final decision. To examine these pathways, we compare Multi-Agent PIU-Only with a condition
that adds the original private source material. In both conditions,
participants receive their assigned PIUs explicitly as facts relevant to the
decision; the added-source condition also supplies source material to
participants assigned no PIU. This tests whether source material changes the
interaction when the relevant facts are already supplied directly. In
Multi-Agent Full-Context, participants instead receive source material from
which they must retrieve their assigned PIUs. Explicit availability does not
ensure that participants communicate or use the facts correctly.
The comparison crosses two PIU distributions, all five communication
structures, and these two evidence conditions.
The complete design contains 8,100 episodes (45 instances $\times$ 3 models
$\times$ 2 evidence conditions $\times$ 2 PIU distributions $\times$
5 structures $\times$ 3 repetitions), with 7,920 primary episodes over
44 instances.

For communication structure $c$ and evidence condition $v$, define
\[
A(v,c)=100\{\operatorname{Success}(\text{All-but-one},v,c)
-\operatorname{Success}(\text{One PIU per participant},v,c)\}.
\]
The source-material interaction is
$A(\text{PIUs and source material},c)-A(\text{Multi-Agent PIU-Only},c)$.
A negative value means that source material makes PIU concentration more
harmful relative to \OneRIUPerParticipant{} under that structure.

The communication-structure interaction compares these allocation effects:
\[
I(v)=A(v,\text{Simultaneous responses})
-A(v,\text{Coordinator-selected speaker}).
\]
The change reported in Section~\ref{sec:piu-interaction-sara-example} is
$I(\text{PIUs and source material})-I(\text{Multi-Agent PIU-Only})$.
It therefore compares the source-material interactions between the two
communication structures, rather than reporting the interaction within
\SimultaneousResponses{} alone.

\paragraph{Who receives source material.}
The preceding comparison changes source-material access for both groups
together. It therefore cannot distinguish a loss originating with
participants assigned PIUs from one originating with participants assigned no PIU. The source-material intervention independently varies source-material access for the two
participants assigned PIUs and for the participants assigned no PIU under
\AllButOneRIUConcentration{}. Exact PIUs, public descriptions, models,
repetitions, and speaker order remain fixed. We compare no added source
material, source material for participants assigned PIUs only, source material
for participants assigned no PIU only, and source material for both groups.
The experiment covers \CoordinatorSelectedSpeaker{}, \BroadcastQuery{},
and \SimultaneousResponses{}, generating 2,430 new episodes, of which
2,376 concern the 44 primary instances. The no-source and both-group conditions reuse the preceding comparison.
Under \OneRIUPerParticipant{}, there are no participants assigned no PIU,
so the corresponding source-material conditions are identical rather than
independent additional observations.

\paragraph{Which messages reach the final decision maker.}
Source-material placement changes a participant's local processing and
public messages together. Even if source material for participants assigned no PIU causes the larger loss, this does not establish whether their messages
impair final synthesis. If the loss is transmitted through those messages,
hiding them only at the final-decision stage should improve episode success
when their authors receive source material, with little improvement when
they do not. The transcript intervention tests this prediction using frozen
\AllButOneRIUConcentration{}--\SimultaneousResponses{}
trajectories from the condition with no added source material and the condition
with source material only for participants assigned no PIU. For each
trajectory, the final decision maker receives either the complete transcript
or only messages from the two participants assigned PIUs. Participant messages,
descriptions, PIU assignment, and speaker order remain fixed; only the final
decision is regenerated. Both visibility conditions use the same replay
implementation, rather than comparing newly generated decisions with archived
ones. A separate complete-transcript replay checks stability against archived
outcomes.

This design contains 1,620 episodes (45 instances $\times$ 3 models $\times$
2 source-material conditions $\times$ 2 message-visibility conditions $\times$
3 repetitions), with 1,584 primary episodes over 44 instances. The primary
interaction is the effect of removing messages from participants assigned
no PIU when those participants receive source material minus the same removal
effect when they do not. Confidence intervals use bootstrap sampling over problem instances,
preserving matched conditions. For the
transcript intervention, overall intervals use 5,000 draws and model-specific
intervals use 2,500 draws.

\subsubsection{SARA: Results and Scope}
\label{app:piu-interaction-sara-results}

\begin{table}[H]
\centering
\small
\setlength{\tabcolsep}{3.5pt}
\caption{\textbf{SARA episode success with exact PIUs and added source material.}
Rates pool three models and three repetitions over 44 problem instances.
Both evidence conditions supply the same exact assigned PIUs.
The PIUs-only rows are Multi-Agent PIU-Only; the other rows additionally
supply the original private source material.}
\label{tab:piu-interaction-sara-factorial}
\begin{tabular}{@{}llrrrrr@{}}
\toprule
& & \makecell{Coordinator-\\selected\\speaker} & \BroadcastQueryCell{} &
\RoundRobinCell{} & \CurrentSpeakerSelectionCell{} & \SimultaneousResponsesCell{} \\
\midrule
\makecell[l]{PIUs\\only} & \makecell[l]{One PIU per\\participant} & 69.7\% & 59.8\% & 73.5\% & 75.3\% & 60.9\% \\
& \makecell[l]{All-but-one PIU\\concentration} & 67.9\% & 60.9\% & 78.0\% & 78.5\% & 69.7\% \\
\midrule
\makecell[l]{PIUs and\\source material} & \makecell[l]{One PIU per\\participant} & 35.9\% & 41.2\% & 39.6\% & 46.0\% & 28.5\% \\
& \makecell[l]{All-but-one PIU\\concentration} & 35.6\% & 28.0\% & 27.8\% & 37.1\% & 23.2\% \\
\bottomrule
\end{tabular}
\end{table}

\begin{table}[H]
\centering
\small
\setlength{\tabcolsep}{4pt}
\caption{\textbf{SARA allocation effects and source-material interactions.}
Allocation effects are \AllButOneRIUConcentration{} minus
\OneRIUPerParticipant{}. The interaction is the added-source allocation
effect minus the Multi-Agent PIU-Only allocation effect.
The table heading ``PIUs only'' denotes Multi-Agent PIU-Only. Entries are percentage points
with 95\% problem-instance bootstrap confidence intervals.}
\label{tab:piu-interaction-sara-context}
\begin{tabular}{@{}lrrr@{}}
\toprule
Communication structure & PIUs only & PIUs and source material & Interaction \\
\midrule
\CoordinatorSelectedSpeaker{} & $-1.8\;[-8.6,4.8]$ & $-0.3\;[-6.1,5.6]$ & $+1.5\;[-7.3,10.1]$ \\
\BroadcastQuery{} & $+1.0\;[-5.3,7.3]$ & $-13.1\;[-19.7,-6.1]$ & $-14.1\;[-22.7,-5.6]$ \\
\RoundRobin{} & $+4.5\;[-2.3,11.1]$ & $-11.9\;[-17.9,-5.5]$ & $-16.4\;[-25.3,-7.6]$ \\
\CurrentSpeakerSelection{} & $+3.3\;[-3.0,9.8]$ & $-8.8\;[-15.2,-2.3]$ & $-12.1\;[-19.4,-4.8]$ \\
\SimultaneousResponses{} & $+8.8\;[2.8,15.2]$ & $-5.3\;[-12.1,1.5]$ & $-14.1\;[-23.0,-5.1]$ \\
\bottomrule
\end{tabular}
\end{table}

Adding source material makes concentration additionally harmful under four
communication structures; the \CoordinatorSelectedSpeaker{} interval
includes zero. The \SimultaneousResponses{}--\CoordinatorSelectedSpeaker{}
communication-structure interaction is $+10.6$ points in Multi-Agent PIU-Only (95\% CI $[2.3,19.0]$) and
$-5.1$ with added source material ($[-13.6,3.0]$). The change is $-15.7$
points ($[-26.8,-4.3]$): the $-14.1$-point source-material interaction
under \SimultaneousResponses{} minus the $+1.5$-point interaction under
\CoordinatorSelectedSpeaker{}, calculated before rounding. The negative point estimate in the added-source
condition alone does not establish an ordering reversal with statistical
confidence, since its interval includes zero.

The interaction is not accompanied by a corresponding reduction in semantic
completeness of communicated PIUs. Source-material interactions in
completeness are $-0.5$ points for \CoordinatorSelectedSpeaker{}, $+2.2$
for \BroadcastQuery{}, $+1.0$ for \RoundRobin{}, $+3.5$ for
\CurrentSpeakerSelection{}, and $+2.8$ for \SimultaneousResponses{}.
This motivates a test of information reaching the final decision maker in
addition to the participants' communication of assigned PIUs.

\paragraph{Source-material recipients.}
Table~\ref{tab:piu-interaction-sara-recipients} localizes the larger loss to
source material supplied to participants assigned no PIU. Under
\SimultaneousResponses{}, supplying only that group with source material
reduces episode success by 47.2 points (95\% CI $[-55.3,-39.1]$).
Supplying only participants assigned PIUs reduces it by 14.9 points
($[-23.0,-7.1]$). Once participants assigned no PIU receive source
material, adding it for participants assigned PIUs changes episode success
by $+0.8$ points ($[-3.0,4.3]$). Source material for participants assigned no PIU also has the larger negative point estimate under the two coordinator
structures; the size of that difference is not itself tested here.

\begin{table}[H]
\centering
\small
\setlength{\tabcolsep}{5pt}
\caption{\textbf{SARA source-material recipients under
\AllButOneRIUConcentration{}.} All participants assigned PIUs receive their
exact PIUs. Column headings specify which participants additionally receive
private source material. Entries are episode success rates.}
\label{tab:piu-interaction-sara-recipients}
\begin{tabular}{@{}lrrrr@{}}
\toprule
Communication structure & None & \makecell{Assigned PIUs\\only} & \makecell{Assigned no PIU\\only} & Both groups \\
\midrule
\CoordinatorSelectedSpeaker{} & 67.9\% & 53.3\% & 38.1\% & 35.6\% \\
\BroadcastQuery{} & 60.9\% & 47.2\% & 28.0\% & 28.0\% \\
\SimultaneousResponses{} & 69.7\% & 54.8\% & 22.5\% & 23.2\% \\
\bottomrule
\end{tabular}
\end{table}

\paragraph{Transcript intervention.}
Table~\ref{tab:piu-interaction-sara-filter} reports newly generated final
decisions under both message-visibility conditions. Removing messages from
participants assigned no PIU improves episode success by 47.7 points when those
participants receive source material (95\% CI $[38.9,56.3]$), compared with
$-1.0$ points when they do not ($[-5.3,3.3]$). The interaction is $+48.7$
points ($[38.9,58.6]$). After removal, the added-source minus no-source
difference is 0.0 points ($[-3.8,3.8]$). This is consistent with recovery to
the Multi-Agent PIU-Only performance level, but does not establish exact equality.

\begin{table}[H]
\centering
\small
\setlength{\tabcolsep}{5pt}
\caption{\textbf{SARA fixed-message intervention at the final-decision stage.}
Participants assigned PIUs receive no added source material in either row.
Only the source-material access of participants assigned no PIU varies.
The effect is episode success with their messages removed minus success
with the complete transcript. The interaction is this removal effect with
source material minus the same effect without it.
Effects and 95\% problem-instance bootstrap intervals are in percentage points.}
\label{tab:piu-interaction-sara-filter}
\begin{tabular}{@{}lrrrr@{}}
\toprule
\makecell[l]{Source material for participants\\assigned no PIU} & \makecell{Complete\\transcript} & \makecell{Messages from participants\\assigned PIUs only} & Effect & 95\% CI \\
\midrule
Absent & 69.9\% & 68.9\% & $-1.0$ & $[-5.3,3.3]$ \\
Present & 21.2\% & 68.9\% & $+47.7$ & $[38.9,56.3]$ \\
\midrule
\multicolumn{3}{l}{Interaction} & $+48.7$ & $[38.9,58.6]$ \\
\bottomrule
\end{tabular}
\end{table}

The interaction is positive for every model: $+25.0$ points for Qwen3-32B
(95\% CI $[9.1,40.9]$), $+65.2$ for Qwen3-235B-A22B ($[51.5,78.0]$),
and $+56.1$ for GPT-5.5 ($[43.2,67.8]$). Complete-transcript replay changes episode success relative
to archived decisions by only $+0.3$ points without added source material
($[-2.8,3.3]$) and $-1.3$ with it ($[-3.8,1.5]$), supporting stability of
the original difference under final-decision regeneration.

Under \SimultaneousResponses{}, total message volume from participants
assigned no PIU increases from approximately 371 tokens per transcript
without source material to 1,396 with source material only for that group.
The messages from participants assigned PIUs remain approximately 204 tokens
per transcript in both conditions. These are pooled group totals, rather than
per-message lengths, and therefore can exceed the 600-token participant cap.
The transcript intervention retains the messages communicating assigned PIUs
while removing the messages through which source material supplied to the
other participants reaches the final decision maker.

\paragraph{Scope.}
\label{app:piu-interaction-sara-scope}
The transcript intervention identifies a harmful contribution of messages
from participants assigned no PIU at the final-decision stage, under
\AllButOneRIUConcentration{} and \SimultaneousResponses{}. It does not
separate transcript length, distracting content, contradictory reasoning,
and the final model's allocation of attention. It directly tests source
material supplied only to participants assigned no PIU; it does not establish
the same removal effect when all participants receive source material or
under other communication structures. Nor does it imply that contributions
from participants assigned no PIU are generally unhelpful. Because the follow-ups
supply exact PIUs as well as source material, they support a mechanism relevant
to the cross-domain Full-Context pattern without constituting a direct
re-estimation or complete causal explanation of that regression interaction.

\par
\endgroup
\FloatBarrier

\FloatBarrier
\subsection{Legal Token-Budget Ablations: Complete Results and Diagnostics}
\label{app:participant-resource-budget-details}

For space, Section~\ref{sec:legal-participant-resource-budget-results}
summarizes the principal findings and defers the complete token-budget analyses
to this appendix. We preserve the full results presentation below, followed by
additional implementation details, budget-binding diagnostics, and results by model
and source corpus.

\subsubsection{Complete Results Deferred from Section~\ref{sec:legal-participant-resource-budget-results}}

We study private-note, message, and local-transcript token budgets under
Multi-Agent Full-Context with \OneRIUPerParticipant{} and
\CoordinatorSelectedSpeaker{}. For a problem instance with \(N\) participants,
every budget setting permits \(N\) participant messages before the final
decision. We hold the coordinator policy and final-decision token budget fixed.

\paragraph{Private-note and message token budgets.}
Each participant message is generated in two stages. The participant first
forms a private working note from its initial local observation and local
transcript. A second stage receives the note and the same local transcript and
produces the message added to the communication transcript; the note is not
shared. The message token budget makes the
departure from \hyperref[assump:unconstrained-communication]{A3} explicit.
Because participants receive source material containing their PIUs
rather than the PIUs alone, they must identify the information needed for a
correct final decision before producing bounded messages, a requirement related to
\hyperref[assump:direct-type-report]{A2}.
We cross private-note token budgets of 64, 128, and 256 tokens with message token budgets
of 32, 64, 128, and 256 tokens. Table~\ref{tab:legal-private-note-message-budget-grid}
reports GPT-5.5 results over 24 CUAD and 29 Multi-LexSum problem instances;
Section~\ref{app:legal-private-note-message-budget-details} provides the complete
implementation and results by model and source corpus.

\paragraph{Local-transcript token budgets.}
We separately limit the newly available message text in each participant's
local transcript. This limit is motivated by work on limited human attention
and working memory \citep{kahneman1973attention,cowan2001magical}, without
assuming that language models share the same cognitive mechanisms. We compare
first-in, first-out (FIFO) and random contiguous-window reading with GPT-5.5 on
the 29 Multi-LexSum problem instances. Section~\ref{app:legal-local-transcript-budget-details}
provides the complete implementation and reports how often the token budgets
bind.

\noindent
\begin{minipage}[t]{0.515\textwidth}
    \vspace{0pt}
    \captionsetup{justification=raggedright,singlelinecheck=false}
    \captionof{table}{\textbf{Private-note and message token budgets.}
    GPT-5.5 episode success rates (\%) over 53 problem instances; small
    \(\pm\) terms are standard errors.}
    \label{tab:legal-private-note-message-budget-grid}
    \vspace{0.2\baselineskip}
    \centering
    \small
    \setlength{\tabcolsep}{1.4pt}
    \renewcommand{\arraystretch}{1.24}
    \begin{tabular*}{\linewidth}{@{\extracolsep{\fill}}crrrr@{}}
    \toprule
    \multirow{2}{*}{\makecell{Private-note\\tokens}} &
    \multicolumn{4}{c}{Message tokens} \\[-0.35ex]
    & 32 & 64 & 128 & 256 \\
    \midrule
    64 & 45.3\uncert{6.8} & 67.9\uncert{6.4} & 71.7\uncert{6.2} & 69.8\uncert{6.3} \\
    128 & 60.4\uncert{6.7} & 69.8\uncert{6.3} & 81.1\uncert{5.4} & 75.5\uncert{5.9} \\
    256 & 60.4\uncert{6.7} & 67.9\uncert{6.4} & 83.0\uncert{5.2} & 81.1\uncert{5.4} \\
    \bottomrule
    \end{tabular*}
\end{minipage}\hfill
\begin{minipage}[t]{0.465\textwidth}
    \vspace{0pt}
    \captionsetup{justification=raggedright,singlelinecheck=false}
    \captionof{figure}{\textbf{Local-transcript token budgets.} The dashed line marks the unlimited baseline; shaded bands show uncertainty.}
    \label{fig:legal-local-transcript-token-budgets}
    \vspace{-3mm}
    \centering
    \includegraphics[width=\linewidth]{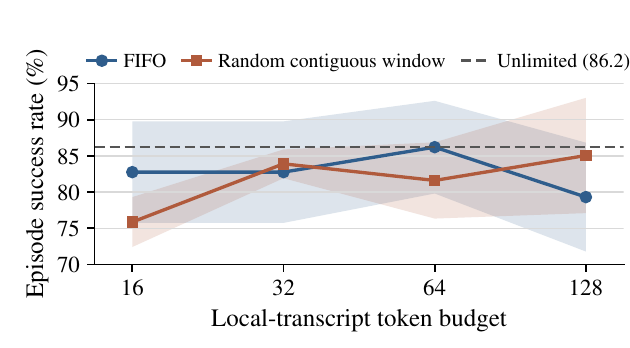}
\end{minipage}

Table~\ref{tab:legal-private-note-message-budget-grid} shows the largest gains
as the message token budget increases from 32 to 128 tokens. Averaged across
private-note token budgets, episode success rate rises from 55.3\% to 78.6\%,
but does not improve further with 256 message tokens; variation across
private-note token budgets is smaller than variation across message token
budgets.

Figure~\ref{fig:legal-local-transcript-token-budgets} shows no monotonic
relationship between the local-transcript token budget and episode success
rate. FIFO remains within 6.9 percentage points of the unlimited baseline,
whereas the random-window mean at 16 tokens is 10.3 points lower. Thus, these
problem instances do not show a uniform performance penalty from a smaller
local-transcript token budget.

\FloatBarrier

\subsubsection{Private-Note and Message Token Budgets}
\label{app:legal-private-note-message-budget-details}

For a problem instance with \(N\) participants, \CoordinatorSelectedSpeaker{}
permits \(N\) participant messages followed by the coordinator's final
decision. Each participant message uses two model generations with the same
local transcript. The first generation receives the participant's initial local
observation and produces a private working note. The second receives that note,
but not the initial local observation, and produces the message added to the
communication transcript. The private working note is not added to the
communication transcript or shown to another decision maker.

The experiment crosses private-note token budgets of 64, 128, and 256 tokens
with message token budgets of 32, 64, 128, and 256 tokens. The token budget for
each participant's initial local observation is 40,000 tokens,
and participant access to the communication transcript is not limited. The
coordinator policy and final-decision token budget remain fixed. The 256-token
message condition is the largest evaluated message token budget within each
private-note row; the grid does not include an unlimited-message condition.

The imposed limits bind as intended. GPT-5.5 participant messages are truncated
in approximately 93\%, 57\%, 23\%, and less than 1\% of episodes as the message
token budget increases, while private-note truncation falls from 82\% to 4\%.
Qwen3-32B does not show a consistent relationship between either token budget
and episode success rate. Table~\ref{tab:legal-private-note-message-budget-by-source-corpus}
reports the source-corpus-specific results for both models.

\begin{table}[!h]
\centering
\scriptsize
\setlength{\tabcolsep}{3pt}
\renewcommand{\arraystretch}{1.06}
\caption{\textbf{Private-note and message token budgets by source corpus and
model.} Episode success rates (\%); small \(\pm\) terms are binomial standard
errors over 24 CUAD or 29 Multi-LexSum problem instances.}
\label{tab:legal-private-note-message-budget-by-source-corpus}
\begin{tabular*}{\textwidth}{@{\extracolsep{\fill}}rrrrrr@{}}
\toprule
\makecell{Private-note\\tokens} & \makecell{Message\\tokens} &
\makecell{GPT-5.5\\CUAD} & \makecell{GPT-5.5\\Multi-LexSum} &
\makecell{Qwen3-32B\\CUAD} & \makecell{Qwen3-32B\\Multi-LexSum} \\
\midrule
64 & 32 & 70.8\uncert{9.3} & 24.1\uncert{7.9} & 79.2\uncert{8.3} & 3.4\uncert{3.4} \\
64 & 64 & 70.8\uncert{9.3} & 65.5\uncert{8.8} & 75.0\uncert{8.8} & 0.0\uncert{0.0} \\
64 & 128 & 70.8\uncert{9.3} & 72.4\uncert{8.3} & 91.7\uncert{5.6} & 3.4\uncert{3.4} \\
64 & 256 & 70.8\uncert{9.3} & 69.0\uncert{8.6} & 83.3\uncert{7.6} & 0.0\uncert{0.0} \\
128 & 32 & 79.2\uncert{8.3} & 44.8\uncert{9.2} & 75.0\uncert{8.8} & 3.4\uncert{3.4} \\
128 & 64 & 79.2\uncert{8.3} & 62.1\uncert{9.0} & 75.0\uncert{8.8} & 6.9\uncert{4.7} \\
128 & 128 & 79.2\uncert{8.3} & 82.8\uncert{7.0} & 83.3\uncert{7.6} & 3.4\uncert{3.4} \\
128 & 256 & 87.5\uncert{6.8} & 65.5\uncert{8.8} & 75.0\uncert{8.8} & 3.4\uncert{3.4} \\
256 & 32 & 79.2\uncert{8.3} & 44.8\uncert{9.2} & 75.0\uncert{8.8} & 0.0\uncert{0.0} \\
256 & 64 & 66.7\uncert{9.6} & 69.0\uncert{8.6} & 79.2\uncert{8.3} & 0.0\uncert{0.0} \\
256 & 128 & 83.3\uncert{7.6} & 82.8\uncert{7.0} & 83.3\uncert{7.6} & 3.4\uncert{3.4} \\
256 & 256 & 79.2\uncert{8.3} & 82.8\uncert{7.0} & 70.8\uncert{9.3} & 3.4\uncert{3.4} \\
\bottomrule
\end{tabular*}
\end{table}

\subsubsection{Local-Transcript Token Budgets}
\label{app:legal-local-transcript-budget-details}

When a participant is selected to send a message, its current directed coordinator
query is read first and charged to the local-transcript token budget. Any
remaining capacity is used to read previously unread message text. FIFO reading
exposes the earliest unread tokens. Random contiguous-window reading instead
selects one contiguous window from the previously unread message text,
reproducibly for a fixed
problem instance, participant, message, and seed. Unread message text remains
available when that participant is next selected. The participant also retains
its initial local observation, its prior messages, and every local-transcript
excerpt it has previously read. The coordinator's observation is not limited and
therefore contains the complete communication transcript.

We fix the private-note and message token budgets at 256 and 128 tokens and
compare local-transcript token budgets of 16, 32, 64, and 128 tokens with
unlimited reading. A 256-token condition provides an additional check at the
largest evaluated token budget. FIFO is evaluated once and random
contiguous-window reading is evaluated with three
seeds. Accordingly,
Table~\ref{tab:legal-local-transcript-token-budget-details}
reports binomial standard errors over the 29 fixed problem instances for FIFO
and standard deviations across the three episode-success rates for random
reading.

\begin{table}[!h]
\centering
\small
\setlength{\tabcolsep}{5pt}
\renewcommand{\arraystretch}{1.05}
\caption{\textbf{Local-transcript token budgets.} Episode success rates (\%)
and changes from the unlimited FIFO baseline. Small \(\pm\) terms are binomial
standard errors for FIFO and standard deviations over three seeds for random
contiguous-window reading.}
\label{tab:legal-local-transcript-token-budget-details}
\begin{tabular}{crrrr}
\toprule
\makecell{Local-transcript\\tokens} & FIFO & \makecell{Change from\\baseline} &
\makecell{Random contiguous\\window} & \makecell{Change from\\baseline} \\
\midrule
16 & 82.8\uncert{7.0} & -3.4 & 75.9\uncert{3.4} & -10.3 \\
32 & 82.8\uncert{7.0} & -3.4 & 83.9\uncert{2.0} & -2.3 \\
64 & 86.2\uncert{6.4} & +0.0 & 81.6\uncert{5.3} & -4.6 \\
128 & 79.3\uncert{7.5} & -6.9 & 85.1\uncert{8.0} & -1.1 \\
256 & 86.2\uncert{6.4} & +0.0 & 82.8\uncert{6.9} & -3.4 \\
Unlimited & 86.2\uncert{6.4} & 0.0 & --- & --- \\
\bottomrule
\end{tabular}
\end{table}

The local-transcript token budget binds at every participant message with
budgets of 16 and 32
tokens and at approximately 95\%, 90\%, and 83\% of participant messages with
budgets of 64, 128, and 256 tokens. At 16 tokens, the coordinator query consumes
the complete local-transcript token budget, so FIFO and random
contiguous-window reading expose the
same prefix; their observed difference reflects run-to-run model variation, not
a different selected excerpt. The 256-token condition reaches 86.2\%
under FIFO and 82.8\uncert{6.9}\% under random reading. Together with the
non-monotonic results in Figure~\ref{fig:legal-local-transcript-token-budgets},
these diagnostics do not support a monotonic relationship between the
local-transcript token budget and episode success rate on these problem
instances.

\subsection{Trajectory Examples}
\label{app:trajectory-examples}

We present two representative episode trajectories that illustrate how PIUs
move from participants' initial local observations into the communication
transcript and final decision\revised{, followed by a matched set of four Finance episodes}.

\subsubsection{Dependency Case and Episode Trajectory: CVS 2022}

\paragraph{Problem and dependency structure.}
This representative FinLongDocQA problem asks for the ratio
of CVS Health's unique digital customers to 2022 revenue in each of three
segments, expressed as millions of customers per \$1 billion of revenue, and
then asks which segment has the largest ratio. Under the \OneRIUPerParticipant{}
condition, Participant A holds the common customer count,
while Participants B--D each hold one segment denominator.

\begin{center}\small
\setlength{\tabcolsep}{5.8pt}
\begin{tabular*}{0.998\textwidth}{@{\extracolsep{\fill}}c
>{\raggedright\arraybackslash}p{0.32\textwidth}
>{\raggedright\arraybackslash}p{0.19\textwidth}
>{\raggedright\arraybackslash}p{0.22\textwidth}@{}}
\toprule
PIU ID & PIU & \makecell{Participant assigned\\the PIU} & \makecell{Answer components\\using the PIU} \\
\midrule
1 & 47M unique digital customers & Participant A & Q1, Q2, Q3, Q4 \\
2 & Health Care Benefits revenue: \$91.4B & Participant B & Q1, Q4 \\
3 & Pharmacy Services revenue: \$169.2B & Participant C & Q2, Q4 \\
4 & Retail revenue: \$106.6B & Participant D & Q3, Q4 \\
\bottomrule
\end{tabular*}
\end{center}

The PIUs used by the four answer components are therefore
\[
Q_1=\{1,2\},\qquad Q_2=\{1,3\},\qquad
Q_3=\{1,4\},\qquad Q_4=\{1,2,3,4\}.
\]
Counting references component by component gives ten PIU uses, but only four distinct
PIUs.  No participant can compute even one requested ratio alone, and the
final comparison requires all four PIUs.

\paragraph{Reference calculation.}
Using the same 47M company-wide customer count as the numerator for each
requested segment calculation gives
\[
\frac{47}{91.4}=0.51,\qquad
\frac{47}{169.2}=0.28,\qquad
\frac{47}{106.6}=0.44.
\]
Thus Health Care Benefits has the highest ratio, at 0.51M unique digital
customers per \$1B of revenue.

\paragraph{Observed success by evaluation condition.}
The table below reports the result for this problem instance under each model
and evaluation condition. A check means that all four answer components were satisfied; a cross
means that all four were missing in the final decision.

\begin{center}\small
\begin{tabular}{lccc}
\toprule
Model & Single-Agent All-PIU & Multi-Agent PIU-Only & Multi-Agent Full-Context \\
\midrule
Claude Sonnet 4.5 & $\checkmark$ & $\checkmark$ & $\checkmark$ \\
Claude Haiku 4.5 & $\checkmark$ & $\times$ & $\times$ \\
\bottomrule
\end{tabular}
\end{center}

\paragraph{Successful episode trajectory: Sonnet 4.5.}
In Single-Agent All-PIU, the model directly read the four facts, computed
0.51, 0.28, and 0.44, and selected Health Care Benefits. In both multi-agent
conditions, all four PIUs entered the communication transcript in the responses
to the first four coordinator queries:
A disclosed 47M; B disclosed \$91.4B; C disclosed \$169.2B; and D disclosed
\$106.6B.  The coordinator used the same company-wide customer count in all
three ratios, performed the divisions, and returned the same ranking.  The semantic
judge found all four answer components correct and grounded in all three conditions.

\paragraph{Failure episode trajectory 1: Haiku 4.5, Multi-Agent PIU-Only.}
The coordinator issued 15 queries to A--D, but every
participant response was empty. No PIU entered the communication transcript.
The final decision explicitly refused to calculate because it lacked the customer
count and segment revenues. The final decision consequently omitted all four
answer components. This is a communication failure rather than an arithmetic
failure: the same model solved the problem in Single-Agent All-PIU.

\paragraph{Failure episode trajectory 2: Haiku 4.5, Multi-Agent Full-Context.}
Participant A found the consolidated 47M customer count, but initially reported
that the needed segment table was unavailable.  Participant C then incorrectly
attached \$169.2B to Health Care Benefits and could not recover the other two
denominators; later participant messages converged on the claim that the
accessible excerpts were insufficient. After 15 coordinator queries, the
coordinator returned ``unable to complete'' instead of the three ratios.  The
final decision omitted all four answer components. Here the source material
created an evidence-retrieval and fact-mapping failure even though every
PIU remained present in the source material provided to its assigned participant.

\paragraph{Case-study implication.}
The evaluation holds the problem instance fixed while exposing two distinct
collaboration bottlenecks. Multi-Agent PIU-Only can fail because a PIU in a participant's initial local observation never
enters the communication transcript; Multi-Agent Full-Context can fail because relevant facts are
lost or misassigned within much longer source material. In Single-Agent All-PIU,
both models solve the problem instance when the PIUs are centralized.

\subsubsection{Composite-Problem Episode Trajectory: Kirkland's 2023}

\paragraph{Generation is not execution.}
This representative FinDVer problem contains six questions, but
Q1--Q5 are not generated during the episode and Q6 is not appended afterward.
All six questions, their links to PIUs, the reference decision, and
the PIU distribution are frozen together as one composite problem instance before
evaluation.  The episode tests whether a coordinator can recover and combine
that already-fixed evidence.

\begin{quote}
\textbf{Data generation:}\quad
10-K source $\rightarrow$ candidate units $\rightarrow$ six PIUs
$\rightarrow$ Q1--Q5 analyses and Q6 synthesis $\rightarrow$ verification of the links between answer components and PIUs
$\rightarrow$ PIU distribution $\rightarrow$ retained problem instance.

\smallskip
\textbf{Evaluation execution:}\quad
initial common observation $\rightarrow$ coordinator queries $\rightarrow$
evidence accumulates in the communication transcript $\rightarrow$ Q1--Q5 and Q6
final decision $\rightarrow$ answer-component correctness and grounding judgment.
\end{quote}

The construction question is therefore ``which facts must jointly determine
the reference decision?''; the evaluation question is ``can the model find those facts and
combine only what the filing supports?''

\paragraph{From filing disclosures to one analyst problem.}
The six-PIU problem joins four review axes that an equity or credit analyst would
normally consider together.

\begin{center}\small
\begin{tabular}{lp{0.45\linewidth}l}
\toprule
Review axis & Filing evidence & Composite item \\
\midrule
Supply chain & Sourcing diversification and country-of-origin mix & Q1 \\
Profitability & 310-bp gross-margin improvement and drivers & Q2 \\
Execution risk & Capital constraints and tightly managed cash flow & Q3 \\
Financing & Term-loan terms, balance-date status, later draws & Q4--Q5 \\
Integrated judgment & Operating improvement plus liquidity pressure & Q6 \\
\bottomrule
\end{tabular}
\end{center}

Q6 asks whether the complete narrative is fully supported: Kirkland's was
improving operations through sourcing diversification and gross-margin
improvement, while still facing liquidity pressure and using new financing as
it executed a financial turnaround strategy.  This is not a vote over Q1--Q5;
it combines evidence with different scopes and dates into one bounded filing
interpretation.

\paragraph{Distributed evidence discovery, queries 1--6.}
Every participant had the generic participant description ``financial filing context
reviewer,'' so the coordinator initially did not know the PIU-ownership mapping.

\begin{center}\small
\begin{tabular}{rllp{0.39\linewidth}}
\toprule
Query & Participant & Item & Evidence recovered \\
\midrule
1 & A & Q4 & No term-loan balance at Feb. 3; \$34M credit borrowing \\
2 & B & Q1 & Diversification and fiscal 2023 country mix \\
3 & C & Q3, Q5 & Tight cash management; later \$6M and \$5M draws \\
4 & D & Q3 & Capital shortage could limit strategic execution \\
5 & E & Q4 & \$12M term loan; at least \$5M initial draw required \\
6 & F & Q2 & 310-bp margin improvement, drivers, and offset \\
\bottomrule
\end{tabular}
\end{center}

Query 1 illustrates genuine discovery rather than replay of a reference-decision ordering.
The coordinator asked A for sourcing information, but A did not own that PIU.
A instead disclosed its financing evidence.  The coordinator used the updated
communication transcript to query another reviewer for the still-missing sourcing component,
and recovered all six PIUs by query 6.

\paragraph{Cross-source consistency checks, queries 7--15.}
Once the decision-relevant facts were present, the remaining questions tested
how strongly they could be connected.  F confirmed that its margin paragraph
did not itself call the improvement part of the financial turnaround.  E
confirmed that its term-loan-terms paragraph did not state a strategic use of
proceeds.  By contrast, C's paragraph jointly disclosed later borrowing,
tightly managed cash flow, and execution of the financial turnaround strategy.
The defensible synthesis is therefore that the company used additional
financing while executing the turnaround under cash pressure--not that a
particular loan directly caused or funded a particular operating improvement.

\paragraph{Successful three-stage episode trajectory: Sonnet 4.5, Multi-Agent PIU-Only.}
\textbf{1. Distributed evidence discovery:} queries 1--6 mapped the six anonymous
reviewers to sourcing, margin, capital-risk, term-loan, balance-date, and
post-year-end evidence. \textbf{2. Checking support for cross-passage conclusions:}
queries 7--15 distinguished direct filing statements from unsupported
cross-passage inferences and
preserved the February 3 balance-date versus subsequent-borrowing distinction.
\textbf{3. Integrated conclusion:} the final decision reported Q1--Q5 and
concluded that Q6 was fully supported. The episode used the complete
budget of 15 participant responses, and all six answer components were
correct and grounded.

\paragraph{Efficiency and interpretation.}
The episode trajectory is deliberately informative rather than optimal.  An initial
misdirected query was necessary because PIU ownership was hidden, whereas
several queries after query 6 repeated checks of whether the filing supported
the proposed cross-passage conclusions after the core facts were already
complete. For presentation, the 15 coordinator queries and responses are most
clearly compressed into the three stages above.  The example then shows more
than parallel subproblem solving: it shows distributed evidence recovery,
verification of permissible cross-source links, and construction of one
realistic operational-and-liquidity judgment.

\revised{%
\subsubsection{Matched Episodes Across PIU Distributions and Communication Structures: Ball 2023}
\label{app:matched-episodes-ball}

\paragraph{Problem and PIUs.}
This FinLongDocQA problem asks for two calculations from Ball Corporation's
2023 annual report: (1) the number of employees who left in 2023, estimated
from the year-end headcount and the employee turnover rate, and (2) the
number of aluminum packaging units shipped in 2023 that correspond to 1\% of
customer emissions. The reference decision is 3,234 employees
($0.154 \times 21{,}000$) and 2.6625 billion units ($106.5/40$). It rests on
three PIUs, which the two PIU distributions assign as follows.

\begin{center}\small
\setlength{\tabcolsep}{5pt}
\begin{tabular*}{0.998\textwidth}{@{\extracolsep{\fill}}c
>{\raggedright\arraybackslash}p{0.40\textwidth}
>{\raggedright\arraybackslash}p{0.19\textwidth}
>{\raggedright\arraybackslash}p{0.21\textwidth}@{}}
\toprule
PIU ID & PIU & \makecell[l]{One PIU per\\participant} & \makecell[l]{All-but-one PIU\\concentration} \\
\midrule
1 & Year-end headcount of 21,000 and employee turnover rate of 15.4\% & Participant A & Participant A \\
2 & Packaging accounts for up to 40\% of customers' emissions & Participant B & Participant A \\
3 & 106.5 billion units of aluminum packaging shipped in 2023 & Participant C & Participant B \\
\bottomrule
\end{tabular*}
\end{center}

\paragraph{Observed outcomes.}
All four episodes use Qwen3-235B-A22B in Multi-Agent Full-Context. They cross
the two PIU distributions and the two communication structures compared in
Section~\ref{sec:piu-interaction-finance-example}. A check means that both
answer components were judged correct.

\begin{center}\small
\begin{tabular}{lcc}
\toprule
PIU distribution & \CoordinatorSelectedSpeaker{} & \RoundRobin{} \\
\midrule
\OneRIUPerParticipant{} (three participants) & $\checkmark$ & $\times$ \\
\AllButOneRIUConcentration{} (two participants) & $\times$ & $\times$ \\
\bottomrule
\end{tabular}
\end{center}

\paragraph{\OneRIUPerParticipant{}, \CoordinatorSelectedSpeaker{}: success.}
The coordinator asked A for the year-end headcount and turnover rate and
received 21,000 and 15.4\%. It then asked B for the shipment volume and the
share of customer emissions. B reported the ``up to 40\%'' statement and said
that its material contained no shipment volume. The coordinator directed the
remaining question to C, who reported 106.5 billion units. The final decision
combined the three PIUs and gave both reference values.

\paragraph{\OneRIUPerParticipant{}, \RoundRobin{}: failure.}
Each participant responded to the whole problem in turn. A reported the
headcount and turnover rate but applied the rate to the 2022 headcount of
23,000. B stated that its material contained neither figure. C cited average
employee counts of 314,775 and 316,433 from its own material and computed
item (2) correctly from the 40\% share and 106.5 billion units. The final
decision gave 2.6625 billion units for item (2) but, citing the inconsistent
employee figures, declined to estimate the number of employees who left.

\paragraph{\AllButOneRIUConcentration{}, \CoordinatorSelectedSpeaker{}: failure.}
A now holds PIUs 1 and 2. The coordinator asked A only for the headcount and
turnover rate, and A reported only those. The coordinator then asked B for
both the shipment volume and the share of customer emissions. B holds only
PIU 3; it reported 106.5 billion units and an emissions share of
``approximately 20\%.'' PIU 2 never entered the communication transcript. The
final decision gave 3,234 employees for item (1) but computed
$106.5/20 = 5.325$ billion units for item (2).

\paragraph{\AllButOneRIUConcentration{}, \RoundRobin{}: failure.}
Without a coordinator query, A reported all of its evidence: the headcount,
the turnover rate, and the 40\% share. B reported 106.5 billion units and
computed item (2) correctly, but stated that its material did not contain the
headcount or the turnover rate. The final decision gave 2.6625 billion units
for item (2) and, citing B's statement, declined to estimate the number of
employees who left.

\paragraph{What the case illustrates.}
Under \CoordinatorSelectedSpeaker{}, moving from \OneRIUPerParticipant{} to
\AllButOneRIUConcentration{} turns a success into a failure, whereas
\RoundRobin{} fails under both distributions. Under concentration, the
coordinator's query to A requested only part of A's evidence, so the PIU that
A newly held was not reported; under \RoundRobin{}, the same participant
reported it without being asked. This is the query-content mechanism tested
in Section~\ref{sec:piu-interaction-finance-example}. The case also shows
what a coordinator can add when the PIUs are spread across participants: a
follow-up query recovered the shipment volume from a third participant, while
the \RoundRobin{} final decision maker received inconsistent employee figures
and did not resolve them.

\paragraph{Scope.}
These four episodes were selected after observing their outcomes. The two PIU
distributions also differ in team size (three versus two participants) and
therefore in the number of participant responses. The case illustrates how
evidence moved in individual episodes and does not isolate the effect of the
PIU distribution. The controlled interventions in
Appendix~\ref{app:riu-distribution-communication-structure-results-interpretation}
provide the corresponding estimates.
}

\FloatBarrier
\revised{%
\subsection{Scope of the Communication and Retrieval Gaps}
\label{app:gap-scope}
Both gaps in Section~\ref{sec:cross-domain-three-evaluation-conditions-results}
are differences in episode success rate between two evaluation conditions.
The two conditions share the problem instance, its PIUs, the reference
decision, the model, and the evaluator, so a gap does not reflect a difference
in problem difficulty or model knowledge. Each gap is an aggregate contrast.
It measures the combined effect of everything that changes between its two
conditions and does not assign separate contributions to the individual
changes.

\begin{center}
\small
\setlength{\tabcolsep}{4pt}
\renewcommand{\arraystretch}{1.15}
\begin{tabularx}{\textwidth}{@{}>{\raggedright\arraybackslash}p{0.19\textwidth}>{\raggedright\arraybackslash}p{0.27\textwidth}X@{}}
\toprule
Gap or condition & Conditions compared & What changes \\
\midrule
Communication gap &
Single-Agent All-PIU $-$ Multi-Agent PIU-Only &
The PIUs are distributed across participants, who must exchange them under the
communication structure and resource budgets before a decision maker makes the
final decision. \\
Retrieval gap &
Multi-Agent PIU-Only $-$ Multi-Agent Full-Context &
Each participant receives source material containing its assigned PIUs instead
of the PIUs themselves. The PIU distribution, communication structure, and
resource budgets are unchanged. \\
Single-Agent Full-Context (supplementary condition) &
Reported separately; not part of either gap &
One model receives all participants' source material instead of the PIUs.
There is no team communication. \\
\bottomrule
\end{tabularx}
\end{center}

\textit{Communication gap.} Moving from one model to a team changes who
initially holds each PIU, replaces a single model call with the interaction
specified by the communication structure and its resource budgets, and gives
the final decision to a decision maker that depends on what the team
communicates. The gap therefore
includes coordination (who is asked and what is asked), transmission of PIUs
in messages, synthesis of the communicated evidence, and the final decision.
This contrast corresponds to a departure from the assumption of nonbinding
communication resources (\hyperref[assump:unconstrained-communication]{A3},
Section~\ref{sec:related-work}): once the PIUs are distributed, the final
decision depends on what participants transmit and read under explicit
resource budgets, whereas a single model that holds all PIUs faces no such
constraint.

\textit{Retrieval gap.} Under the direct-report idealization
(\hyperref[assump:direct-type-report]{A2}, Section~\ref{sec:related-work}),
each participant knows its private information and can report it. Multi-Agent
PIU-Only corresponds to this idealization: each participant is given the
pivotal facts it holds and has only to communicate them. Multi-Agent
Full-Context removes it. A participant must first identify, within its own
source material, the evidence that matters for a decision made jointly with
participants whose material it cannot see, and then express that evidence in
a bounded message. The retrieval gap measures what this added step costs the
team's decision, which is why it compares the two multi-agent conditions.
When one model both reads the source material and makes the decision, there
is no report to another decision maker.

In \ours{}, all required evidence is already contained
in the participants' initial observations, and no external search is involved.
We use \emph{retrieval} to mean identifying the relevant evidence within the
supplied source material and using it. Relative to Multi-Agent PIU-Only,
Multi-Agent Full-Context gives participants longer inputs that contain
material irrelevant to the decision, and the amount of source material can
differ across participants. The retrieval gap therefore includes identifying
the evidence, handling irrelevant content, processing long inputs, and
integrating the identified evidence. It does not separate the loss from
identifying evidence from the loss due to long-context processing.

\textit{Single-Agent Full-Context (supplementary condition).} Single-Agent
Full-Context
(Appendix~\ref{app:cross-domain-model-source-corpus-results}) gives one model
the source material of all participants combined. It shows how one model uses
the pooled source material when no communication takes place.
Appendix~\ref{app:cross-domain-single-full-context} reports the evaluation
coverage of this condition. The condition is not part of either gap. Its
difference from Single-Agent All-PIU
cannot be subtracted from the retrieval gap to isolate the part that is due
to long inputs, because the two settings differ. The single model reads all
the source material at once, whereas each participant reads only its own
share, and evidence that a participant fails to identify is also missing from
the communication that follows.

The transcript interventions in Section~\ref{sec:selected-mechanism-results}
and the trajectory replay in Section~\ref{sec:current-speaker-participant-policy-training}
examine specific components under controlled conditions.
In particular, Section~\ref{sec:piu-interaction-finance-example} varies what
the coordinator asks participants to report, and
Section~\ref{sec:piu-interaction-sara-example} varies participants' access to
source material and the messages that reach the final decision maker.
}

\section{Literature Motivation for the PIU-Distribution--Communication-Structure Interaction}
\label{app:riu-distribution-communication-theory}

{Section~\ref{sec:cross-domain-riu-distribution-communication-structure-results}
tests whether differences in episode success rate between communication
structures change across PIU distributions. Three complementary literatures motivate
this comparison and the interaction specification.

\paragraph{Team decision-making under a decentralized information structure.}
Team decision theory studies decision makers who pursue a shared objective but
have different information available to them
\citep{radner1962team,marschak1972economic,nayyar2013common}. This is the
basic structure of our evaluation conditions with multiple participants: the
participants share the episode objective, while PIU distribution determines
which PIUs are available to which participants. Consistent with this
formulation, we hold the problem instance and model fixed while varying PIU
distribution. The literature does not imply that a particular PIU distribution
must perform best, because performance also depends on whether the distributed
PIUs enter communication and support a correct final decision.

\paragraph{Communication under a decentralized information structure.}
Communication complexity and distributed-function computation study how
separately informed parties exchange the information needed to compute a
shared target \citep{yao1979complexity,orlitsky2001coding}. Work on knowledge
in distributed systems likewise formalizes communication as changing the
information available to decision makers \citep{halpern1990knowledge}. In our
setting, communication structure determines who may send a message
at each point in the episode and which earlier messages that participant can
observe, while PIU distribution determines where the relevant information is
initially available. We study PIU distribution and communication structure
jointly because differences in episode success rate among communication
structures may vary across PIU distributions.

\paragraph{PIUs entering communication.}
Hidden-profile experiments show that information observed by only one
participant may fail to enter group discussion \citep{stasser1985pooling}.
This evidence is directly relevant to PIU distributions that place a PIU with
only one participant or concentrate multiple PIUs with one participant. It
also explains why the proportion of participants who send a message is not by
itself sufficient: episode success depends on whether the communication
structure elicits and preserves the PIUs needed for a correct final decision.

\paragraph{Implications for the empirical specification.}
The specification includes PIU-distribution and communication-structure main
effects and their interaction, allowing differences among communication
structures to vary across PIU distributions. Because both are represented as categorical
conditions and every non-reference pair receives an interaction term, the
specification does not impose that their relationships with episode success
are independent. The Bernoulli model and logit link reflect that episode
success is binary, while problem-family and model fixed effects adjust for
differences across problem families and models. Standard errors are clustered
by problem instance, and
Appendix~\ref{app:problem-instance-riu-distribution-communication-structure-results}
reports the stricter specification with problem-instance fixed effects. The
analysis estimates empirical relationships among the evaluated conditions;
these literatures motivate the comparison but do not make the interaction
estimates causal tests of any one theory.
}

\section{Policy-Training Data, Evaluation, and Diagnostic Analyses}
\label{app:policy-training}

This appendix documents the data, evaluation, training configurations, and
diagnostic analyses for the two policy-training demonstrations in
Section~\ref{sec:policy-training}. We first describe the CUAD-derived and
MAUD-derived free-form decisions and their evaluator. We then report the
demonstration under \CoordinatorSelectedSpeaker{} and the demonstration under
\CurrentSpeakerSelection{}, with the participant that ends communication
making the final decision.

\paragraph{Relation to assumptions A1, A2, and A3.}
Both demonstrations address
\hyperref[assump:strategic-computation]{A1} by training and comparing
language-model policy collections rather than assuming that a jointly optimal
policy collection is available before execution. The
\CoordinatorSelectedSpeaker{} demonstration additionally departs from
\hyperref[assump:direct-type-report]{A2} because participants derive messages
from source material, whereas the \CurrentSpeakerSelection{} demonstration
provides PIUs directly. Both demonstrations depart from
\hyperref[assump:unconstrained-communication]{A3} by enforcing explicit
observation and communication budgets.

\subsection{Free-Form Legal Decisions and Evaluation}
\label{app:policy-training-free-form-decisions}

\subsubsection{Construction and Data Splits}
\label{app:policy-training-free-form-construction}

We construct CUAD-derived and MAUD-derived problem instances from the
corresponding distributed CUAD and MAUD problem instances while preserving the
PIU distribution and source material. For each input problem instance, GPT-5.5
receives the original decision task, the reasoning supporting its reference
decision, the PIUs and their source text, and the PIU distribution. It produces
a standalone legal question, a source-grounded reference decision,
PIU-grounded answer components, a component requiring synthesis across all
PIUs, material contradictions that disqualify a final decision, and
problem-specific evaluation criteria. The question identifies the dimensions
to resolve without exposing the original decision alternatives or their
labels.

A separate GPT-5.5 call first checks the proposed question, reference decision, and
evaluation criteria against the source evidence. The check rejects disguised
multiple choice, reference-decision leakage, questions answerable from one PIU,
unsupported content, incomplete PIU coverage, and criteria that would accept a
partial or vague final decision. Deterministic checks then enforce the length
and format requirements, require at least one answer component grounded in
every PIU and one component synthesizing all PIUs, and verify that the PIU
distribution is unchanged. If either stage fails, the proposal is revised
using the resulting findings. Of 396 retained problem instances, 283 pass on
the first draft; the remaining instances require up to seven revision rounds.

\begin{table}[!h]
\centering
\small
\setlength{\tabcolsep}{7pt}
\caption{\textbf{Free-form legal problem instances.} Only CUAD-derived
training problem instances supply gradients. CUAD-derived validation contracts
are absent from training, and MAUD-derived problem instances are opened only
after CUAD-derived checkpoint selection.}
\label{tab:policy-training-data-counts}
\begin{tabular}{lrrr}
\toprule
Use & CUAD-derived & MAUD-derived & Total \\
\midrule
Training & 334 & 0 & 334 \\
Validation & 36 & 0 & 36 \\
MAUD-derived evaluation & 0 & 26 & 26 \\
\bottomrule
\end{tabular}
\end{table}

Across the 396 problem instances, the PIU count has median 2, mean 2.13, and
range 2--6. Questions contain 47 words at the median, and reference decisions
contain 100 words at the median. On the 62 CUAD-derived validation and
MAUD-derived evaluation instances, the evaluator receives a median of six
answer components: five grounded in individual PIUs and one requiring
cross-PIU synthesis. All 62 reference decisions receive an LLM-judged score of
1.0. By contrast, 62 single-PIU partial decisions and 62 original incorrect
decision alternatives have a mean score of 0.193. Policies never observe the reference decisions, answer-component
annotations, PIU identifiers, construction checks, or evaluation criteria.

\subsubsection{GPT-5.5 Evaluator}
\label{app:policy-training-free-form-evaluation}

Free-form decisions admit multiple valid surface forms, so the evaluator
compares a candidate final decision with the reference decision, PIUs, answer
components, problem-specific evaluation criteria, and material contradictions.
GPT-5.5 directly assigns an \emph{LLM-judged score} in \([0,1]\). The
implementation clips the returned value to this interval and assigns zero when
the score is unavailable. The arithmetic mean across problem instances is the
reported LLM-judged score and the only semantic training reward. The evaluation
criteria require the final decision to reach the reference conclusion, cover
the required content, satisfy the applicable numeric and legal conditions, and
contain no disqualifying error. A malformed or missing judge response receives
a score of zero. All evaluations use the AWS Bedrock-hosted GPT-5.5 at
temperature zero with the same prompt and normalization procedure.

\subsection{\CoordinatorSelectedSpeaker{}: Participant and Coordinator Policy Training}
\label{app:coordinator-controlled-policy-training}

All episodes in this subsection use Multi-Agent Full-Context with
\CoordinatorSelectedSpeaker{}. A shared participant policy generates bounded
messages from each selected participant's initial local observation, while a
coordinator policy generates coordinator queries and the final decision. Each initial local observation has a 28,000-token
budget; participant messages have a 48-token budget within a 64-token
structured completion; coordinator queries have a 64-token budget; and the
final decision has a 256-token budget within a 384-token structured completion.
Participants do not generate private notes.

\subsubsection{Optimization and Training Configurations}

Each update samples eight shuffled CUAD-derived training problem instances and
eight rollouts per problem instance, yielding 64 episode rollouts. GRPO
standardizes the eight rewards within each problem instance and broadcasts the
resulting episode advantage to the trainable actions. An invalid trainable
action instead receives a local advantage of \(-0.1\) and one retry. We update
all parameters of each trainable policy with AdamW for one epoch per update
using a PPO-style clipped
surrogate, ratio clip 0.2, dual-clip bound 3.0 for negative advantages,
sequence-mean/token-mean loss aggregation, learning rate \(10^{-6}\), weight
decay 0.01, and a \(10^{-3}\) KL loss relative to the base
Qwen2.5-7B-Instruct policy. Training runs for 80 updates, validation occurs
every 10 updates with greedy decoding, and each configuration uses four H200
GPUs.

\begin{table}[!h]
\centering
\small
\setlength{\tabcolsep}{3pt}
\renewcommand{\arraystretch}{1.12}
\caption{\textbf{Training configurations under \CoordinatorSelectedSpeaker{}.} The
participant-only stage and coordinator-only continuation form one two-stage
policy-training experiment. Joint two-policy training begins independently
from the base policy for both trainable policies.}
\label{tab:coordinator-controlled-training-configurations}
\begin{tabular}{@{}>{\centering\arraybackslash}m{0.16\textwidth}
>{\centering\arraybackslash}m{0.258\textwidth}
>{\centering\arraybackslash}m{0.258\textwidth}
>{\centering\arraybackslash}m{0.258\textwidth}@{}}
\toprule
Setting & Participant-only stage & Coordinator-only continuation & Joint two-policy training \\
\midrule
Updated policy
& Participant policy
& Coordinator policy
& Participant and coordinator policies \\
\arrayrulecolor{black!18}\specialrule{0.3pt}{0.8pt}{0.8pt}\arrayrulecolor{black}
Fixed policy
& Base coordinator policy
& Trained participant policy
& Neither \\
\bottomrule
\end{tabular}
\end{table}

The LLM-judged score is the training reward in all configurations. Which
participant the coordinator queries, query order, which PIUs appear in
participant messages, and whether the coordinator queries every participant do
not enter the reward.

\subsubsection{Two-Stage Policy Training}
\label{app:two-stage-policy-training}

The participant-only stage updates the participant policy while the base
coordinator policy remains fixed. We evaluate the participant policy every 10
updates; the version after update 60 is the first to attain the highest mean
LLM-judged score on CUAD-derived validation and is therefore used in the
coordinator-only continuation. This continuation initializes the coordinator
policy from the base Qwen2.5-7B-Instruct policy and holds the trained
participant policy fixed. The coordinator policy selected after this stage is
the update-80 checkpoint.

\begin{table}[!h]
\centering
\small
\setlength{\tabcolsep}{8pt}
\caption{\textbf{CUAD-derived validation results during two-stage policy
training.} Values are percentages.}
\label{tab:two-stage-policy-training-results}
\begin{tabular}{lc}
\toprule
Training stage & LLM-judged score \\
\midrule
Base policies & 12.14 \\
Participant-only stage & 16.84 \\
Coordinator-only continuation & \textbf{23.33} \\
\bottomrule
\end{tabular}
\end{table}

The participant-policy comparison uses episodes generated by the base and
trained participant policies under the same fixed coordinator policy and
decoding configuration. Table~\ref{tab:participant-only-training-examples} shows
that training causes participant messages to include information needed for a
correct final decision and changes which PIUs appear across participant
messages. In the insurance-obligations example, the trained
participant policy produces a different first message. Although the coordinator
policy remains fixed, it consequently observes a different transcript and
queries participant B instead of repeating participant A.

\begin{table*}[!h]
\centering
\scriptsize
\setlength{\tabcolsep}{5pt}
\renewcommand{\arraystretch}{1.12}
\caption{\textbf{Illustrative changes after participant-only training.} Only
the participant policy changes. Values are LLM-judged scores; the examples are
not frequency estimates and do not enter training or checkpoint selection.}
\label{tab:participant-only-training-examples}
\begin{tabular}{@{}
>{\centering\arraybackslash}m{0.15\textwidth}
>{\raggedright\arraybackslash}m{0.31\textwidth}
>{\raggedright\arraybackslash}m{0.31\textwidth}
>{\raggedright\arraybackslash}m{0.13\textwidth}@{}}
\toprule
Case & Base participant policy & Trained participant policy & Observed change \\
\midrule
Liability limits
& Both participant messages deny that the agreement contains liability limits;
the final decision receives 0.00.
& One participant message includes the mutual exclusion of lost profits and
indirect, special, consequential, and punitive damages, including the product
context and even-if-advised condition; the final decision receives 0.33.
& Participant message includes information needed for a correct final decision \\
\addlinespace
Compliance audit and access
& Participant messages leave the right holder unspecified and substitute
unrelated consumer-credit services; the final decision receives 0.00.
& The two participant messages separately include Equidata's at-any-time
site-inspection right and its compliance audit at Equidata's expense; the final
decision receives 0.33 but still merges the two rights.
& Different PIUs appear across participant messages \\
\addlinespace
Insurance obligations
& The first participant message includes the \$1M coverage, but the second
coordinator query repeats participant A and obtains an unrelated fiber-rights
clause; the final decision receives 0.40.
& After the trained participant policy produces a different
first message, the fixed coordinator observes the changed transcript and
queries participant B, which supplies the cancellation, non-renewal,
modification, and 30-day notice condition; the final decision receives 0.80.
& Changed participant message leads to a different next coordinator query \\
\bottomrule
\end{tabular}
\end{table*}

The coordinator-policy comparison holds the trained participant policy fixed.
Table~\ref{tab:coordinator-only-training-examples} shows one example in which
changed coordinator queries cause participant messages to include both the
certificate-on-request clause and a reciprocal insurance obligation. In
another, the trained coordinator policy produces the correct final decision
from participant messages that are substantively unchanged. These examples
separate changes arising from participant messages from changes arising through
coordinator queries or the final decision.

\begin{table*}[!h]
\centering
\scriptsize
\setlength{\tabcolsep}{5pt}
\renewcommand{\arraystretch}{1.12}
\caption{\textbf{Illustrative changes after the coordinator-only continuation.} Only
the coordinator policy changes; the trained participant policy remains fixed.
Values are LLM-judged scores; the examples are not frequency estimates and do
not enter training or checkpoint selection.}
\label{tab:coordinator-only-training-examples}
\begin{tabular}{@{}
>{\centering\arraybackslash}m{0.15\textwidth}
>{\raggedright\arraybackslash}m{0.32\textwidth}
>{\raggedright\arraybackslash}m{0.32\textwidth}
>{\raggedright\arraybackslash}m{0.10\textwidth}@{}}
\toprule
Case & Base coordinator policy & Trained coordinator policy & Observed change \\
\midrule
Distributor insurance
& Participant messages include unrelated regulatory and committee-cost facts;
the final decision receives 0.00.
& Changed coordinator queries cause participant messages to include both the
certificate-on-request clause and the reciprocal insurance obligation; the
final decision receives 1.00.
& Changed coordinator queries \\
\addlinespace
Co-brand minimum commitment
& Both participant messages contain the 5,000-unit initial order and the
10,000-unit/32-month guarantee, but the final decision confuses their roles;
the final decision receives 0.33.
& With substantively unchanged participant messages, the final decision
correctly separates the initial order from the longer-term guarantee and
receives 1.00.
& Changed final decision from the same participant messages \\
\bottomrule
\end{tabular}
\end{table*}

On the 26 MAUD-derived problem instances held out from training and checkpoint
selection, two-stage policy training does not improve the aggregate LLM-judged
score.

\subsubsection{Joint Two-Policy Training}
\label{app:joint-two-policy-training}

Joint two-policy training removes weight sharing between the participant and
coordinator policies. Both policies are initialized independently from
Qwen2.5-7B-Instruct, maintain separate optimizer states and clipped-surrogate
losses, and receive the same team payoff. Every update has a nonzero gradient
norm for each policy, ruling out the failure mode in which one shared policy
generates or updates both participant and coordinator actions.

\revised{Figure~\ref{fig:policy-training-validation-cuad} summarizes three
joint-training histories through update~80. The shaded band shows their
across-run SD.}
The representative run uses random seed 29, 80 updates, and 5,120 episode rollouts.
The update-80 pair attains the highest mean LLM-judged score among the
scheduled CUAD-derived validation evaluations. The score increases from
11.97\% to 22.38\%. The coordinator queries every participant on 100\% of validation
problem instances by update 20 and maintains that rate through update 80.

The coordinator-query diagnostics use one greedy rollout per
CUAD-derived validation problem instance. A first query copies the problem
question only when the query and question match after normalization for case,
punctuation, and whitespace. A later query is lexically reply-conditioned when
it contains an explicit follow-up marker or at least two terms from a preceding
participant message that are absent from the problem question. This diagnostic
does not constitute a causal test of adaptation. The trained coordinator
queries every participant, uses short queries, and does not copy the problem
question in the first query; lexically reply-conditioned later queries remain
infrequent. Thus, the learned coordinator-query behavior is distinguished by
querying every participant with short queries, rather than by a high rate of
lexically reply-conditioned later queries.

In one CUAD-derived validation example, the base participant and coordinator
policies elicit only the parties and the Type I material label and produce a
one-directional final decision with an LLM-judged score of 12.5\%. The jointly
trained policies instead elicit the IBM-to-Customer Type II permissions from
one participant and the Customer-to-IBM downstream-authorization right for
Type I materials from another. Their final decision represents both grant
directions and receives 62.5\%, although it remains incomplete. This example
illustrates coordinated evidence decomposition but does not establish that the
interaction between the two trained policies is necessary.

Under the jointly trained participant and coordinator policies, the coordinator
queries every participant and every participant sends a public message on all
26 MAUD-derived problem instances; both behaviors occur on 20 of 26 instances
under the base participant and coordinator policies. The LLM-judged score
changes only slightly, from 5.12\% to 5.58\%. The query and messaging patterns
therefore transfer more clearly than task-specific evidence use or
final-decision quality.

\revised{%
\FloatBarrier
\subsubsection{\revised{CUAD-to-MAUD Transfer: Paired Information-Flow Analysis}}
\label{app:cuad-maud-transfer-analysis}

The preceding results distinguish transfer of query and messaging behavior
from transfer of decision quality. We further examine this distinction using
the exact base and trained trajectories from the reported joint-training run
(seed 29, CUAD-selected update 80). The paired cohort contains 36 CUAD
validation problem instances from 22 source agreements and 26 held-out MAUD
problem instances from six agreements, with one greedy episode per policy and
instance. Training and checkpoint selection use only CUAD. The paired records
have matching normalized problem-instance inputs. We use the original
reported CUAD trajectories and locked MAUD evaluations; separate CUAD reruns,
older staged-training audits, and finite-choice evaluations are excluded.
No additional policy training or controlled policy replay is performed here.

\paragraph{Paired audit and uncertainty.}
A separate diagnostic call to GPT-5.5 evaluates the base and trained
trajectories together for each instance. The auditor receives the reference
PIUs and their ownership, required evidence and synthesis components, public
queries and replies, and final decisions. It traces whether relevant evidence
is requested, faithfully communicated, and used in the final decision.
Diagnostic scores are zero for absent or incorrect content, one-half for
partially correct content that omits material detail, and one for complete,
correct treatment including relevant qualifiers. Scores are averaged within
each task and then across all tasks; unqueried PIUs receive zero coverage
credit. These fractional diagnostic scores are separate from the original
reward judgments and are not proportions of fully correct tasks. A
two-instance schema and consistency check precedes the complete audit.

Paired 95\% percentile intervals use 50,000 bootstrap samples of source
agreements, retaining all instances and both policy endpoints within each
sampled agreement. They condition on this training seed, selected checkpoint,
rollout, and fixed judge outputs. The different PIU and answer-component
granularities across corpora make within-corpus changes the primary
comparison.

\begin{table}[H]
\centering
\small
\setlength{\tabcolsep}{6pt}
\renewcommand{\arraystretch}{1.12}
\captionsetup{skip=4pt}
\caption{\revised{\textbf{Participation transfers more clearly than evidence transmission
on MAUD.} Values compare the base and trained policies on the same 36 CUAD
and 26 MAUD instances. Original outcomes and new fractional audit scores are
shown separately; all audit values are percentages of their scoring scale.}}
\label{tab:cuad-maud-transfer-pipeline}
\revised{%
\begin{tabularx}{\linewidth}{@{}Xcc@{}}
\toprule
Measurement & \makecell{CUAD\\Base $\to$ trained} &
\makecell{MAUD\\Base $\to$ trained} \\
\midrule
\multicolumn{3}{l}{\textit{Original outcomes}} \\
LLM-judged score (\%) & $11.97\to22.38$ & $5.12\to5.58$ \\
Every participant queried and replies & $29/36\to36/36$ & $20/26\to26/26$ \\
A final written answer exists & $29/36\to36/36$ & $21/26\to26/26$ \\
\midrule
\multicolumn{3}{l}{\textit{Paired information-flow audit: mean partial-credit scores (\%)}} \\
Relevant PIU request coverage & $31.71\to37.27$ & $31.03\to39.62$ \\
Faithful PIU transmission & $25.93\to42.13$ & $10.26\to13.01$ \\
Required evidence communicated & $24.77\to37.37$ & $12.31\to12.65$ \\
Required evidence in the final decision & $17.99\to31.31$ & $15.15\to17.03$ \\
Final evidence supported by messages & $19.90\to35.11$ & $16.31\to14.78$ \\
Cross-PIU integration & $13.89\to25.00$ & $5.77\to7.69$ \\
\bottomrule
\end{tabularx}}
\end{table}

\paragraph{Where evidence is lost.}
Reaching every evidence owner does not ensure that the relevant legal content
enters the conversation. Required-evidence communication increases by 12.60
percentage points on CUAD (95\% interval $[3.39,22.21]$), but by only 0.35
points on MAUD ($[-2.19,4.15]$). Faithful PIU transmission increases by 16.20
points on CUAD ($[6.79,26.16]$), compared with an uncertain 2.76 points on
MAUD ($[-1.80,8.21]$). The MAUD cross-PIU integration change is also uncertain:
1.92 points ($[-5.88,15.62]$).

The underlying counts make the distinction concrete. Across the MAUD cohort,
the number of PIUs whose owners are queried rises from $69/87$ to $87/87$,
but only $36/87$ and $43/87$ PIUs, respectively, are at least partly requested.
Of those requested PIUs, $16/36$ and $21/43$ are at least partly transmitted,
and none receives full transmission credit. Of the 157 required evidence
steps, only three receive full communication credit under either policy.
These PIU and evidence-step denominators are distinct from the task-mean
scores in Table~\ref{tab:cuad-maud-transfer-pipeline}. The audit therefore
locates a persistent request-to-message bottleneck; it does not establish
which policy component causes it.

\paragraph{Completion and decision quality.}
The original MAUD mean score increases by 0.46 points, with an
agreement-cluster interval of $[-1.97,3.32]$. Among the 21 pairs where both
policies produce a final answer, the mean original score change is zero
($[-2.68,3.08]$). The five newly completed final answers have a mean score of
2.40\%; their contribution accounts arithmetically for the aggregate gain.
Individual semantic improvements and regressions can cancel within the
already-completed subset. This is a descriptive breakdown, not a causal
adjustment for completion. Missing-final/format cases decrease from seven to
zero on CUAD and from five to zero on MAUD; no original reward-judge API
failure is recorded in these paired endpoints.

\begin{table}[H]
\centering
\small
\setlength{\tabcolsep}{7pt}
\renewcommand{\arraystretch}{1.12}
\captionsetup{skip=4pt}
\caption{\revised{\textbf{Evidence demands in the paired legal cohorts.} Entries are
means per task. Assigned source words sum the private source assignments
across participants; they are not the length of a single participant's input.}}
\label{tab:cuad-maud-transfer-demands}
\revised{%
\begin{tabularx}{\linewidth}{@{}Xrr@{}}
\toprule
Cohort property & CUAD ($n=36$) & MAUD ($n=26$) \\
\midrule
Assigned source words & 6,814 & 37,773 \\
Reference evidence words & 104 & 458 \\
Participants & 2.03 & 3.08 \\
PIUs & 2.03 & 3.35 \\
Required answer components & 4.67 & 7.04 \\
\bottomrule
\end{tabularx}}
\end{table}

\paragraph{Source demands and observed budget effects.}
Both corpora contain conditions, exceptions, and party directionality; the
reference-evidence audit finds such examples in all 36 CUAD and 26 MAUD
tasks. The data therefore do not support a claim that CUAD entirely omits
these requirements. Instead, this MAUD cohort combines longer sources and
reference evidence with more distributed PIUs and required answer components
under the same short-message settings
(Table~\ref{tab:cuad-maud-transfer-demands}).

The recovered packing records retain each owner's full assigned source, with
no detected source truncation. Maximum assigned-source lengths are 16,711
tokens on CUAD and 23,992 on MAUD, below the configured 28,000-token
participant-source cap. These counts concern assigned source text, not every
token in the rendered prompt. Some participant generations reach the
64-token structured-output limit. Across both policy endpoints, 13 of 133
successfully parsed CUAD utterances and 22 of 149 MAUD utterances exceed the
48-token public-message field cap and are subject to field truncation.
No valid final decision terminates at its generation-length limit.
Qualifier loss also appears in short replies, so observed clipping alone
does not establish that the message budget causes weak transfer. No
counterfactual budget intervention is performed.

\paragraph{Interpretation and limits.}
The paired audit supports the existing distinction between behavioral and
semantic transfer: a complete query schedule and reliable final-answer
completion transfer to MAUD, while faithful evidence transmission and
integration show small, statistically uncertain changes on MAUD. The cohort has longer and more distributed evidence, and completing the
query schedule does not ensure substantial gains in communicated evidence.
The separate contributions of retrieval, query targeting, message limits, policy
specialization, and final synthesis remain unresolved. The analysis uses one
CUAD-selected training run and only six MAUD source agreements. CUAD
validation gains are subject to checkpoint-selection optimism. Audit labels
are model-generated and use the same model family as the reward evaluator;
the intervals do not capture
training-seed, rollout, or audit-model uncertainty. These results are a
descriptive account of the observed transfer pattern, not a proof of intrinsic
MAUD difficulty or a causal decomposition of the trained policies.
\FloatBarrier
}

\subsection{\CurrentSpeakerSelection{}: Shared Participant Policy Training}
\label{app:coordinator-free-policy-training}

\subsubsection{Problem Instances and Communication Structure}

This experiment uses Multi-Agent PIU-Only with \OneRIUPerParticipant{} under
\CurrentSpeakerSelection{}. The participant that ends communication makes the
final decision; there is no separate final decision maker. Training contains
350 FinDVer problem instances with two to eight participants and 188
FinLongDocQA problem instances with two or three participants. Validation
contains 39 FinDVer and 21 FinLongDocQA problem instances. The held-out test
set contains 40 four-participant FinLongDocQA problem instances. Canonical task
identifiers and source-material groups do not overlap across splits.
Four-participant FinLongDocQA is absent from training, although
four-participant FinDVer occurs in training.

Each participant receives exactly one PIU in its initial local observation. The
initial common observation contains the decision objective, decision format,
and a directory of participant identifiers and public participant descriptions.
The descriptions are short role names generated from the broad subject of the
assigned PIU. Deterministic checks and a separate leakage check reject names
that expose a value, polarity, condition, conclusion, or decision implication.
A prompt-construction check of all 638 normalized instances and 4,988 rendered prompts confirms
that every directory contains the permitted participants and that no
participant receives another participant's PIU.

In each round, the current speaker sends a public message and either selects
the next participant or ends communication and makes the final decision. If the
15-round communication budget is exhausted, the last speaker makes the final
decision. Public messages have a 600-token budget, the final-decision field has
a 1,200-token budget, participant actions have 1,200 private reasoning tokens,
and the final decision has 1,800 private reasoning tokens. The runtime also
applies a 750-token structured-generation cap, a 16,384-token prompt cap, and a
24,576-token training-sequence cap.

\subsubsection{Optimization and Checkpoint Selection}

A single Qwen2.5-7B-Instruct participant policy is shared across participants.
Participant messages, next-participant selection, decisions to end
communication, and the final decision all receive participant-policy gradients.
Each update samples eight problem instances and 16 rollouts per instance,
yielding 128 rollouts per update and 10,240 rollouts across 80 updates. Rollouts
use temperature 0.7 and top-\(p=0.95\); validation uses greedy decoding.

Training updates all parameters of the shared participant policy using FSDP in
bfloat16 on four H200 GPUs. GRPO
standardizes the 16 rewards within each problem instance. Invalid JSON,
schema-invalid actions, and invalid next-participant identifiers receive a
local advantage of \(-0.1\) and one retry. Optimization uses AdamW with learning
rate \(10^{-6}\), weight decay 0.01, gradient-norm clip 1.0, PPO ratio clip 0.2,
dual-clip bound 3.0 for negative advantages, and sequence-mean/token-mean loss
aggregation. This run does not use a KL loss.

Checkpoint selection uses the mean LLM-judged score on the 39 FinDVer
validation problem instances, with the earlier update winning a tie. The 21
FinLongDocQA validation problem instances are evaluated on the same schedule but
do not enter checkpoint selection. The selected update is update 70. Main-paper
Table~\ref{tab:finance-current-speaker-policy-training} reports the separate
post-selection greedy evaluations of the base and trained participant policies.

\subsubsection{Communication Diagnostics}

Final-decision completion records whether an episode produces a final decision.
``Every participant sends a message'' records whether every participant appears
at least once as a public-message sender. Participant coverage is the number of
distinct public-message senders divided by team size, and the number of unique
speakers is its unnormalized counterpart. Self-handoff rate is the fraction of
valid handoffs in which the current speaker selects itself as the next
participant. These diagnostics use one greedy rollout per problem instance and
retain valid public actions produced before an episode failure. None enters the
training reward.

\subsubsection{Separating the Communication Trajectory from the Final-Decision Policy}

The cross-replay analysis separates the communication trajectory from the
final-decision policy. It uses the 40 validation problem instances for which
both the base and trained communication trajectories produce a final decision.
For each trajectory, we freeze the communication transcript, the identity and
assigned PIU of the participant making the final decision, and the final-decision prompt,
then regenerate the final decision with both the base and trained participant
policies.

Let \(Y_{BB}\), \(Y_{BT}\), \(Y_{TB}\), and \(Y_{TT}\) denote the scores from
crossing the base and trained communication trajectories with the base and
trained final-decision policies. We define the communication-trajectory effect
as
\[
\frac{1}{2}\big[(Y_{TB}-Y_{BB})+(Y_{TT}-Y_{BT})\big]
\]
and the final-decision-policy effect as
\[
\frac{1}{2}\big[(Y_{BT}-Y_{BB})+(Y_{TT}-Y_{TB})\big].
\]
Intervals are percentile 95\% intervals from 100,000 paired
problem-instance-bootstrap samples.

On validation, the communication-trajectory effect is 6.95 points with a 95\%
interval of \([0.49,13.78]\). The final-decision-policy effect is \(-0.20\)
points with an interval of \([-0.99,0.61]\), and is not statistically
distinguishable from zero. The cross-replay cell means represent
policy-conditioned regenerated decisions rather than deterministic
reproductions of the original final-decision strings.

\paragraph{Illustrative trajectory pair.}
\revised{In one three-participant FinLongDocQA validation example, the team
estimates Mohawk's 2022 residential replacement sales using the approximately
50\% residential replacement share and the Flooring North America (NA) and
Rest of World (ROW) net sales. Participant A reviews the share, B reviews NA
sales, and C reviews ROW sales. The correct calculation is
$(\$4,207.0\mathrm{m}+\$3,222.3\mathrm{m})\times0.5=\$3,714.65\mathrm{m}$.
The following selected excerpts retain the agents' wording.}
\begin{quote}\small
\revised{\textbf{Base, round 3, participant B.}
``\$3,222.3 million * 50\% = \$1,611.15 million.''\par\smallskip
\textbf{Base, round 12, participant B, handing off to itself.}
``To complete the calculation, we need the Flooring NA net sales for 2022.
Could you, @participant\_b, provide this information so we can calculate the
total flooring sales for 2022?''\par\smallskip
\textbf{Trained, round 1, participant C.}
``Based on the report, the Flooring ROW net sales for 2022 were
\$3,222.3 million.''\par\smallskip
\textbf{Trained, round 2, participant B.}
``From the report, we have the Flooring NA net sales for 2022, which is
\$4,207.0 million.''\par\smallskip
\textbf{Trained, round 3, participant A.}
``The calculation would be: (\$3,222.3 million + \$4,207.0 million) * 0.5 =
\$3,714.65 million. This is an estimate because the share is approximate.''}
\end{quote}
\revised{Both trajectories use all 15 communication rounds. The base
trajectory contains seven self-handoffs and ends by turn-budget exhaustion;
its public messages give an estimate based only on ROW sales. The trained
trajectory contains no self-handoffs, reaches the correct calculation in
round 3, and largely repeats it through round 15, when B requests finalization.
The first two trained messages also include hypothetical 40\% examples,
omitted from the excerpts above.
The separate final decision is produced at turn 16 in both episodes, by A
for the base trajectory and B for the trained trajectory. In cross-replay,
both final-decision policies score 35\% with the base trajectory and 100\%
with the trained trajectory. The original base final answer scored 25\%.
The base public transcript supplies no numerical NA sales amount. This
example illustrates improved evidence sharing and a better decision with
the final-decision policy held fixed.}

\subsubsection{Additional Analysis of the Communication Trajectories}
\label{app:finance-current-speaker-trajectory-analysis}

The completion-conditioned cross-replay excludes problem instances for which
either original trajectory does not produce a final decision. We therefore run
a second analysis on all 39 paired FinDVer validation problem instances and all
40 paired FinLongDocQA test problem instances. One GPT-5.5 final-decision
policy receives the task and the valid communication transcript from each base and
trained trajectory. It does not receive the original final decision, the
participants' assigned PIUs, original score, or an indication of which policy produced the
transcript. All incomplete episodes end after an invalid participant response;
for these, the input ends with the last valid public message. We evaluate
the resulting decision with the same LLM-judged score used in the original
evaluation.

We separately show GPT-5.4 the task, reference answer, answer components,
reference PIUs, and communication transcript, without the policy identity,
original final decision, or original score. For every PIU, the evaluator
records public messages that communicate it and separately records messages
that communicate an incompatible value, unit, period, sign, or scope. It also
records whether each answer component is
fully supported by public messages. Exact repetition is computed directly
after case-folding and whitespace normalization and does not use an evaluator.

\begin{table}[H]
\centering
\scriptsize
\setlength{\tabcolsep}{2.9pt}
\renewcommand{\arraystretch}{1.07}
\caption{\textbf{Communication-trajectory analysis.} Values are means across
problem instances. Percentages are shown except for public messages. The same
39 FinDVer validation and 40 held-out FinLongDocQA problem instances are
paired across policies.}
\label{tab:finance-current-speaker-trajectory-analysis-full}
\begin{tabular}{@{}>{\raggedright\arraybackslash}p{0.33\linewidth}rrrrrr@{}}
\toprule
& \multicolumn{3}{c}{FinDVer validation} &
\multicolumn{3}{c}{FinLongDocQA test} \\
\cmidrule(lr){2-4}\cmidrule(l){5-7}
Measurement & Base & Trained & $\Delta$ & Base & Trained & $\Delta$ \\
\midrule
Same final-decision policy: LLM-judged score
& 34.10 & 46.72 & +12.62 & 49.60 & 44.82 & $-4.78$ \\
Public messages
& 7.28 & 12.46 & +5.18 & 9.30 & 13.43 & +4.12 \\
PIUs appearing in the communication transcript (\%)
& 38.0 & 53.9 & +15.9 & 63.1 & 63.1 & 0.0 \\
PIUs appearing in the communication transcript with no incompatible message (\%)
& 23.3 & 26.3 & +3.0 & 55.6 & 58.8 & +3.1 \\
Answer components fully supported (\%)
& 23.5 & 33.0 & +9.5 & 26.2 & 25.0 & $-1.2$ \\
PIUs communicated incompatibly at least once (\%)
& 30.6 & 48.0 & +17.4 & 26.9 & 25.6 & $-1.2$ \\
Exactly repeated public messages (\%)
& 18.9 & 50.9 & +31.9 & 23.1 & 48.6 & +25.5 \\
\bottomrule
\end{tabular}
\end{table}

For each measurement, the reported change is the mean within-problem-instance
trained-minus-base difference. Percentile 95\% intervals use 100,000 paired
problem-instance bootstrap samples with seed 20260922. Under the same
final-decision policy, the score changes by \(12.62\) points on FinDVer
(\([4.38,21.59]\)) and by \(-4.78\) points on FinLongDocQA test
(\([-17.75,8.20]\)). On FinDVer, the increase in the proportion of PIUs
appearing in the communication transcript is
\(15.9\) points with an interval of \([4.4,28.1]\), and the increase in fully
supported answer
components is \(9.5\) points with an interval of \([0.4,19.2]\). On
FinLongDocQA test, the corresponding changes are \(0.0\) points
(\([-11.2,11.2]\)) and \(-1.2\) points (\([-12.5,10.4]\)). Exact repetition
increases by \(31.9\) points (\([21.7,42.1]\)) on FinDVer and \(25.5\) points
(\([16.2,34.3]\)) on FinLongDocQA test.

\revised{%
\subsubsection{Variation Across Training Runs}
\label{app:proposal-finance-training-seeds}
\revised{Table~\ref{tab:proposal-finance-qwen-seeds} summarizes the multi-run
endpoint evaluations. The trajectory replay and communication diagnostics
in Appendix~\ref{app:deferred-communication-trajectory-table} and
Appendix~\ref{app:finance-current-speaker-trajectory-analysis} retain the
original run's recorded episodes and baselines.}
\begin{table}[H]
\caption{\revised{\textbf{Qwen2.5-7B-Instruct finance scores and selected updates across three training runs.}
Scores are percentages. Test gains are percentage points with paired 95\%
source-group bootstrap intervals; the final row reports mean $\pm$ sample SD
across runs.}}
\label{tab:proposal-finance-qwen-seeds}
\centering\small
\setlength{\tabcolsep}{3.5pt}
\renewcommand{\arraystretch}{1.08}
\revised{%
\begin{tabular}{@{}lrrrrl@{}}
\toprule
Run & \makecell{Selected\\update}
& \makecell{FinDVer val.\\$n=39$}
& \makecell{FinLongDocQA\\val. ($n=21$)}
& \makecell{FinLongDocQA\\test ($n=40$)}
& \makecell{Test gain\\{}[95\% CI]} \\
\midrule
Base & --- & 25.95 & 54.05 & 25.53 & --- \\
A & 70 & 41.08 & 69.52 & 41.13 & $+15.60\;[4.85,26.35]$ \\
B & 80 & 40.77 & 69.52 & 49.50 & $+23.98\;[13.13,34.60]$ \\
C & 70 & 34.51 & 64.05 & 31.88 & $+6.35\;[-5.95,18.98]$ \\
\midrule
Mean $\pm$ SD & --- & $38.79\pm3.70$ & $67.70\pm3.16$ & $40.83\pm8.82$ & --- \\
\bottomrule
\end{tabular}}
\end{table}
}

\revised{%
\subsubsection{Llama-3.1-8B-Instruct}
\label{app:proposal-llama-finance-pilot}
We train one Llama-3.1-8B-Instruct policy from its own frozen starting weights
using the same finance tasks and 80-update budget. FinDVer validation selects
update~70; Table~\ref{tab:proposal-finance-llama-pilot} reports its results.

\begin{table}[H]
\caption{\revised{\textbf{Finance results with Llama-3.1-8B-Instruct.}
Scores are percentages; gains are percentage points with paired 95\%
source-group bootstrap intervals.}}
\label{tab:proposal-finance-llama-pilot}
\centering\small
\setlength{\tabcolsep}{4pt}
\revised{%
\begin{tabular}{@{}lrrrr@{}}
\toprule
Cohort & $n$ & Base & Trained & Gain [95\% CI] \\
\midrule
FinDVer val. & 39 & 15.46 & 34.64 & +19.18 [+11.56, +27.26] \\
FinLongDocQA val. & 21 & 27.62 & 50.95 & +23.33 [+8.33, +37.86] \\
FinLongDocQA test & 40 & 17.50 & 36.33 & +18.83 [+9.83, +27.95] \\
\bottomrule
\end{tabular}}
\end{table}
}

\end{document}